\RequirePackage{fix-cm}
\documentclass[smallextended]{svjour3}       % onecolumn (second format)
\smartqed  % flush right qed marks, e.g. at end of proof
\usepackage{graphicx}
\usepackage{soul}
\usepackage{bm}
\usepackage[ruled,linesnumbered,noend]{algorithm2e}
\usepackage{amsfonts}
\usepackage{url}
\usepackage{booktabs}
\usepackage{tikz-cd}
\usepackage{amssymb}
\usepackage{lscape}
\usepackage{amsmath}
\usepackage{subcaption}
\usepackage{rotating}
\DeclareMathOperator*{\argmax}{arg\,max}

\DeclareMathOperator{\arccosh}{arcCosh}
\SetKwComment{Comment}{$\triangleright$\ }{}

\newcommand{\norm}[1]{\left\lVert#1\right\rVert}
\begin{document}

\title{ReliefE: Feature Ranking in High-dimensional Spaces via Manifold Embeddings
%High dimensional feature ranking via manifold embeddings
}

\author{Bla\v{z} \v{S}krlj \and Sa\v{s}o D\v{z}eroski  \and Nada Lavra\v{c} \and Matej Petkovi\'{c}}


\institute{Bla\v{z} \v{S}krlj, Sa\v{s}o D\v{z}eroski, Matej Petkovi\v{c} \at
             Jo\v{z}ef Stefan Institute, Jamova 39, 1000 Ljubljana, Slovenia \\Jo\v{z}ef Stefan International Postgraduate School, Jamova 39, 1000 Ljubljana, Slovenia \\ \email{blaz.skrlj@ijs.si, saso.dzeroski@ijs.si, matej.petkovic@ijs.si} \\
             Nada Lavra\v{c}\\
             Jo\v{z}ef Stefan Institute, Jamova 39, 1000 Ljubljana, Slovenia \\
             University of Nova Gorica, Vipavska 13, 5000 Nova Gorica, Slovenia \\
             \email{nada.lavrac@ijs.si}\\
             }

\date{Received: date / Accepted: date}

\maketitle
\begin{abstract}
Feature ranking has been widely adopted in machine learning applications such as high-throughput biology and social sciences. The approaches of the popular Relief family of algorithms assign importances to features by iteratively accounting for nearest relevant and irrelevant instances. Despite their high utility, these algorithms can be computationally expensive and not-well suited for high-dimensional sparse input spaces. In contrast, recent  embedding-based methods learn compact, low-dimensional representations, potentially facilitating down-stream learning capabilities of conventional learners. This paper explores how the Relief branch of algorithms can be adapted to benefit from (Riemannian) manifold-based embeddings of instance and target spaces, where {a given embedding's} dimensionality is intrinsic to the dimensionality of the considered data set.
The developed ReliefE algorithm is faster and can result in better feature rankings, as shown by our evaluation on 20 real-life data sets for multi-class and multi-label classification tasks. The utility of ReliefE for high-dimensional data sets is ensured by its implementation that utilizes sparse matrix algebraic operations.  
{Finally, the relation of ReliefE to other ranking algorithms is studied via the Fuzzy Jaccard Index}.
\end{abstract}

\section{Introduction}
\label{intro}

Contemporary machine learning has found its use in many scientific disciplines, ranging from biology, sociology, logistics, engineering sciences to physics. Data sets are often available in tabular form, and consist of instances (rows) and features (columns), where attributes denote column names, and individual features correspond to individual attribute values.

Even though predictive models can offer insights into how well a certain aspect of a given system can be predicted, researchers and industry practitioners are frequently interested in \emph{which} parts of the input space are the most relevant. Having such knowledge can yield novel insights into relevant aspects of the studied problem, leading to improved human understanding of the studied phenomenon. For example, in modern molecular and systems biology, discovery of novel biomarkers is of {high} relevance---once the researchers know which, e.g., compounds or proteins indicate the presence of the studied condition, they can be used for preliminary condition detection, but also to {advance} the human understanding of the conditions leading to the construction of novel hypotheses. {We next discuss the types of feature ranking algorithms.}

{Feature ranking algorithms can be split into two main groups: myopic and non-myopic. Myopic algorithms do not consider multiple features simultaneously and thus potentially neglect \emph{interactions} between features. Examples of myopic feature ranking algorithms include, e.g., information gain-based ranking.}
Algorithms from the Relief branch, originating in the early Relief algorithm \cite{kira1992feature}, are among the most widely used non-myopic algorithms for \emph{feature ranking}, where each feature is assigned a real-valued score, offering insights into its importance. 
The Relief family of algorithms has been successfully applied to numerous real-life problems \cite{stiglic2010stability,stokes2012application,PETKOVIC2021104143}.
In this work we propose ReliefE, an embedding-based feature ranking algorithm built on the ideas of the original Relief and ReliefF~\cite{robnik2003theoretical}, as well as their extensions to a multi-label classification setting \cite{mlcrelief}. ReliefE does not compute feature importances based on the original, high-dimensional feature space, but via low-dimensional embeddings of the input and/or output spaces. 
{The} key contributions of this work are summarized as follows:
\begin{itemize}
    \item We {present} ReliefE, an algorithm for feature ranking implemented using sparse matrix algebraic computation of distances between low-dimensional manifold embeddings of both instances and targets ({when considering multi-label classification}).
    \item The latent dimension of the space, in which the distances are computed, is \emph{inferred automatically} in an efficient manner.
    \item We {show that} the number of neighbors to be considered can be {automatically} inferred based on the distribution of distances to the considered instances, rather than hard-coded as part of the input.
    \item Theoretically grounded sparsification of the input was considered as a preprocessing step, potentially decreasing {the} execution time.
    \item We offer evidence that ReliefE performs significantly faster than many state-of-the-art methods, especially in high-dimensional input (and output) spaces.
    \item Theoretical properties of ReliefE, including the properties of the embedding spaces, their relations and computational complexity analysis are {analysed}.
    \item We showcase the ReliefE's performance against six strong (widely used) baselines on 20 real-life multi-class and multi-label classification data sets.
    \item We perform extensive Bayesian and frequentist performance comparisons assessing the statistical evidence of ReliefE's utility and potential drawbacks.
\end{itemize}

The {rest} of this paper is structured as follows. In Section~\ref{sec:related}, we discuss the key ideas that have led to the developments described in this paper. Section~\ref{sec:proposed} presents the proposed ReliefE methodology, followed by a description of the experimental setting in Section~\ref{sec:experimental} and the results of the empirical evaluation in Section~\ref{sec:results}. The overall findings are discussed in Section~\ref{sec:discussion}, followed by the conclusions and plans for further work in Section~\ref{sec:conclusions}. The paper includes numerous appendices presenting detailed results of empirical comparisons and case studies.

\section{{Background}}
\label{sec:related}
In this section we discuss the works that have impacted this paper the most, starting with the notion of feature ranking and the description of the Relief branch of feature ranking algorithms. Next, we discuss how embedding-based learning can be of use when solving otherwise intractable problems, serving as a motivation for the proposed work.

\subsection{Feature ranking}

Feature ranking can be considered as the process of learning to prioritize the feature space with respect to a given learning task. {Algorithms that rank features can operate in non-myopic (considering feature interactions) or myopic manner (ignoring feature interactions).} One of the first and most widely used algorithms for non-myopic feature ranking is Relief \cite{kira1992feature}, introduced in the early 1990s. This iterative algorithm operates by randomly selecting parts of the instance space (e.g., rows), followed by iterative update of weights corresponding to individual features based on the closest instances from the positive and negative classes. {The} original Relief {performs} well for binary classification, however was unable to generalize to more complex learning tasks such as multi-class classification. Its extension, ReliefF \cite{robnik2003theoretical}, introduced a prior-based weighting scheme that can take different classes into account. In the following years, multiple adaptations of both Relief and ReliefF were introduced, varying mostly {in terms of} schemes for taking into account a given instance's neighborhood and its (aggregated) properties. For example, 
SURF \cite{greene2009spatially} unifies the considered per-class neighborhoods, whereas SURFstar \cite{greene2010informative} additionally considers
distant neighbors.
Further, MultiSURFstar \cite{granizo2013multiple} takes neighborhood boundaries into account based on the average and standard deviation of distances from the target instance to all others. The TuRF adaptation \cite{moore2007tuning} of any Relief algorithm also employs recursive feature elimination, whilst applying the dedicated Relief implementation iteratively during feature pruning. TuRF attempts to address some of the problems that arise in very large feature spaces (e.g., more than 20{,}000 features), yet at the cost of higher computational complexity. In terms of scalability, for example, VLSReliefF \cite{eppstein2008very} samples \emph{random subspaces} whilst simultaneously offering competitive performance at a far lower computational cost. The above ReliefF variants mostly attempt to correct some of the original ReliefF drawbacks by re-considering the update step and how the neighbors are taken into account.

In recent years, the Relief algorithms have been also extended to a multi-label classification setting {(MLC)} -- a learning problem where multiple labels for an instance are simultaneously possible. {Examples of this task include, for example, gene function prediction.}
The first attempt of extending ReliefF to MLC that scales well with the number of labels \cite{mlcrelief},
uses the Hamming distance as the distance measure between two label sets. 
Since Hamming loss can be expressed as a sum of component-wise differences, it is not able to detect the possible label-label interactions, which may lead to sub-optimal quality of the obtained rankings as shown recently \cite{mlcrelief}, where other MLC error measure-based distances
between labels were shown to offer superior performance.

\subsection{Embeddings}
The rise of embedding-based learning can {be nowadays} observed in virtually all areas of science and industry. Since the 2010s, for example, deep neural networks are successfully used in fields such as computer vision, where state-of-the-art results are consistently demonstrated, e.g., in face recognition and anomaly detection \cite{lecun2015deep,pouyanfar2018survey}. Further, in recent years, natural language processing has been transformed first by the introduction of recurrent neural networks, followed by attention-based neural networks (transformers), showing prominent results in the areas of question answering, language understanding and text classification \cite{vaswani2017attention}. Similar results have been observed in the areas of network mining \cite{kipf2016semi} and time series analysis (based on the initial work of \cite{connor1994recurrent}).

Neural networks offer an elegant alternative for learning a latent representation of the input data set, that can be \emph{transferred}, as well as directly used for problem solving. More recent works on representation learning, however, suggest that a low-dimensional manifold is a suitable topological object for learning rich and transferable representations \cite{bronstein2017geometric,masci2015geodesic}. Even though representations learned by neural networks can be associated with manifold learning, the development of algorithms capable of approximating such low-dimensional manifolds {has} been an active research area even before the era of deep learning, and is of particular relevance to this work. For example, algorithms such as Isomap \cite{balasubramanian2002isomap} and Locally linear embedding \cite{roweis2000nonlinear} have been successfully used for data visualization and more efficient learning. Further, modern \emph{omics} sciences have widely adopted t-SNE \cite{maaten2008visualizing}, an approximation method capable of producing two dimensional embeddings of high-dimensional spaces, making it suitable for e.g., gene expression visualization. 
Hyperbolic embeddings have been also recently demonstrated {to be useful} when considering hierarchical multi-label classification~\cite{stepisnikHyperbolic}.

This work builds on the notion of uniform manifold projections (UMAP) \cite{mcinnes2018umap,mcinnes2018umap-software}, a recently introduced algorithm with a strong theoretical grounding in \emph{manifold theory}. We explore, whether low-dimensional manifold approximations, derived from sparse input spaces, can be a natural extension to the Relief family of algorithms.

\section{Proposed methodology}
\label{sec:proposed}
One of the main criticisms of the Relief branch of algorithms is their lack of ability to handle very high-dimensional, potentially sparse input spaces, where problems arise either due to increased space complexity or too incremental weight update steps that result in similar importance scores for many features (i.e. non-informative rankings). We have developed the proposed ReliefE algorithm with the goal to addresses these issues. In this section, we discuss in detail ReliefE, the proposed embedding-based feature ranking algorithm for multi-class and multi-label classification problems, along with its implementation, currently one of the fastest Python-based implementations compiled to machine code capable of handling both multi-class and multi-label classification problems. 
\begin{figure}[h]
    \centering
    \includegraphics[width = .83\linewidth]{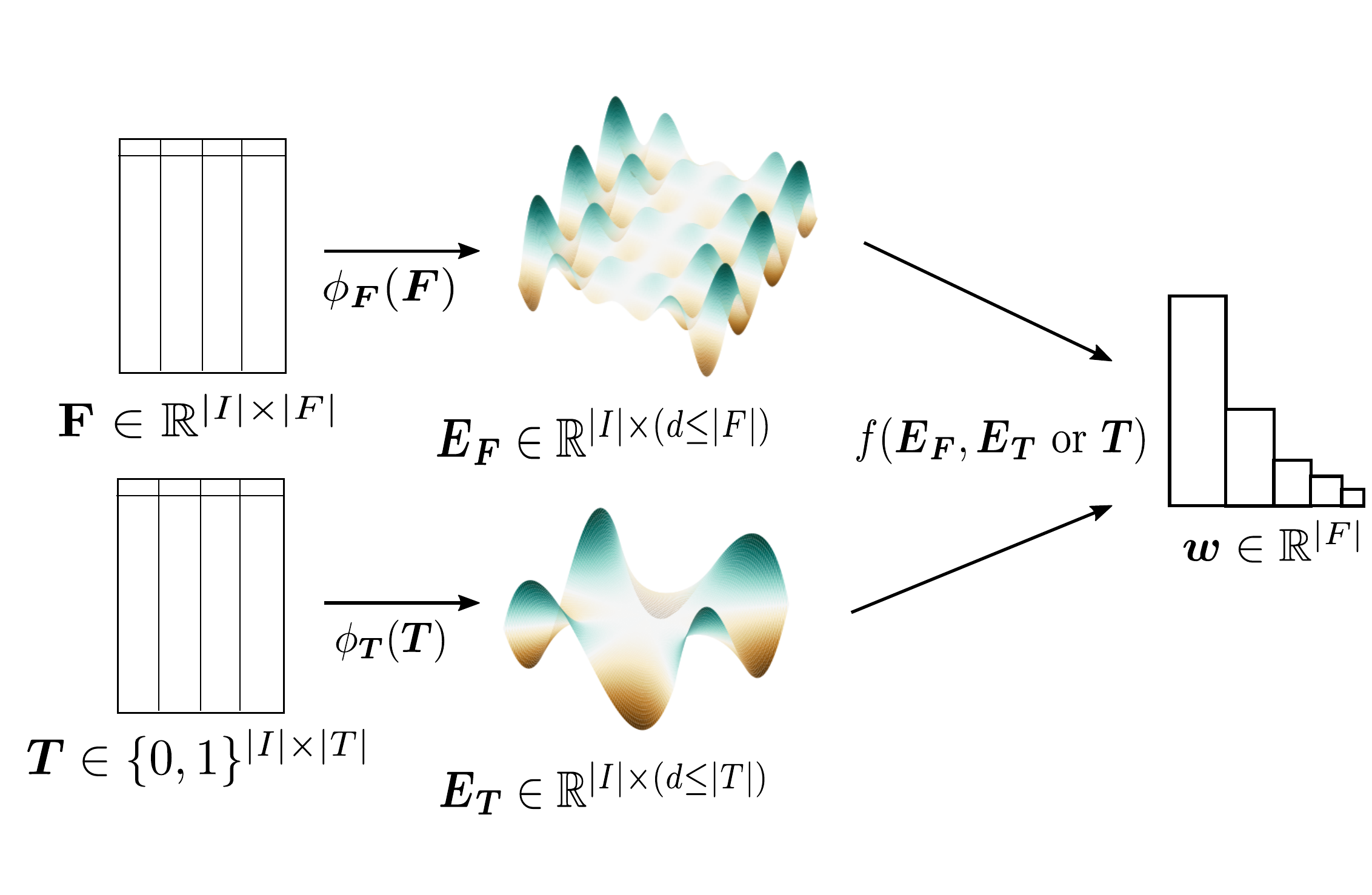}
    \caption{Overview of the core idea behind ReliefE. {Distances in both the feature and target space are computed based on instance embeddings.}}
    \label{fig:scheme}
\end{figure}
The proposed ReliefE algorithm is summarized in Figure~\ref{fig:scheme}. Here, the input feature space ($\boldsymbol{F}$) is mapped (by $\phi_{\boldsymbol{F}}$) to its corresponding low-dimensional approximation ($\boldsymbol{E_F}$). Finally, the Relief feature ranking ($f$) is adapted to operate via this low-dimensional representation to obtain final feature rankings $\boldsymbol{w}$. Further, in a multi-label setting, distances between targets ($\boldsymbol{T}$) can also be computed in the latent space $\boldsymbol{E_T}$, constructed via $\phi_{\boldsymbol{T}}$.

%The key insight conveyed in the image is the fact that feature importances are estimated based on the distances between instances in either the input or the target space.
%\emph{regardless of their representation}.

\subsection{Rationale for embedding-based ranking}
\label{sec:rationale}
Embedding the {instances in the} feature space prior to feature ranking is sensible due to the fact that volume increases \emph{exponentially} with dimension. Many classifiers benefit from increasing the number of dimensions, however, once a certain dimensionality is reached, their performance starts to degrade \cite{hughes}. Feature ranking aids this problem by prioritizing parts of the feature space that are relevant for learning.

Higher-dimensional feature spaces render feature importane detection in the {initial feature space} harder, as more subtle differences between instances need to be taken into account. Embedding (in an unsupervised manner) offers {the construction} of instance representations that potentially carry additional 
\emph{semantic context}, as comparing instances in the embedded space does not  compare only the considered pairs of instances but also their potential \emph{roles} in the joint latent space, offering more meaningful comparisons (as shown in e.g., \cite{NIPS2013_5021} for words and phrases). We next discuss the embedding method considered throughout this work.

\subsection{Uniform Manifold Approximation and Projection}
\label{sec:umap}
This work builds on the recently introduced idea of Uniform Manifold Approximation and Projection (UMAP) \cite{mcinnes2018umap} for the task of low-dimensional, unsupervised representation learning. Even though a detailed treatment of the theoretical underpinnings of UMAP are beyond the scope of this paper, we discuss some of the key ideas underlying the actual implementation, and refer the interested reader to \cite{mcinnes2018umap} for a detailed overview of the theory.

UMAP is formulated with concepts from both topological manifold theory and applied category theory. Riemannian manifolds are topological spaces that have a locally constant metric, and are locally connected. UMAP assumes that high dimensional data can be uniformly mapped across a low-dimensional Riemannian manifold.
The locality of the metric, connectivity and uniformity of the mapping are the three main assumptions of this method.
Even though such assumptions are not necessarily {fulfilled}, appropriate selection of the metric {used by UMAP} can offer better performance and generalization when {the} learned representations are used for down-stream learning tasks. In contrast to t-SNE, which is effective for two dimensional embeddings,  UMAP is highly efficient {also} for embeddings into \emph{higher-dimensional} vector spaces. UMAP thus serves as a general unsupervised representation learner\footnote{UMAP can also perform supervised embeddings, yet this functionality is not considered in this work.}. It has been successfully used for exploration of biological and other high-dimensional data sets \cite{Cao2019}.
{In summary,} the topological manifold theory underlying UMAP offers a very general {representation} learning framework, applicable beyond the current implementation of UMAP, which we discuss in more detail below.

Even though the original formulation assumes continuity, in practice, discrete graph-theoretical analogs of the continuous concepts are employed. 
{The representations are a result of a two-step procedure, where in the first step, a weighted graph is constructed based on the distances between the closest instances. The second step resembles conventional graph layout computation, which is normally computed in two or three dimensions for visualization purposes, where, e.g., two coordinates of a 2D layout represent two features. Analogously, UMAP extends the idea to higher dimensions, where the instances are positioned in a $d$-dimensional space (with $d$ going up to several hundred in most cases). The resulting space is not useful for direct visualization, but serves as a representation suitable for a down-stream learning task---in this work, feature ranking.}
The computation of the {UMAP} embedding can be described as follows.
\begin{description}
\item \textbf{Weighted graph construction}. 
Assume a user-specified dissimilarity measure $\delta: \mathbb{R}^{|F|} \times \mathbb{R}^{|F|} \to [0, \infty)$ and the number of nearest neighbours $k$, computed via an approximation algorithm \cite{dong2011efficient}. We refer to the ordered set of instances nearest to instance $\boldsymbol{r}_i$ as $\{\boldsymbol{r}_{i_{1}}, \dots, \boldsymbol{r}_{i_{k}}\}$. For each instance, let
\begin{equation*}
    \omega_i = \min{ \{\delta(r_i, r_{i_{j}}) | 1 \leq j \leq k, \delta(r_i, r_{i_{j}}) > 0 \} },
\end{equation*}
\noindent and $\beta_i$ such that
\begin{equation*}
    \sum_{j = 1}^{k} \exp \bigg ( { \frac{ - \max(0, \delta(r_i, r_{r_{j}}) - \omega_i)}{\beta_i}} \bigg ) = \log_2(k).
\end{equation*}
\noindent Next, a weighted directed graph $G = (N,E,w)$ is constructed, where $N$ is the set of considered instances, $E = \{ (r_i,r_{i_{j}}) | 1 \leq j \leq k\}$ and 
\begin{equation*}
    w((r_i,r_{i_j})) = \exp \bigg ( \frac{- \max(0, \delta (r_i, r_{i_j}) - \omega_i)}{\beta_i} \bigg ).
\end{equation*} 
The final adjacency matrix $\boldsymbol{B}$ is computed as $\boldsymbol{B} = \boldsymbol{A} + \boldsymbol{A}^T - \boldsymbol{A} \odot \boldsymbol{A}^T$ where, $\boldsymbol{A}$ is the adjacency matrix of $G$ and $\odot$ denotes the Hadamard product.

\item \textbf{Layout computation}. In the second step, the graph undergoes a process resembling energy minimization, where repulsive and attractive forces are iteratively applied across pairs of instances, resulting in a $d$-dimensional layout, which is effectively a real-valued embedding.
The attractive forces ($F_+$) are computed, for a given pair of vertices $n_i$ and $n_j$ and their corresponding coordinate representations (embeddings) $\boldsymbol{r}_i$ and $\boldsymbol{r}_j$ as follows:
\begin{equation*}
    F_+ = \frac{-2 \cdot a \cdot b \cdot \vert \vert \boldsymbol{r_i} - \boldsymbol{r_j} \vert \vert_{2}^{2(b-1)}}{1 + \vert \vert \boldsymbol{r_i} - \boldsymbol{r_j} \vert \vert_{2}^{2}} \cdot w(n_i,n_j) \cdot (\boldsymbol{r_i} - \boldsymbol{r_j}),
\end{equation*}
\noindent and similarly, the repulsive forces ($F_-$) are computed as
\begin{equation*}
    F_- = \frac{b \cdot (1 - w(n_i,n_j)) \cdot (\boldsymbol{r_i} - \boldsymbol{r_j})}{(\eta + \vert \vert \boldsymbol{r_i} - \boldsymbol{r_j} \vert \vert_{2}^{2}) (1+ \vert \vert \boldsymbol{r_i} - \boldsymbol{r_j} \vert \vert_{2}^{2})}.
\end{equation*}
\noindent Here, $\eta$ is introduced for numerical stability (a small constant), while $a$ and $b$ are hyperparameters. Note that the $F_-$ update is computationally very expensive, and is addressed via sampling. The initial coordinates are computed by using spectral layout---here, the two eigenvectors corresponding to the two largest eigenvalues are used as the starting set of coordinates.
\end{description}
The two steps, when implemented efficiently, offer fast construction of $d$-dimensional, real-valued representations. 
The considered UMAP implementation exploits the Numba framework \cite{lam2015numba} for compiling parts of Python code, making it scalable whilst maintaining user-friendly API execution. Note that UMAP's computational complexity depends heavily on the approximate $k$-nearest neighbor computation.
In the following sections, we discuss how we further facilitate the embedding computation, as the current version of UMAP still has \emph{high space complexity} (graph construction).

\subsection{Input sparsification}
\label{sec:sparsification}
The proposed methodology is capable of handling highly sparse inputs without additional memory overheads. However, real-life {data frequently} comes in {the} form of dense matrices, as is the case with, e.g., gene expression data sets. As a part of the proposed methodology, we explore whether input sparsification can speed up the feature ranking process with minimal ranking quality degradation. We implement the recently introduced, theoretically grounded, Probabilistic Matrix Sparsification algorithm (PrMS) for matrix sparsification \cite{arora2006fast}, given in Algorithm~\ref{algo:sparsify}.

\begin{algorithm}[b!]
\DontPrintSemicolon
\caption{Probabilistic Matrix Sparsification (PrMS)
\cite{arora2006fast}}
\label{algo:sparsify}
\KwData{Input matrix $\boldsymbol{A}$, number of elements $n$, approximation constant $\epsilon$}
\For{$i,j \in n$}{
    \eIf{$|\boldsymbol{A}_{ij}| > \frac{\epsilon}{\sqrt{n}}$}{
        $\hat{\boldsymbol{A}_{ij}} = \boldsymbol{A}_{ij}$\;
    }{
        $\hat{\boldsymbol{A}}_{ij} = \begin{cases} \textrm{sgn}(\boldsymbol{A}_{ij}) \cdot \frac{\epsilon}{\sqrt{n}}; \textrm{ with probability }p_{ij} = \frac{\sqrt{n}|\boldsymbol{A}_{ij}|}{\epsilon} \\
        0; else
        \end{cases}$
    }
}
\KwRet $\hat{\boldsymbol{A}}$\;
\end{algorithm}
The mathematical intuition behind PrMS is as follows. Given a real-valued matrix $\boldsymbol{A} \in \mathbb{R}^{n \times n}$, let $S = \sum_{ij}|A_{ij}|$. A single PrMS pass through $\bm{A}$ guarantees at least $\mathcal{O}(\frac{\sqrt{n}\cdot S}{\Delta})$ elements with probability $1 - \exp{(- \Omega(\frac{\sqrt{n}\cdot S}{\Delta}))}$. Here, $\Omega$ represents the lower bound, and $\Delta = \epsilon / \norm{\bm{A}}_2$, where parameter $\epsilon > 0$ determines the approximation level.
Further, with probability $1 - \exp{ (- \Omega(n))}$, $\lvert \lvert \boldsymbol{A} - \hat{\boldsymbol{A}} \rvert \rvert_2 \leq \mathcal{O}(\epsilon)$ holds\footnote{For extensive theoretical treatise, please consult \cite{arora2006fast}.}. Note that, as the majority of real-life data sets are not {represented by} symmetric matrices (typically, they are not even square matrices), the transformation $\bm{B}\mapsto \bm{A} = \begin{bmatrix}0 &\bm{B}\\ \bm{B}^T& 0 \end{bmatrix}$ of the initial matrix $\boldsymbol{B} \in \mathbb{R}^{m \times n}$ has to be employed since $\bm{A}$ is symmetric and $\norm{\bm{A}}_2 = \norm{\bm{B}}_2$. 
 We consider {$\epsilon = \norm{\bm{A}}_\infty / (m + n)$}, i.e. {the maximal column-average (of absolute values) of matrix $\bm{A}$.}
\noindent The sparsification procedure remains highly dependent only on $\epsilon$, the parameter determining the reconstruction error that is allowed.

We have used this estimate of $\epsilon$ as it is fast to compute and avoids the need for user specification of $\epsilon$, whilst simultaneously guaranteeing reasonable performance (see Section~\ref{sec:results}). One of the most crucial hyperparameters related to representation learning in general is the dimension of the space in which the constructed representation resides. We have attempted to automate the choice of this---otherwise hard-coded---parameter and discuss the {considered} estimate next.
\subsection{Estimation of latent dimension}
\label{sec:intrinsic}
Following \cite{facco2017estimating}, we improve upon the idea of latent dimension estimation via top two nearest neighbors. {To compute the latent dimension $d$ of the data (under the assumption of a locally constant probability density), it suffices to define two (hyper)spheres $S_1$ and $S_2$. Both are centred at a random data sample and have radii equal to the distances between the sample and its two nearest neighbors} \cite{facco2017estimating}. {The radii and the dimension $d$ define the volumes of the spheres and it turns out that the value of $d$ can be easily estimated from the empirical probability distribution of the ratio $V(S_2\backslash  S_1)/ V(S_1)$ of the volumes of the shell $S_2\backslash S_1$ and sphere $S_1$.} The method, implemented as a part of ReliefE is summarized in Algorithm~\ref{algo:dim}.

\begin{algorithm}[h!]
\DontPrintSemicolon
\caption{Latent dimension estimation  \cite{facco2017estimating}.}\label{algo:dim}
\KwData{Set of instance indices $I$, (sparse) feature matrix $\textbf{F} \in \mathbb{R}^{|I| \times |F|}$}
distanceMatrix $\leftarrow $ pairwiseDistancesNonzero($\boldsymbol{F}$)\;
$\mu \leftarrow $ distanceMatrix[:,1]/ distanceMatrix[:,0]\;
empiricalDist $\leftarrow$ computeEmpiricalDistribution($\mu$)\;
$d \leftarrow$ leastSquaresLine(log($\mu$), log(1 - empiricalDist))\;
\KwRet $d$\;
\end{algorithm}

\begin{figure}[b!]
    \centering
    \includegraphics[width = .7\linewidth]{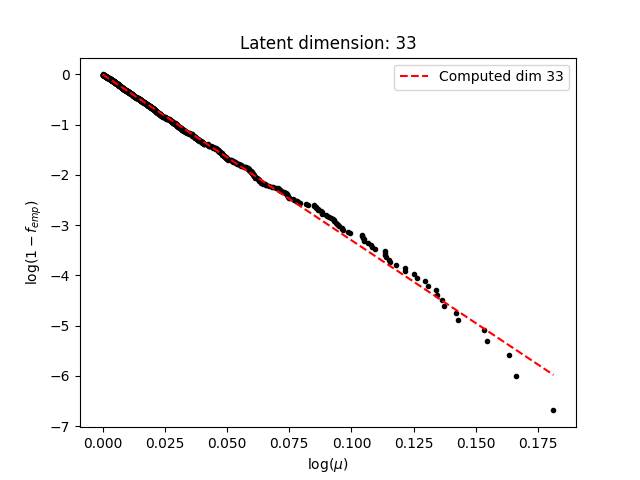}
    \caption{Intrinsic dimension for the \emph{genes} \cite{weinstein2013cancer} data set.}
    \label{fig:intrinsic}
\end{figure}

The algorithm first computes distanceMatrix, i.e. a matrix comprised of distances to the top two nearest neighbors. In this work, we improve upon the original idea of simply computing the
two nearest neighbors of a given instance.
{Instead, we ignore the neighbors that are equal to a given instance (are at distance $0$), and take into account the two nearest neighbors that are at a positive distance to the given instance (method pairwiseDistancesNonzero). The rationale for this step is that this method is also used on the output space of multi-label classification data, where two (or more) examples can often have the same output value (vector describing the labels assigned to the instance). If we had followed the original method to the letter, some components of the vector $\mu$ would equal $\infty$ or NaN. Using the modified procedure, numerical instability is avoided whilst observing the same, or very similar, results.}
Next, $\mu$, a quotient between the two distance vectors is computed (second closest against closest). An empirical distribution is derived from $\mu$. This distribution can be defined as:
\begin{equation*}
    \textsc{EMP}_n(x) = \frac{1}{|I|} \sum \mathbb{I}_{x_i < x},
\end{equation*}
\noindent where $\mathbb{I}$ represents the indicator function (presence). Thus, for a given $x$, $F_n$ represents the relative number of elements that are smaller than $x$.
The logarithms of $\mu$, as well as (1 - $\textsc{EMP}_n(x)$) are computed next.
The line between the two quantities intersects 0, and its coefficient, when rounded to the nearest integer, corresponds to the estimated latent dimension.
For example, the intrinsic dimension of the \emph{genes} data set is estimated to 33 (see Figure \ref{fig:intrinsic}).

%We next discuss the improvements we introduced to the original ReliefF, followed by the formal description of the proposed algorithm.

\subsection{Scaling up UMAP: Learning to embed based on representative subspaces}
\label{sec:subspaces}
As the most apparent memory (space) bottleneck, we (empirically) identified the UMAP's graph construction phase, where the entirety of the instances is used for learning to approximate the low-dimensional manifold (embedding). 
{This paper explores whether representative subspaces of the instance space can serve similarly well for learning to embed.}
The idea was inspired by recent work on random output space selections for ensemble learning \cite{Breskvar2018}.
The two adaptations we needed to consider were the initialization and training on a representative subspace of the instances. {In the original UMAP, the initialization is based} on {the} spectral decomposition of the instance graph. We instead considered random initialization, which already notably reduced the memory requirement, and kept the quality of the ranking at the same level. However, on the larger data sets (see Section~\ref{sec:experimental}), the embedding computation was still beyond the capabilities of an {off-the-shelf computer (Lenovo Carbon X1)}. To overcome this issue, we employ the following two-step sampling scheme. In the first step, the set of target values appearing in the data is ordered into a list. In the second step, we \emph{cyclically} iterate through the list and (without repetition) choose an element with a given target value, or skip this value if all {samples labeled with the considered target value} have already been chosen.

The multi-class (and binary) classification examples thus adhere to straightforward mapping from possible classes to the corresponding \emph{multisets} of instances.
We also extended this idea to the task of multi-label classification (MLC) where multiple labels can be simultaneously assigned a given example.
However, a finite number of possible different label sets allows for first enumerating them, and then proceeding as in the multi-class classification scenario.
We consider the theoretical properties of this procedure in Section~\ref{sec:theoretical}.

\subsection{Adaptive neighbor selection and comparison to average neighbor representation}
\label{sec:adaptive}

The first improvement we introduce {next} removes the hyperparameter $k$---the number of neighbors considered for an update step w.r.t. a single randomly chosen instance. Commonly, $k$ is a user defined hyperparameter: However, we explored whether this part of the user input can be removed entirely and replaced by a distance distribution-based heuristic that \emph{dynamically allocates} a number of neighbors suitable for a given randomly selected instance, albeit at some additional computational cost.
The rationale for such a heuristic is that real-life data sets are often not uniformly distributed \cite{liu2005investigation}, indicating only a few other instances are mostly well connected with a given one (scale-free property).

We implement this (optional) estimation as follows. For each instance, ReliefF computes its distance to the remainder of the other instances in order to obtain the top $k$ nearest ones. Such a hard-coded selection scheme is not optimal, as it does not take into account the shape of the distance distribution with respect to an individual instance. To overcome this issue, we propose the following procedure:
\begin{enumerate}
    \item Sort distances w.r.t. the selected instance $\boldsymbol{r}_i$.
    \item Compute the difference between each pair of consequent entries in the distance vector.
    \item Select the value of $k$ based on the maximum difference observed.
\end{enumerate}
This procedure intuitively takes into account the shape of the distance distribution as follows.  Assuming that all instances are similarly far from the selected instance, the difference vector will mostly consist of small values (a given pair of distances is not all that different). This can result in large $k$, as the global difference maximum can occur very late. On the contrary, if only a handful of instances are close, and the remainder is very far, only this handful shall serve to determine $k$, which will be consequently lower. Furthermore, if very high $k$ is selected and all distances are approximately the same, their mean will be similar no matter how many are selected. If a similar situation is considered for highly non-uniform distance distribution, the mean of selected $k$ nearest instances should represent only the ones that are indeed similar to the selected instance, and do not take into account the remainder which is further and possibly not as relevant.

Even if the standard IID assumption holds when sampling a data set, we are always given a finite sample. The neighborhoods of the samples on the border of the convex hull of the data set are most probably different than the neighborhoods of those in the interior of the hull. Additional theoretical properties of this neighbor number selection are given in Section~\ref{sec:theoretical}, where computational complexity is also considered.

The proposed adaptation of ReliefF was further extended with an additional option to use the \emph{average neighbor} for the update step, instead of performing updates based on all neighbors and averaging the updates. Albeit subtle, this difference potentially improves the update part's robustness, making the update step less prone to possible outliers amongst the nearest neighbors---such situations could emerge, if the value of $k$ is hard-coded, as is commonly the practice. The intuition behind this incremental change in weight update is as follows. As the distance computation is conducted in {the} \emph{latent} space, averaging the representation of the neighborhood can also be considered as constructing a \emph{semantic representation} of the considered instance's neighborhood. %As it is not entirely clear whether this consideration improves the final ranking, we implemented it as a part of this work. We next present the proposed algorithm in a formal manner.

\subsection{ReliefE - ranking via manifold embeddings}
\label{sec:reliefE}
In the following sections, we discuss more formally the proposed ReliefE algorithm  incorporating the possible improvements stated in the previous sections.
The solution (that handles both multi-class and multi-label classification) is given as Algorithm~\ref{algo:ttv}.
We can see that this is an iterative algorithm where a random sample $\bm{r}$ is chosen on each of the $n$ iterations, and distances between $\bm{r}$ and the remaining instances are computed (lines \ref{line:sample} and \ref{line:distances}).
We next discuss the two main parts of ReliefE with respect to the addressed classification task.
\begin{algorithm}[t!]
\caption{ReliefE.}
\label{algo:ttv}
\DontPrintSemicolon
\KwData{ feature set $F$, instance index set $I$, (sparse) feature matrix $\textbf{F} \in \mathbb{R}^{|I| \times |F|}$,  target classes $C$, (sparse) target space $\textbf{T} \in \{0,1\}^{|I| \times |C|}$, neigborhood size $k$, boolean adaptiveThreshold, distance score $\delta$, latent dimension $d$, boolean estimateLatentDimension, MLC distance $\tau$, number of iterations $s$, indices $\nu$ of dimension estimation examples}
$\boldsymbol{w} \leftarrow $ zero list of length $|F|$ \Comment*[r]{Initiate importances.}
\If{estimateLatentDimension}{
    $d \leftarrow$ latentDimension$(\textbf{F},\nu)\quad(\in \mathbb{N}_{+})$\;
}
{$\textbf{E} \leftarrow $ manifoldProjection$(|\textbf{F}|$,$d$,$\nu)\quad\left(\in \mathbb{R}^{|I| \times d}\right)$}\;
\For{ $i$ in 1 $\dots$ $s$}{
    $\boldsymbol{r}_i \leftarrow$ randomInstance($\textbf{F}$)\Comment*[r]{Sample instance.}\label{line:sample}
\If{MCC}{
\For{$c \in C$}{\label{line:for-c}
    dists $\leftarrow$ computeDistances($\boldsymbol{r}_i$,$\delta$,$\textbf{E}$,$c$)\;
    sortedIndices $ \leftarrow $ argSortDistances(dists)\;
    \If{adaptiveThreshold}{
    $k \leftarrow$ adaptiveThresholdMethod(sortedIndices, dists)\;
    }
    nearestNeighbors $ \leftarrow $ select(sortedIndices,$k$)\;
    
    \For{$j \textrm{ in } 1 \dots |F|$}{
        priorWeight $\leftarrow$ ComputeWeight($c$, dists, $k$)\;
        $\boldsymbol{w}[j] \leftarrow \boldsymbol{w}[j]$ +  updateScore(priorWeight, nearestNeighbors, j)\;\label{alg:line:cc-update}
    }
   
    }
    }\ElseIf{MLC}{
    dists $\leftarrow$ computeDistances($\boldsymbol{r}_i$,$\delta$,$\textbf{E}$)\;\label{line:distances}
    sortedIndices $ \leftarrow $ argSortDistances(dists)\;
    \If{adaptiveThreshold}{
    $k \leftarrow$ adaptiveThresholdMethod(sortedIndices, dists)\;
    }
    nearestNeighbors $ \leftarrow $ sortedIndices[:$k$]\;  
    targetDistances $\leftarrow$ \{\}\Comment*[r]{Note the direct use of $\textbf{T}$ compared to MCC.}
    descriptiveDistances $\leftarrow$ \{\}\;
    \For{neighborIndex $\in$ nearestNeighbors}{
        distTar $\leftarrow$ distanceToTarget($\boldsymbol{T}[i]$, $\boldsymbol{T}$[neighbor], $\tau$)\;\label{alg:line:target-dist}
        distDes $\leftarrow$ distanceToDesc($\boldsymbol{E}[i]$, $\boldsymbol{E}$[neighbor], $\delta$)\;
        targetDistances.add(distTar)\;
        descriptiveDistances.add(distDes)\;
    }
    meanTarDist $\leftarrow$ mean(targetDistances)\;
    meanDesDist $\leftarrow$ mean(descriptiveDistances)\;
    \textsc{td-diff}, \textsc{d-diff}, \textsc{t-diff} $\leftarrow$ expectedDist(meanTarDist, meanDesDist)\;
     \For{$j \textrm{ in } 1 \dots |F|$}{
        $\boldsymbol{w} \leftarrow \boldsymbol{w} +  \frac{\textsc{td-diff}}{\textsc{t-diff}} - \frac{\textsc{d-diff}- \textsc{td-diff}}{1 - \textsc{t-diff}}.$ \Comment*[r]{See Eq.~\ref{mlc:update}.}
     }
    }
}
\KwRet $\boldsymbol{w}$\;
\end{algorithm}
The adaptive threshold step can be further formalized as shown in Algorithm~\ref{algo:adaptiveThreshold}.
\begin{algorithm}
\caption{adaptiveThresholdMethod.}
\label{algo:adaptiveThreshold}
\DontPrintSemicolon
\KwData{distances, sortedIndices}
sortedDist $\leftarrow$ reorderAscending(dists, sortedIndices)\;
{sortedDiff[i]$\leftarrow$ sortedDist$[i+1]-$sortedDist$[i]\quad\left(\in \mathbb{R}_{+}^{|F|-1}\right)$}\;
$k \leftarrow \argmax{\textrm{sortedDiff}}$ \Comment*[r]{Peak point.}
\KwRet $k$\;
\end{algorithm}
\subsubsection{Multi-class classification}
We first discuss the part of ReliefE algorithm that handles multi-class classification (MCC) tasks. Here, the classes are traversed (line \ref{line:for-c}) as follows.
If \textit{adaptiveThreshold} is enabled, the number of neighbors to be considered is determined dynamically for each sample (see Section~\ref{sec:adaptive} for more details). Next, the neighbors are selected and used for the final weight update, where the prior probabilities of individual classes are also considered. The \textit{updateScore} method iteratively updates the weights (line \ref{alg:line:cc-update}), and is in this work for the $j$-th feature and $i$-th sample defined as follows:
\begin{equation*}
    w[j] \textrm{+=} \underbrace{\Big| \boldsymbol{\boldsymbol{r}}_i^j - \mathbb{E}[\textrm{nearestNeighbors}(i)][j] \Big|}_{\textrm{ absMean weight update}}  
    \cdot  \underbrace{\begin{cases}
    P[c] / (1 - P[c_i])  &;\; c \neq c_i \\
    -1 &;\; c = c_i
    \end{cases}}_{\textrm{Prior information}}.
\end{equation*}
\noindent In the proposed update step, an instance is compared directly to {the} mean of its neighbors which reduces the noise compared to 
the original updates of Relief \cite{kira1992feature} where the order of the averaging and $|\cdot|$ operators is reversed.

The nearestNeighbors represents the ordered set of indices of the top $k$ neighbors, thus $\mathbb{E}[\textrm{nearestNeighbors}(i)][j]$ represents the first moment w.r.t. the $j$-th feature based on the nearest neighbors of the $i$-th instance (there are $k$ such neighbors).
We set the prior to -1 and the offset for considering the nearest neighbors to +1 if $c = c_i$, i.e., the considered class $c_i$ is equal to currently considered class $c$.

\subsubsection{Multi-label classification}
For multi-label classification (MLC option), the weight update step differs substantially from the MCC case, after selecting a random instance ($\boldsymbol{r}_i$) and determining its distance to the other examples. First, the indices of the closest $k$ neighbors are stored in nearestNeighbors. As the values in the target space $\boldsymbol{T}$ are sets of (multiple) labels per instance, the simple iteration considered in the MCC case does not take {the} interactions between classes into account (label co-occurrences). Hence, distances are also computed between the target values of $\boldsymbol{r}_i$ and its nearest neighbors (line \ref{alg:line:target-dist}), by using one of the implemented options of ReliefE, which are given in Table~\ref{tbl:mlc1}.
\begin{table}[]
    \centering
     \caption{Considered distances between rows in the multi-label output matrix $\boldsymbol{T}$. The $\boldsymbol{t}_1$ and $\boldsymbol{t}_2$ correspond to two rows, the nnz represents the count of non-zero elements in a given row vector. Note that {the} considered vectors are binary in all but the cosine and hyperbolic cases (last two rows).}
     %{vsote so vecinoma odvec, sm pobrisal in ppravu druge napake}}
    \begin{tabular}{rl}
    \hline
    Distance & Definition \\ \hline
      F1   &  $\textrm{dist} = \begin{cases} 1-
       \frac{2 \boldsymbol{t}_1 \boldsymbol{t}_2^{T}}{(\textrm{nnz}(\boldsymbol{t}_1) + \textrm{nnz}(\boldsymbol{t}_2))} &; (\textrm{nnz}(\boldsymbol{t}_1) + \textrm{nnz}(\boldsymbol{t}_2)) > 0\\ \vspace{0.1cm}
       0 &; \textrm{ otherwise}
      \end{cases}$ \\ 
      \hline
        Accuracy & $\textrm{dist} = \begin{cases}
        1 - \frac{\boldsymbol{t}_1  \boldsymbol{t}_2^{T}}{(\textrm{nnz}(\boldsymbol{t}_1 + \boldsymbol{t}_2))} &; (\textrm{nnz}(\boldsymbol{t}_1 + \boldsymbol{t}_2)) > 0 \\ \vspace{0.1cm}
        0 &; \textrm{ otherwise}
        \end{cases}$\\ 
      \hline
        Subset & $\textrm{dist} = \begin{cases}
        1 &;\;\boldsymbol{t}_1 == \boldsymbol{t}_2; \\ \vspace{0.1cm}
        0 &; \textrm{ otherwise}
        \end{cases}$\\
      \hline
        Hamming & $ \textrm{dist} = \sum_{i = 1}^{|\bm{t}_1|} |\boldsymbol{t}_{1, i} - \boldsymbol{t}_{2, i}| / |\boldsymbol{t}_1|$\\ 
      \hline
        Cosine (if embedded) & $\textrm{dist} = \boldsymbol{t}_1  \boldsymbol{t}_2^{T} / (\norm{\boldsymbol{t}_1}_2 \norm{\boldsymbol{t}_2}_2)$\\ 
      \hline
        Hyperbolic (if embedded) & dist $= \arccosh{}(-\bm{t}_1 \bm{t}_2^T)$ %\arccosh(-\sum(\boldsymbol{t}_1 \odot \boldsymbol{t}_2))$ {bolje/preprosteje \arccosh{}(-\bm{t}_1 \bm{t}_2^T)$}
        \\
        \hline
    \end{tabular}
    \label{tbl:mlc1}
\end{table}
Note that we also consider the cosine and hyperbolic distances which are applicable if the label space is embedded prior to the ranking step. We believe employment of manifold projections that operate on sparse spaces can be of relevance for high-dimensional output spaces, as for example seen when considering gene function prediction \cite{urbanowicz2018benchmarking}.
Once the distances are obtained both based on the input space and the output space, this information is used for updating feature weights as follows. 
Let $K$ represent the set of considered nearest neighbors. 
Let \textsc{t-diff} represent the mean of the target distances \textsc{tar-diff}. Let \textsc{d-diff} represent the mean of the (descriptive) distances to neighbors (\textsc{des-diff}), as also considered for the MLC case, i.e. the absolute difference between the selected instance $\boldsymbol{r}_i$ and the mean of the nearest neighbors. More formally
\begin{equation*}
    \textsc{t-diff} = \mathbb{E}[\textrm{dist}(\boldsymbol{t}_i, \boldsymbol{T}[\bm{n} \in K])] \quad \textrm{ and } \quad \textsc{d-diff} = \mathbb{E}[\textrm{dist}(\boldsymbol{r}_i, \boldsymbol{E}[\bm{n} \in K])],
\end{equation*}
where $\bm{X}[\bm{n} \in K]$ keeps only the rows of the matrix $\bm{X}$ that correspond to the neighbors $\bm{n} \in K$.
\noindent We further define
\begin{equation*}
    \textsc{td-diff} =
    \mathbb{E}[\textsc{tar-diff} \odot \textsc{des-diff}],
\end{equation*}
 \noindent and the weight update can be defined as:
\begin{equation}
    w[j] \textsc{+=} \frac{\textsc{td-diff}}{\textsc{t-diff}} - \frac{\textsc{d-diff}- \textsc{t-diff}}{1 - \textsc{td-diff}}.
    \label{mlc:update}
\end{equation}
\noindent The weight update concludes the ranking for multi-label classification.

\subsection{Theoretical analysis}
\label{sec:theoretical}

We next discuss the relevant theoretical aspects of ReliefE, ranging from computational complexity  analysis (both time and space) to the implications of the adaptations considered.

\subsubsection{Time complexity}
The time complexity of ReliefE can be studied with respect to the two main modes of function -- multi-class and multi-label classification.
\begin{description}
\item \textbf{Dimensionality estimation}. The dimensionality estimation step, as optionally considered in this work, requires pairwise distance computation between the instances. Thus, $\Theta(|\nu|^2 \cdot |F|)$ operations are required, where $\nu$ represents the indices of the samples for dimension estimation, and $|\nu|< |I|$, as discussed in Section~\ref{sec:subspaces}. Note that if all instances are considered, the complexity rises to $\Theta(|I|^2 \cdot |F|)$.
\item \textbf{Manifold projections}. Learning low-dimensional representations represents one of the computationally more intensive parts of ReliefE. Following \cite{mcinnes2018umap-software}, the UMAP's complexity can be split into two main parts. First, approximate nearest neighbor computation was shown to have empirical complexity of $\mathcal{O}(|I|^{1.12} \cdot |F|)$ \cite{dong2011efficient}. However, in the sampling limit, if all instances are considered, the computational complexity is equal to that of pairwise comparisons -- $\mathcal{O}(|I|^{2} \cdot |F|)$
The optimization of embeddings requires additional  $\mathcal{O}(k \cdot |I|)$ steps, where $k$ is the number of nearest neighbors (a hyperparameter). Overall complexity is in the worst case thus $\mathcal{O}(|I|^{2} \cdot |F|)$. Note that the proposed cyclic sampling scheme (Section~\ref{sec:subspaces}) implies $I \rightarrow \nu$ for all cases in this paragraph.
\item \textbf{Multi-class}.
Given a fixed number of samples $s$, ReliefE traverses each of the classes $|T|$, and for each one performs the sampling. The adaptive neighbor selection scheme does not cost any additional time w.r.t. $|I|$, as the distances are already computed. The feature update step requires $\mathcal{O}(|F|)$ operations, for each neighbor. The complexity of the original, re-implemented ReliefF is thus $\mathcal{O}(|I| \cdot |F| \cdot s)$  \cite{robnik2003theoretical}. The absMean update does not change this complexity, however, when adaptive scoring is considered, distances to the class members need to be sorted. We re-use the sorted indices of top neighbors to obtain closest distances, thus no additional time is spent on sorting. If ReliefE is considered, the complexity needs to be adapted for input dimension estimation, as well as lower dimension in which distances are computed. The final complexity is thus: $\mathcal{O}(|\nu|^2 \cdot|F| + |I| \cdot d \cdot s)$, where $d$ is the dimensionality of the embedding. Assuming the ``empirical complexity" from the previous paragraph holds, the multi-class complexity can also be stated as $\mathcal{O}(|\nu|^{1.12} \cdot |F| + |I| \cdot d \cdot s)$
\item \textbf{Multi-label}.
The complexity of multi-label classification needs to additionally account for the distances computed between the target instances. Effectively, {$\mathcal{O}(|\nu|^2 |F| + s\cdot(|I| \cdot d_{F} +  k \cdot d_{T}))$} operations are required, where $d_T$ and $d_F$ correspond to embedding dimensions of the input and output space -- for each sample, first distances need to be computed between the instances to that sample within the input space ($|I|$). Once top $k$ nearest instances are identified, the distances of the target instance to these $k$ other target instances are computed in $d_T$ space.
\item \textbf{Down-stream ranking}. Commonly, Relief algorithms operate in the raw feature space, however, as ReliefE operates via embedding-based distance computation, we consider the option that embeddings are \emph{pre-computed}. This is possible due the fact that many contemporary embedding algorithms \emph{refine} the representation, once the new data is obtained, and do not (necessarily) re-compute the embedding for each new instance. In this case, the initial complexity of $\mathcal{O}(|\nu|^2 \cdot |F| + |I| \cdot d \cdot s)$ reduces to $\mathcal{O}(|I| \cdot d \cdot s)$.
\end{description}

\subsubsection{Space complexity}
The proposed implementation of ReliefE in comparison with, e.g., state-of-the-art Python-based implementations (as found in \cite{urbanowicz2018benchmarking}) operates easily in very high-dimensional, sparse vector spaces. In practice, we adopt the CSR formalism for matrix representation. Here, a sparse matrix is stored as three arrays, a data array, a pointer array and an index array. All three code for non-zero entries in the input space. Note that such representation is not optimal for dense matrices, as it results in some (minor) space overhead. This design decision means that every computationally-intensive operation that is part of ReliefE was re-written with handcrafted CSR-friendly, Numba-compilable methods. More formally, let nnz correspond to the number of non-zero elements in a given matrix. The space complexity of ReliefE can thus be stated as:
$\mathcal{O}( \textrm{max}(\textrm{nnz}(\boldsymbol{F}),\textrm{nnz}(\boldsymbol{E_F})) + \textrm{max}(\textrm{nnz}(\boldsymbol{T}), \textrm{nnz}(\boldsymbol{E_T})))$. 

The complexity thus depends on the relation between the embedded space and the input space, which can be very context-dependent, however very low-dimensional embeddings normally do not result in space overhead, and neither do highly sparse input matrices. More formally, if
%\begin{equation*}
    $\textrm{nnz}(\boldsymbol{F}) \geq |I| \cdot d \textrm{ or } \textrm{nnz}(\boldsymbol{T}) \geq |I| \cdot d$,
%\end{equation*}
\noindent the embedded space will require less (or equal) memory.
Note that $\boldsymbol{T}$, corresponding to a potentially very sparse output space, is similarly considered as a sparse matrix, meaning that classification problems with very high-dimensional target spaces can also be considered, which is to our knowledge one of the first such Python-based, user-friendly implementations. As dimensionality estimation only requires the two closest neighbors, we do not keep all others in memory, the space complexity becomes linear, i.e., $\mathcal{O}(|I|)$ (in fact, exactly $2 \cdot |I|$). We empirically discovered that UMAP's memory requirements are the main space bottleneck, and, based on the evaluation on the larger data sets, require $\mathcal{O}(|I|^2)$ (empirical) space. 
Such complexity potentially arises from the dense computational graph {derived} by UMAP.
This observation led us to introduce the representative (cyclic) sampling scheme, which reduced this complexity to $\mathcal{O}(|\nu|^2)$, making ReliefE executable even on an off-the-shelf computer (Lenovo Carbon X1). Note that the number of samples is lower-bounded by the number of classes or unique label sets.

\subsubsection{absMean update step and its implications}
Compared to the original ReliefF, one of the proposed modifications implemented in ReliefE is the comparison of a given instance
{directly to the average nearest neighbor. We believe that this approach is advantageous in two ways.}
\begin{figure}[ht!]
    \centering
    \begin{tabular}{cc}
\subcaptionbox{$\bm{r}$ is in the center of its class\label{fig:abs-mean-center}}{\includegraphics[width = .45\linewidth]{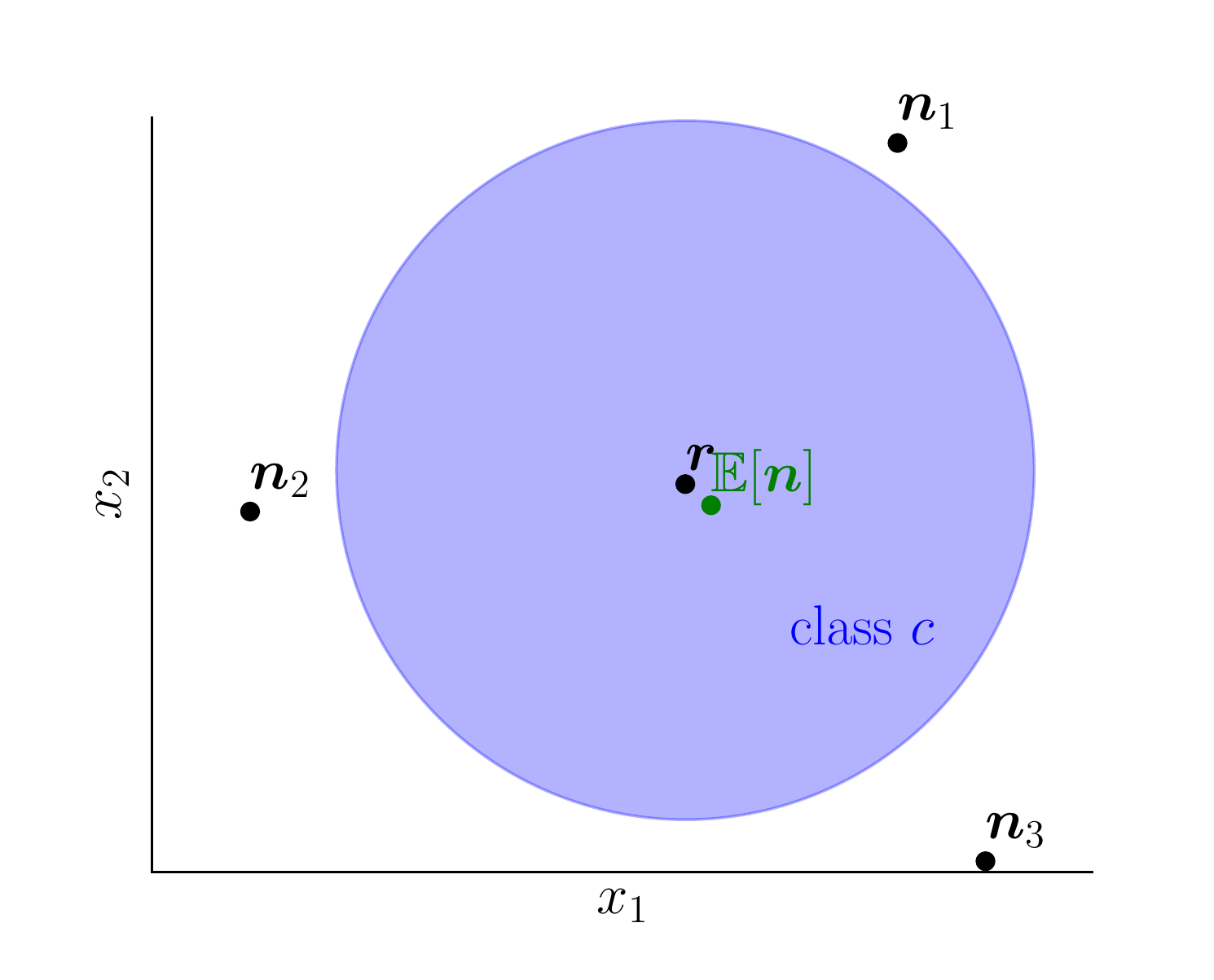}} &
\subcaptionbox{$\bm{r}$ is on the border of its class\label{fig:abs-mean-border}}{\includegraphics[width = .45\linewidth]{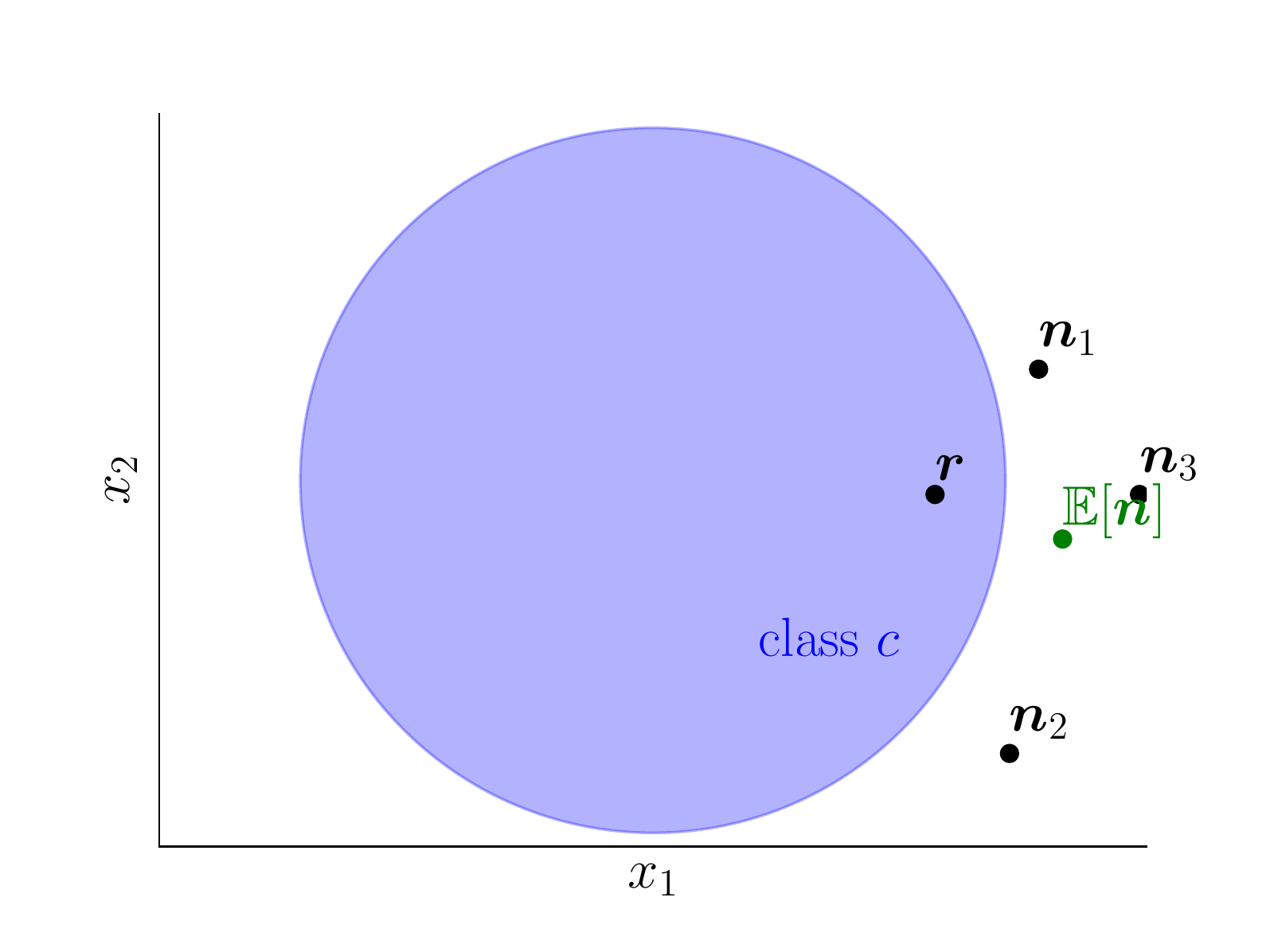}}
\end{tabular}
    \caption{Updating weights with the absMean approach. The $n_{1,2,3}$ represent the instance $\textbf{r}$'s neighbors.}
        \label{fig:abs-mean}
\end{figure}
First, as shown in Fig.~\ref{fig:abs-mean-center}, if a sample $\bm{r}$ that is far away from the class border is chosen, we cannot capture the local structure of the data in the other classes, so such samples $\bm{r}$ should not influence the updates considerably. This is not the case in the standard ReliefF, since the differences in feature values are necessarily large. This is overcome by computing the average neighbor first, and then updating the weights.

Second, when the sample $\bm{r}$ is close to the border (Fig.~\ref{fig:abs-mean-border}), averaging neighbors results in correctly detecting that the general direction
of the neighbors should be perpendicular to the class borders when the number of samples goes to infinity.
For example, in the situation depicted in Fig.~\ref{fig:abs-mean-border}, only $n_1$ should be rewarded. Again, computing the mean neighbor $\mathbb{E}(\bm{n})$ first, brings us closer to the optimal direction.
The reduction of noise can be also proven by using the triangle inequality, $\frac{1}{k}\sum_{j = 1}^k |n_i^0 - n_i^j|\geq |n_i^0 - \frac{1}{k}\sum_{j = 1}^k n_i^j|,$ from which it directly follows that this approach results in smaller weight updates.

\subsubsection{Adaptive neighbor selection and its behavior}
\begin{figure}[h!]
    \centering
    \begin{tabular}{cc}
\subcaptionbox{Dispersed neighborhood\label{fig:adaptive1}}{\includegraphics[width = .30\linewidth]{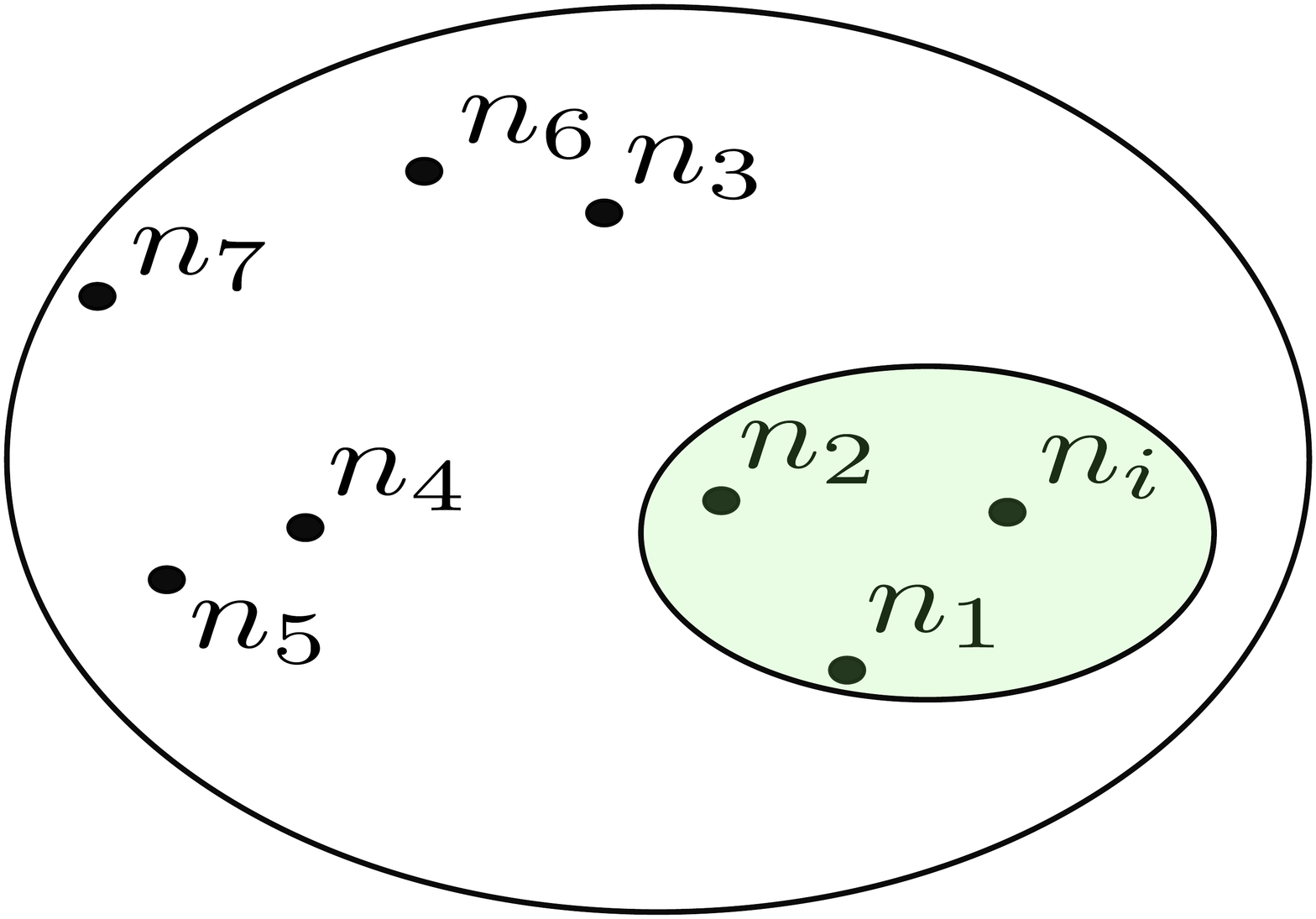}} &
\subcaptionbox{Local neighborhood\label{fig:adaptive2}}{\includegraphics[width = .30\linewidth]{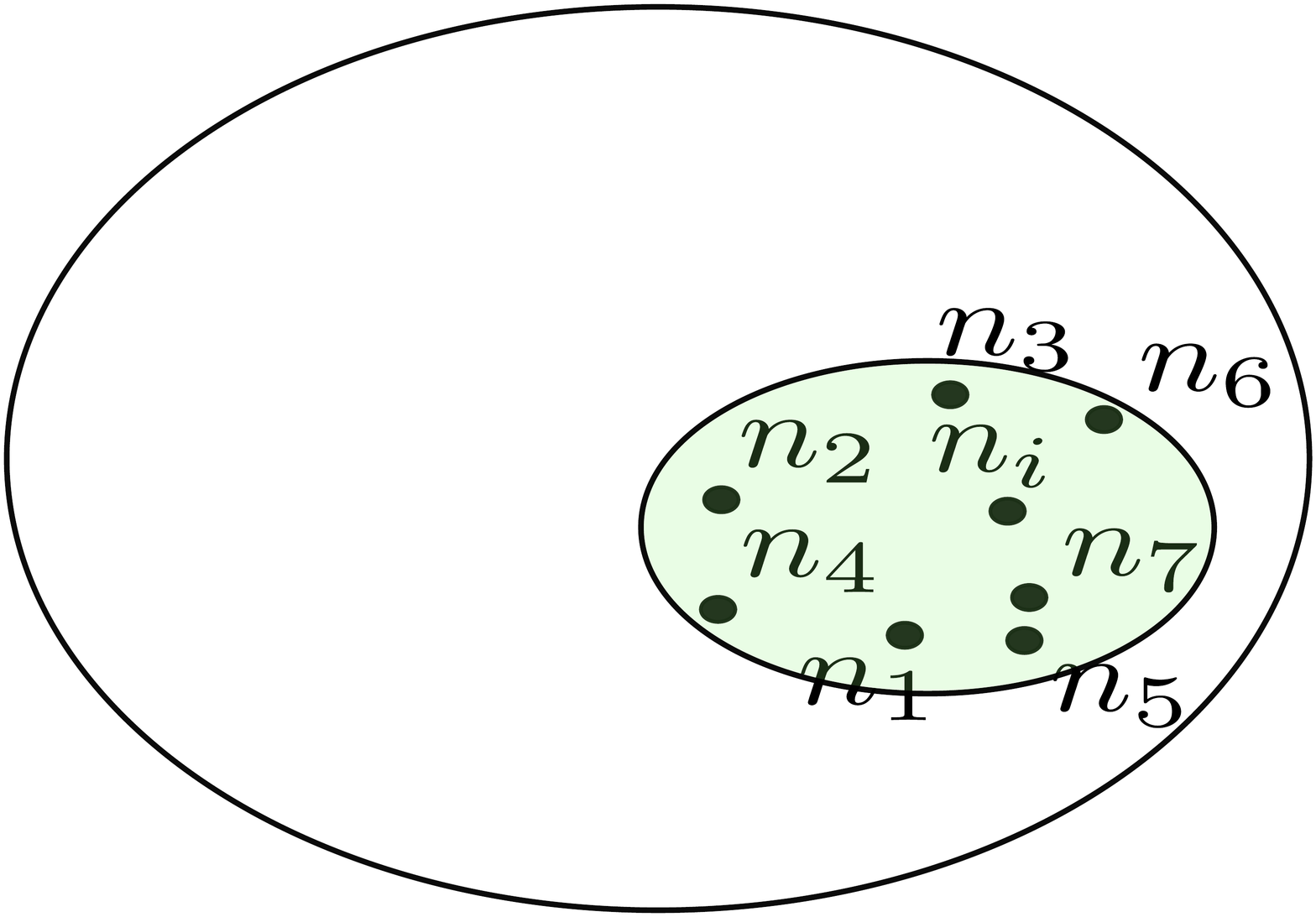}}
\end{tabular}
    \caption{Adaptive neighbor selection. The $n_{1,2,3,\dots}$ represent neighbors.}
        \label{fig:adaptive}
\end{figure}

The considered adaptive neighbor selection attempts to reduce the number of hyperparameters by one ($k$), potentially saving $\mathcal{O}(k)$ optimization iterations, should this parameter be tuned. Furthermore, by considering neighbors, potentially relevant for a given instance, less noise is considered during the weight update step. For example, assume $k=7$, with only three other instances very close and the remaining four much further (by a large margin).  The latter 4 instances will impact the weight update significantly, as the average distance will be heavily biased towards their mean, and thus potentially not representative of the close neighborhood of a given instance that naturally appears in the data. A visualization in such a situation is shown in Figure~\ref{fig:adaptive}.

In both panels ((a) and (b)), the outer circle represents the neighbourhood for a hard-coded value of $k$. In Figure~\ref{fig:adaptive1}, very distant instances are also considered for the update (e.g., from $n_3$ onward) and the adaptive estimation only selects the closest neighbors (green). However, in Figure~\ref{fig:adaptive2}, all instances are very close, thus the value of $k$ is equal to the automatically selected choice. 

This follows the intuition behind the Relief family of algorithms, where an instance is compared to its slight perturbations. Another downside of having $k$ fixed,
is that taking into account more distant nearest neighbors would (on average),
increase the importance of more noisy features, since the distance values directly influence the importance.
Irregularities in distance distribution were shown to hold for many real-life data sets, see for example the assumptions and their implementation in \cite{dong2011efficient}.
Finally, as ReliefE operates in a latent, low-dimensional space, obtained by instance similarity comparison, comparison to the closest instances only is potentially meaningful.
%Hence, hard-coding the number of neighbors potentially results in comparison to semantically different neighborhoods, which defeats the purpose of using embedding-based distance computations to some extent.

\subsubsection{Parallelism aspects}
The proposed implementation exploits the Numba framework for just-in-time compilation \cite{lam2015numba}. Numba offers parallelism at the level of individual methods that get compiled, meaning that the proposed implementation offers parallelism at the level of weight updates. During compilation, parts of the code that are sensible to compile get detected \emph{automatically}. 
Many operations such as scalar-vector addition and similar can easily be parallelized. 
With auto-parallelization, Numba attempts to identify such operations in the ReliefE weight update step, and fuse adjacent ones together, to form one or more kernels that are automatically run in parallel. In practice, however, we observed that such auto-parallelism does not necessarily offer superior performance in terms of speed. However, it represents an elegant, array-level parallelism detection which, when improved/updated, shall speed up the execution time even more. We omit the discussion regarding different spaces considered during ReliefE to Appendix~\ref{appendix:theory}.

\subsubsection{How powerful is ReliefE?}
Throughout this paper, we propose and demonstrate the utility of ReliefE when tabular data is considered. However, as ReliefE requires merely the \emph{representations} of instances (training or target), the proposed approach generalizes well beyond tabular data with a single adaptation: the embedding method needs to be suitable for the considered data type. For example, if an instance is described by an ordered list of graphs, the plethora of graph embedding methods \cite{goyal2018graph,9265235} could be used to prioritize the graphs based on their (learned) representations.
Similarly, ReliefE could be adapted for learning in the context of relational data bases, via Wordification \cite{perovvsek2015wordification} and other propositionalization-like algorithms.

\section{Empirical evaluation setting}
\label{sec:experimental}
Our empirical evaluation of ReliefE consists of many sub-studies, and can be summarized as follows. First, we discuss the evaluation of ReliefE against state-of-the-art ranking algorithms on eight multi-class classification data sets. Next, we present the empirical {evaluation} setup where ReliefE's capabilities are shown on nine multi-label classification data sets. Finally, we conducted a series of experiments where we investigated in more detail the convergence and time performance. We conclude this section by describing the Bayesian and frequentist approaches, which aided understanding of the results.

\subsection{Multi-class classification data sets}
\label{sec:cc-empirical}
Multi-class classification remains one of the most widely adopted forms of learning. Here, the input space is associated with a single, integer-valued vector, where each instance can belong to one of the many possible classes. In this work, we consider a wide spectrum of data sets, summarized in Table~\ref{tbl:cc}.
\begin{table}[b!]
\centering
\caption{The properties of the considered multi-class classification data sets. The last column denotes the proportion of non-zero elements in the data table.}
\begin{tabular}{lrrrr}
\toprule
           Data set &    Instances &  Features &  Classes &  Proportion of non-zero entries \\
\midrule
              chess \cite{shapiro1984role} &    3196 &        38 &        2 &  0.726558 \\
 biodeg-p2-discrete \cite{dvzeroski1999experiments} &     328 &        61 &        4 &  0.107357 \\
          optdigits \cite{alpaydin1998cascading} &    5620 &        62 &       10 &  0.528117 \\
            madelon \cite{guyon2005result} &    2000 &       500 &        2 &  0.999999 \\
          php88ZB4Q \cite{php} &   10299 &       561 &        6 &  0.999860 \\
pd\_speech\_features  \cite{sakar2013collection} &     756 &       753 &        2 &  0.995294 \\
              dlbcl \cite{armstrong2002mll} &      77 &      7070 &        2 &  0.997388 \\
           tumors C \cite{tumors-c} &      60 &      7129 &        2 &  0.995722 \\
    AP\_Ovary\_Kidney \cite{stiglic2010stability} &     458 &     10935 &        2 &  1.000000 \\
          ohscal.wc \cite{ohscal} &   11162 &     11465 &       10 &  0.005270 \\
              genes \cite{weinstein2013cancer} &     801 &     20531 &        5 &  0.857824 \\
\bottomrule
\end{tabular}
\label{tbl:cc}
\end{table}
The data sets are from multiple domains, incl. biological, social and other domains (e.g., chess). The data sets are of different dimensions, in terms of the numbers of rows and also columns.

\subsection{Multi-label classification data sets}
Feature ranking for multi-label classification remains an active research area. Many of the approaches considered in the previous section (multi-class classification) are not able to handle {the} multi-label setting, where a single instance can belong to many classes simultaneously. Such a setting, for example, naturally emerges when gene function prediction is considered---a single gene is associated with many functions and contexts. The considered multi-label data sets are summarized in Table~\ref{tbl:mlc}.

Similarly to the multi-class setting, we selected data sets from various domains to maintain diversity. Note that multiple repetitions of 10 fold cross validation were needed to perform Bayesian comparisons.

\begin{table}[ht!]
    \centering
    \caption{The properties of the considered MLC data sets. The last column denotes the proportion of non-zero elements in the data table.}
\begin{tabular}{lrrrr}
\toprule
       Data set &    Instances &  Features &  Classes &  Proportion of non-zero entries\\
\midrule
      delicious \cite{delicious} &   16105 &      1000 &      983 &  0.500000 \\
           imdb \cite{imdb}&  120919 &      1001 &       28 &  0.019363 \\
        medical \cite{medical} &     978 &      2898 &       45 &  0.500000 \\
         bibtex \cite{bibtex}&    7395 &      3672 &      159 &  0.500000 \\
     Education1 \cite{ueda}&   12030 &     27534 &       33 &  0.004059 \\
        Health1 \cite{ueda}&    9205 &     30605 &       32 &  0.003555 \\
 Entertainment1\cite{ueda} &   12730 &     32001 &       21 &  0.004552 \\
       Science1 \cite{ueda}&    6428 &     37187 &       40 &  0.004659 \\
        Social1 \cite{ueda} &   12111 &     52350 &       39 &  0.002949 \\
\bottomrule
\end{tabular}
    \label{tbl:mlc}
\end{table}

\subsection{Additional experiments and statistical evaluation of results}
{For MCC, logistic regression with its default parameters was used as the learner. The first reason for this choice is the fact that this very learner is commonly used to evaluate the quality of a given data representation (in our case a subset of the feature space), and is known to be sensitive to noisy features. The second reason is computational: With all repetitions required for Bayesian analysis, additional grid search would be out of reach as it could further increase the computational time beyond reasonable capabilities. For the MLC setting, the default random forest parametrization was used, as it has been previously shown to perform competitively/ well in such a setting.}
Throughout the experiments, we set the regularization term of logistic regression (C) to one, the default value \cite{pedregosa2011scikit}. For multi-label classification, we considered the RandomForest classifier with default settings as set in \cite{pedregosa2011scikit}.

As we consider either multi-class or multi-label problems, we compute either relative F1 or micro-averaged relative F1 scores, defined as:
\begin{equation*}
    \textrm{rF1}(f) = \frac{\textrm{F1}_f}{\textrm{F1}_{f = |F|}},
\end{equation*}
{where F1 is the harmonic mean of precision and recall, and $f$ is the number of features.}
The {macro} rF1 is defined in the same fashion. Considering relative performance offers direct insights into how performant a given ranking is with how many top-ranked features. Note that by considering relative performance, it can be directly observed when the feature ranking algorithm identifies a ranking that outperforms the situation where all features are considered -- a reasonable baseline. {We performed ten fold stratified cross-validation ten times, as required for the statistical analysis discussed next.}

In order to summarize the overall performance of a given ranking, we believe taking into account the ranking's quality over all possible values of top $f$ features needs to be considered. Hence, we introduce the area under rF1 (AUrF1), i.e., the integral of rF1 normalized by the number of considered top $f$ rankings (to be more comparative across data sets), where we numerically integrate with the Simpson's method.

Recent criticisms of the frequentist non-parametric comparison of multiple classifiers~\cite{demvsar2006statistical} has given rise to a novel spectrum of Bayesian t-tests, that directly offer insight into a probability space corresponding to the differences in algorithm performance~\cite{benavoli2017time}. In this work we adopt the hierarchical t-test, which is capable of comparing pairs of classifiers.
{The hierarchical Bayesian t-test is used to assess the probability of observing a given difference in performance between a pair of classifiers. As noted by Benavoli et al.~{\cite{benavoli2017time}}, it requires that e.g., ten repetitions of ten fold cross validation need to be considered in order to reliably fit a hierarchical model. The approach attempts to model the probability of observing a given difference in performance between a pair of classifiers, which can be in favor of either of the classifiers or undetermined - practically equivalent (rope region). The plotted results are given in the form of triangular schemes, where each point represents a sample from the posterior distribution. Such samples, when aggregated, directly represent a probability of observing a given state (in this case difference between the classifiers).}
We set the rope region to 5\%---if the difference in quality between two rankings is less than 5\%, they are considered equal. The remaining setting is the same as in the original paper \cite{benavoli2017time}, we compare the top ranking for each fold. For a given pair of ranking algorithms, the pairwise Bayesian tests were performed on the data sets common to both algorithms.
Finally, results of time performance are presented in computation time (in seconds) diagrams with standard deviations. Such a comparison is not necessarily informative/useful when multiple classifiers are simultaneously considered, thus we also offer the results in the form of average rank diagrams~\cite{demvsar2006statistical}. We believe that having both local and global insights into the relations between classifiers, their differences are easier to study, even though looking at the classifier ranks alone can be misleading \cite{benavoli2017time}.

\subsection{Considered implementations and baselines}
We next discuss the implementations considered. For multi-class classification, the considered Relief variants were MultiSURF, MultiSURFstar, ReliefF, all from the scikit-rebate library \cite{urbanowicz2018benchmarking}. We also used RandomForest (RF)-based importances (Genie3) and Mutual information (MI)-based ones~\cite{pedregosa2011scikit}. The multi-class Relief variants that are the original contribution of this work include: ReliefE, ReliefE-absMean, ReliefE-adaptive and ReliefE-absMean-adaptive. The suffix \emph{adaptive} denotes the use of an adaptive threshold and \emph{absMean} the use of absMean update step.

Multi-label classification is not supported (at all) in scikit-rebate \cite{urbanowicz2018benchmarking}, and thus we considered the multi-label variants of ReliefE and ReliefF (re-implemented in this work with Numba) with all of the possible distances given in Table~\ref{tbl:mlc1}. We emphasize that when multi-label distances are considered, only the cosine and hyperbolic distances operate on target space embeddings (the other distances do not). The computation of these distances is also more efficient.
\begin{table}[h!]
    \centering
    \caption{Computational complexity of feature importance estimation. For Relief algorithms, we used $s = |I|$. The $t$ corresponds to the number of trees.}
    \begin{tabular}{lcc}
    \hline 
      Algorithm   &  time complexity & space complexity \\ \hline
        ReliefE &$\mathcal{O}(|\nu|^2 \cdot |F| + |I| \cdot d \cdot s)$ & $\mathcal{O}(|\nu|^2)$ \\
        ReliefF & $\mathcal{O}(|F| \cdot |I|^{2})$ & $\mathcal{O}(|F|)$ \\
        Random Forest & $\mathcal{O}(t \cdot |F| \cdot |I| \cdot \log^2 |I|)$ & $\mathcal{O}(|I| + |F|)$\\
        Mutual Information & $\mathcal{O}(|F| \cdot |I|)$ & $\mathcal{O}(|I|)$ \\ \hline
    \end{tabular}
    \label{tbl:c1}
\end{table}
Note that all versions of ReliefF, implemented or re-implemented in this work\footnote{Implementation's official repository is \url{https://github.com/SkBlaz/reliefe}}, natively operate on sparse spaces, which is on its own a contribution of this work. In terms of sparsification, we set the sparsification threshold to 0.15, meaning that if a matrix's density is higher than 15\%, it is sparsified with the proposed procedure (there are many of such matrices amongst the considered data sets). 
{Detailed results of investigating the ablation of the considered data sets' (induced) sparsities are given in Appendix~}\ref{appendix-sparseness}.
{Similarly, the behavior of the adaptive $k$ statistic was also studied in more detail in Appendix~}\ref{apppendix:adaptive}.
Further, $\nu$ (the sample for intrinsic dimension estimation) was set to 2048. 
{The dimension number was set so that the algorithm runs normally on an off-the-shelf-computer (Lenovo Carbon X1) even for larger data sets.}
Thus, if a given data set consisted of more than 2048 instances, a representative subset of 2048 instances was considered for estimating the intrinsic dimension and consequent embedding. The UMAP's setting is left to its defaults, with the dimension being set to the estimated one\footnote{Extensive evaluation of UMAP's capabilities w.r.t. the proposed implementations is beyond the scope of this paper, and is left for further work.}. The value of $k$ is set to 15 for our implementation for ReliefF, and left at its defaults for the baselines. The time and space complexity of the baselines and ReliefE are summarized in Table~\ref{tbl:c1}.
Note that, even though ReliefF (and its other variants') space complexity is linear w.r.t $|F|$, their implementations, should they not consider the sparse input structure, in fact require $\mathcal{O}(|I| \cdot |F|)$ space (as found, e.g., in \cite{urbanowicz2018benchmarking}).

Finally, the considered experiments for multi-label classification consider both Euclidean embeddings, as well as non-Euclidean ones (Poincar\'e ball).

\section{Results}
\label{sec:results}
{This section presents} the results of the empirical evaluation. We begin by discussing the performance comparisons for the task of multi-class classification. We follow on by discussing the results of the experiments on multi-label classification tasks. Finally, we present additional investigations of ReliefE's behavior.

\subsection{Multi-class classification}
\label{sec:results-cc}
\begin{figure}[ht!]
    \centering
    \begin{tabular}{l}
         \subcaptionbox{\label{fig:allCDCC2}}{\includegraphics[width =1.1\linewidth]{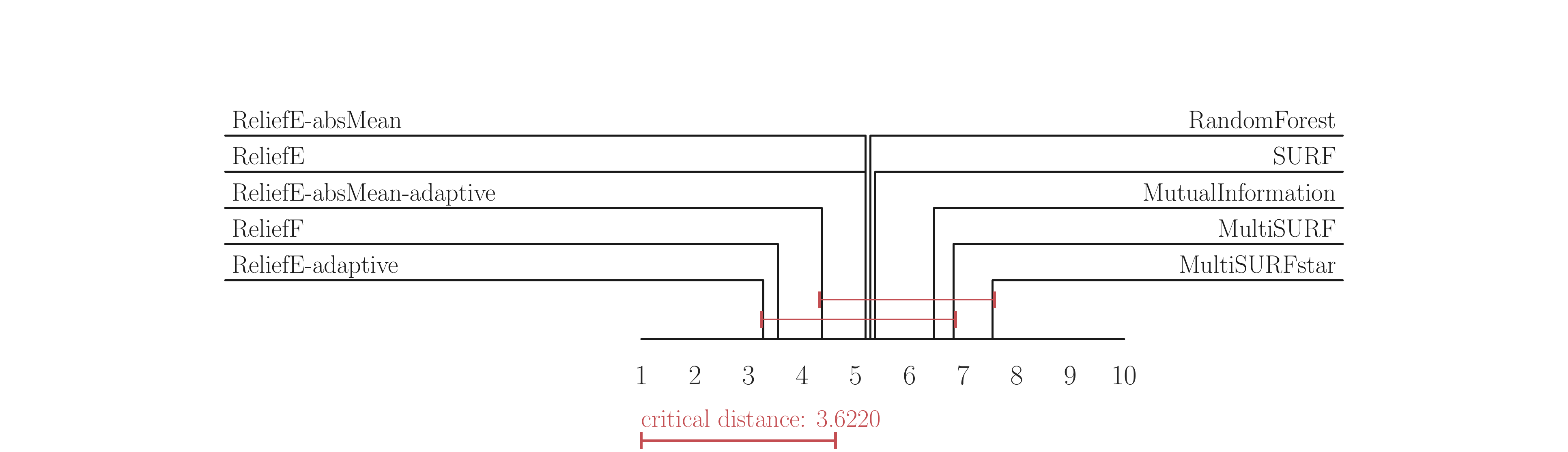}} \\
         \subcaptionbox{\label{fig:allCDCC1}}{\includegraphics[width =1.1\linewidth]{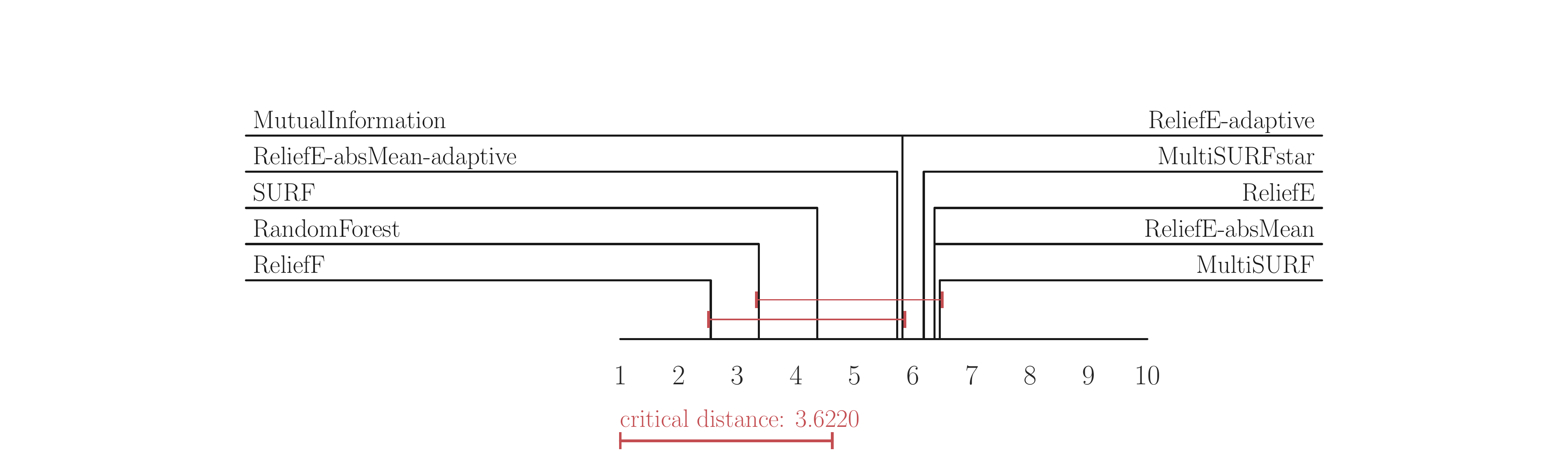}} \\
    \end{tabular}
    \caption{Max (a) and mean (b) F1 scores across all feature rankings.}
\end{figure}

We first present two average rank diagrams depicting the relative performance on the different ranking methods for MCC in terms of the quality of the produced rankings, as measured by the corresponding average and maximum F1 scores (Figures~\ref{fig:allCDCC2} and~\ref{fig:allCDCC1}, respectively). The diagrams  include critical distances, representing the minimum differences in performance that are statistically significant.  It can be observed that the ReliefE variants yield the best performing rankings (with lowest average ranks, Figure~\ref{fig:allCDCC2}), but there are not many such rankings (Figure~\ref{fig:allCDCC1}).
The AUrF1 values (Appendix~\ref{appendix:aurf1}) indicate that the performances of the top 5 feature ranking algorithms are highly similar (within the confidence interval).

\begin{figure}[ht!]
    \centering
    \begin{tabular}{c}
         \subcaptionbox{Average absolute running times with standard deviations (in seconds).\label{fig:timesCC1}}{\includegraphics[width =\linewidth]{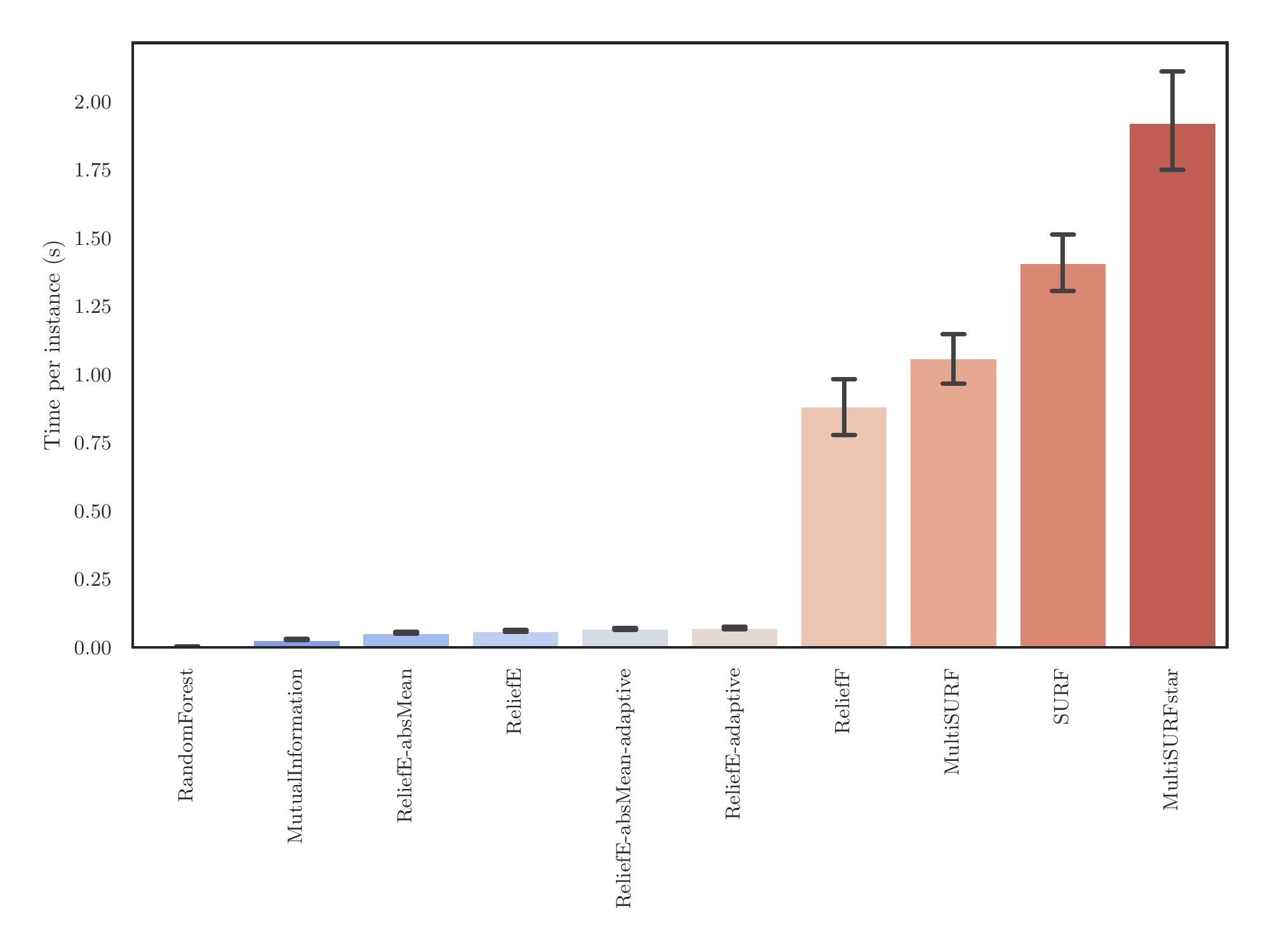}} \\
         \subcaptionbox{Average rank diagram (times).\label{fig:timesCC2}}{\includegraphics[width =\linewidth]{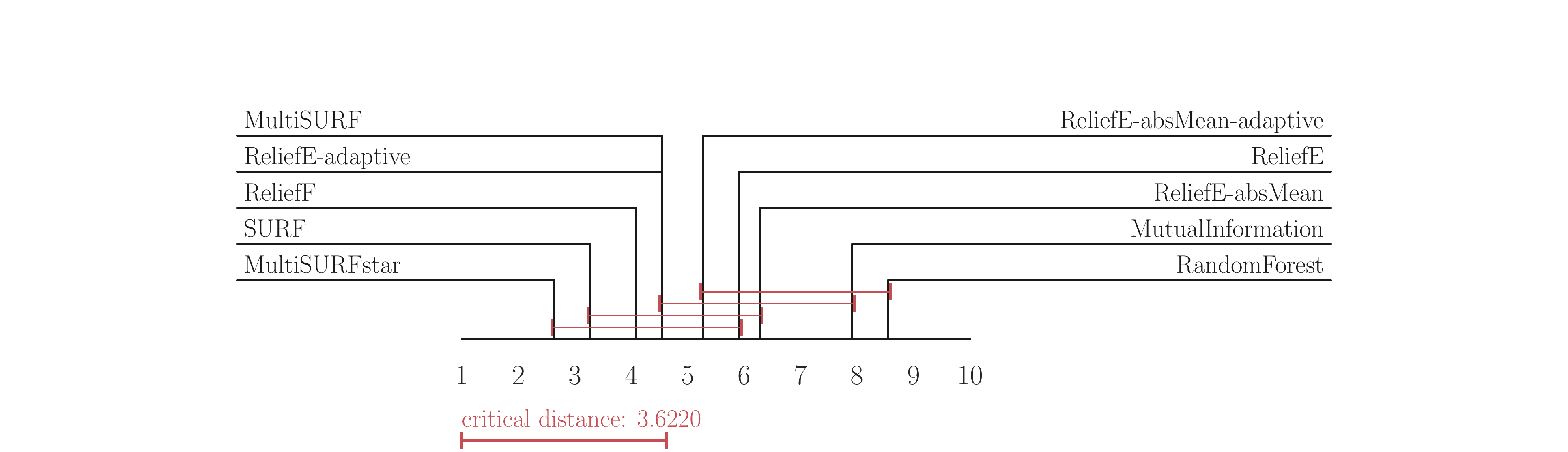}} \\
    \end{tabular}
     \caption{Speed comparison of ranking approaches for MCC. Absolute running times (a) show that ReliefE variants perform an order of magnitude (or more) faster. Relative running times are given in terms of average ranks (b), where lower ranks mean worse performance, i.e., longer running times.}
\end{figure}

We next present the mean time consumption averaged across data sets. Consistently slower SURF variants of ReliefF can be observed in the rightmost part of Figure~\ref{fig:timesCC1}. The average rank diagram is shown in Figure~\ref{fig:timesCC2}.
Additional analysis of the proportions of time spent at different parts of the algorithm is presented in Appendix~\ref{appendix:time}, showing that most time is spent on feature weight updates. Average rank diagrams comparing the rankings in terms of the top 50 and 100 features are given in Appendix~\ref{appendix:additional-rank}. 
%We next present pairwise Bayesian comparisons.

\subsection{Bayesian ranking comparison of ranking approaches for MCC}
\label{appendix-bayesian}
\begin{figure}[h!]
    \centering
\begin{tabular}{cc}
\subcaptionbox{MultiSURFstar vs. MI}{\includegraphics[width = .45\linewidth]{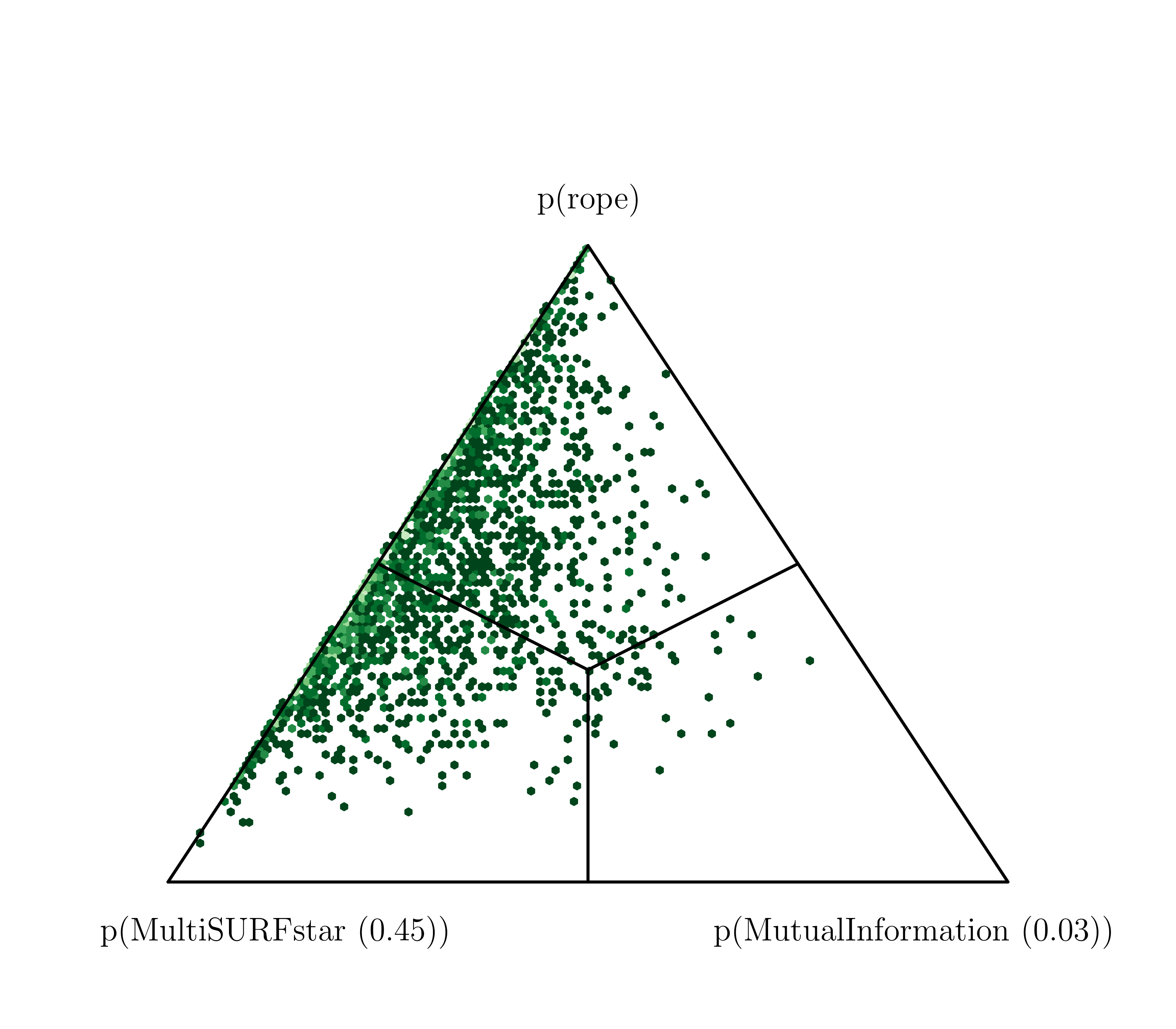}} &
\subcaptionbox{MultiSURFStar vs. RF}{\includegraphics[width = .45\linewidth]{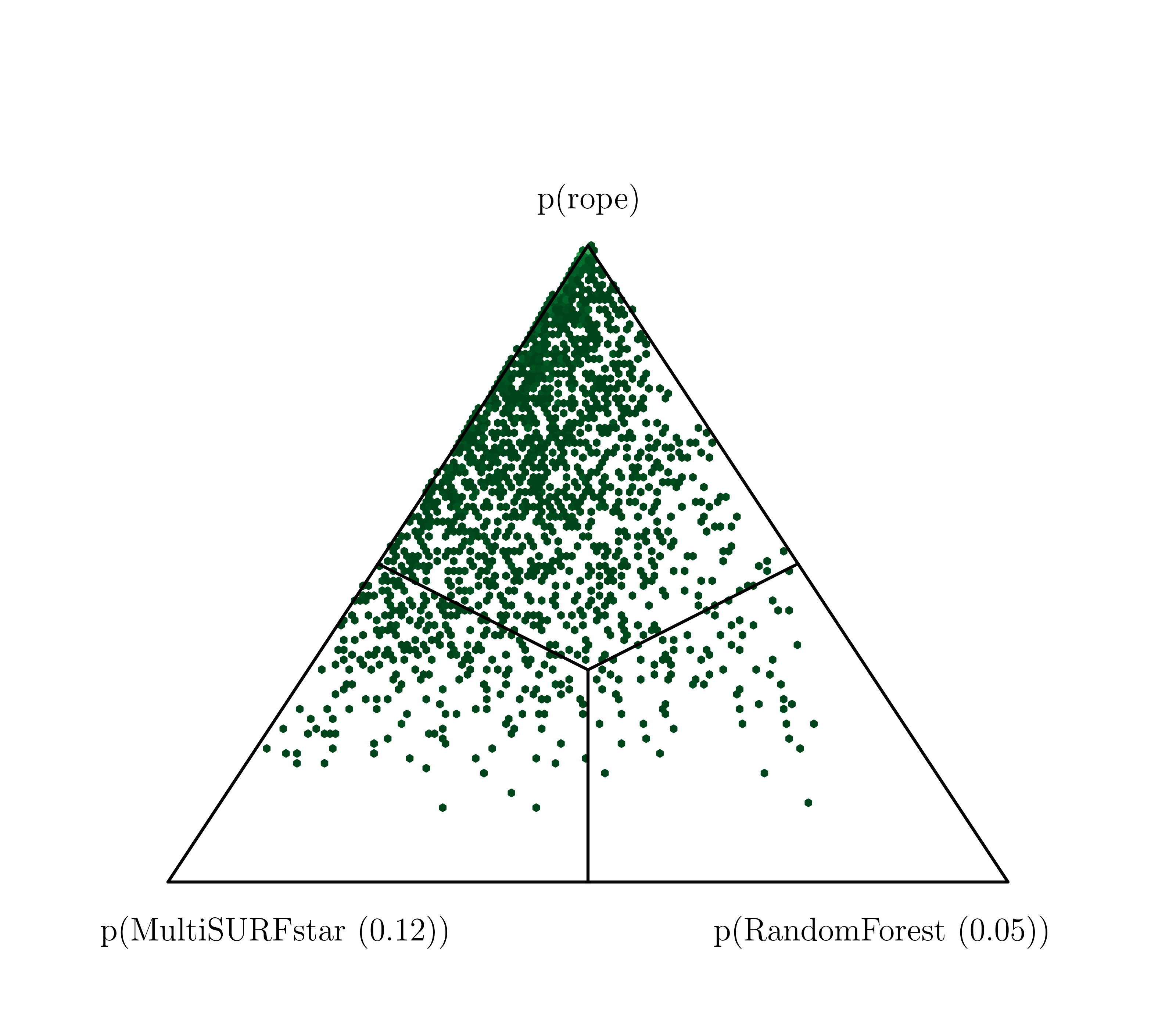}} \\
\subcaptionbox{MultiSURFstar vs. ReliefE-absMean-adaptive}{\includegraphics[width = .45\linewidth]{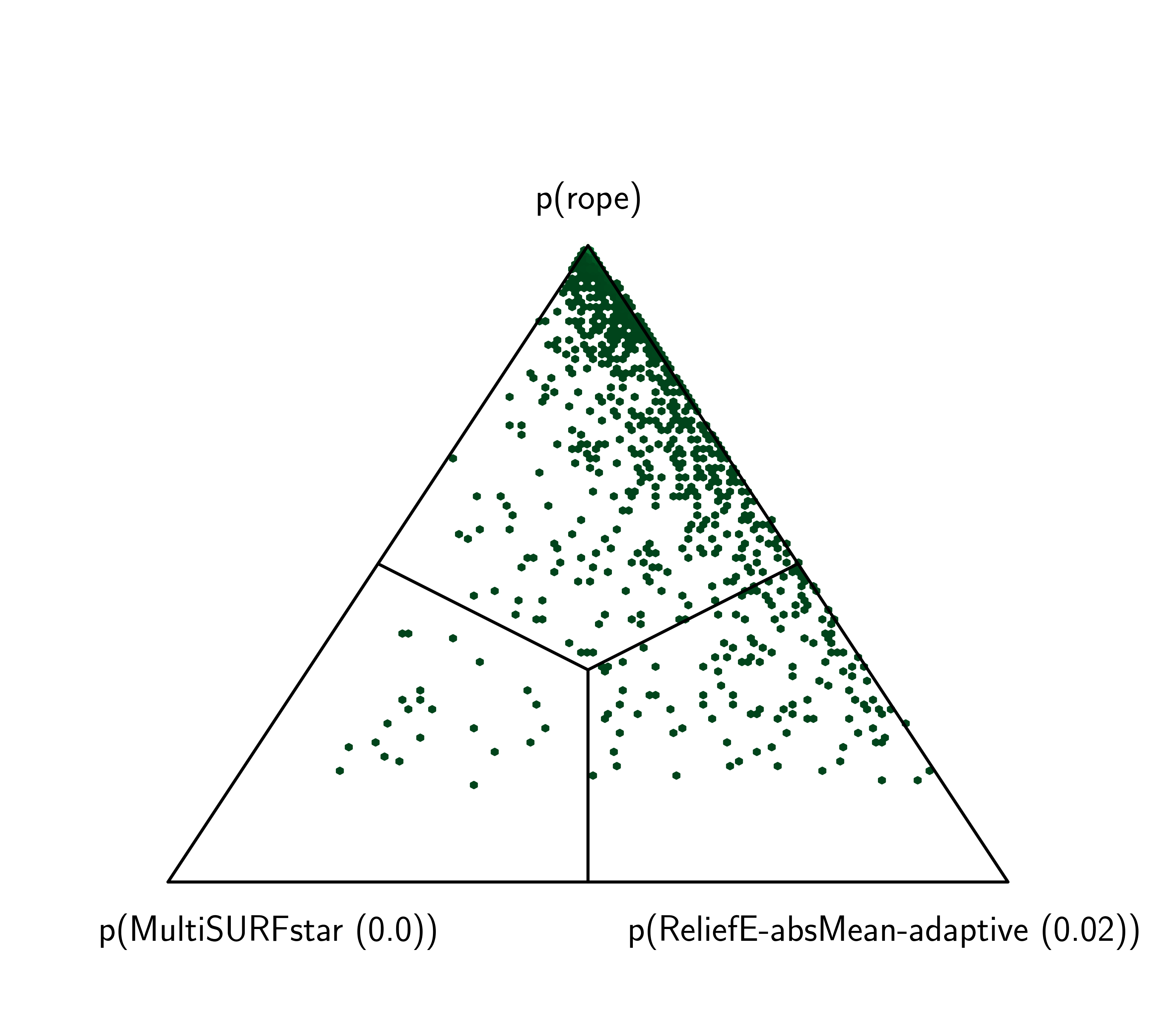}} &
\subcaptionbox{MultiSURFstar vs. ReliefE}{\includegraphics[width = .45\linewidth]{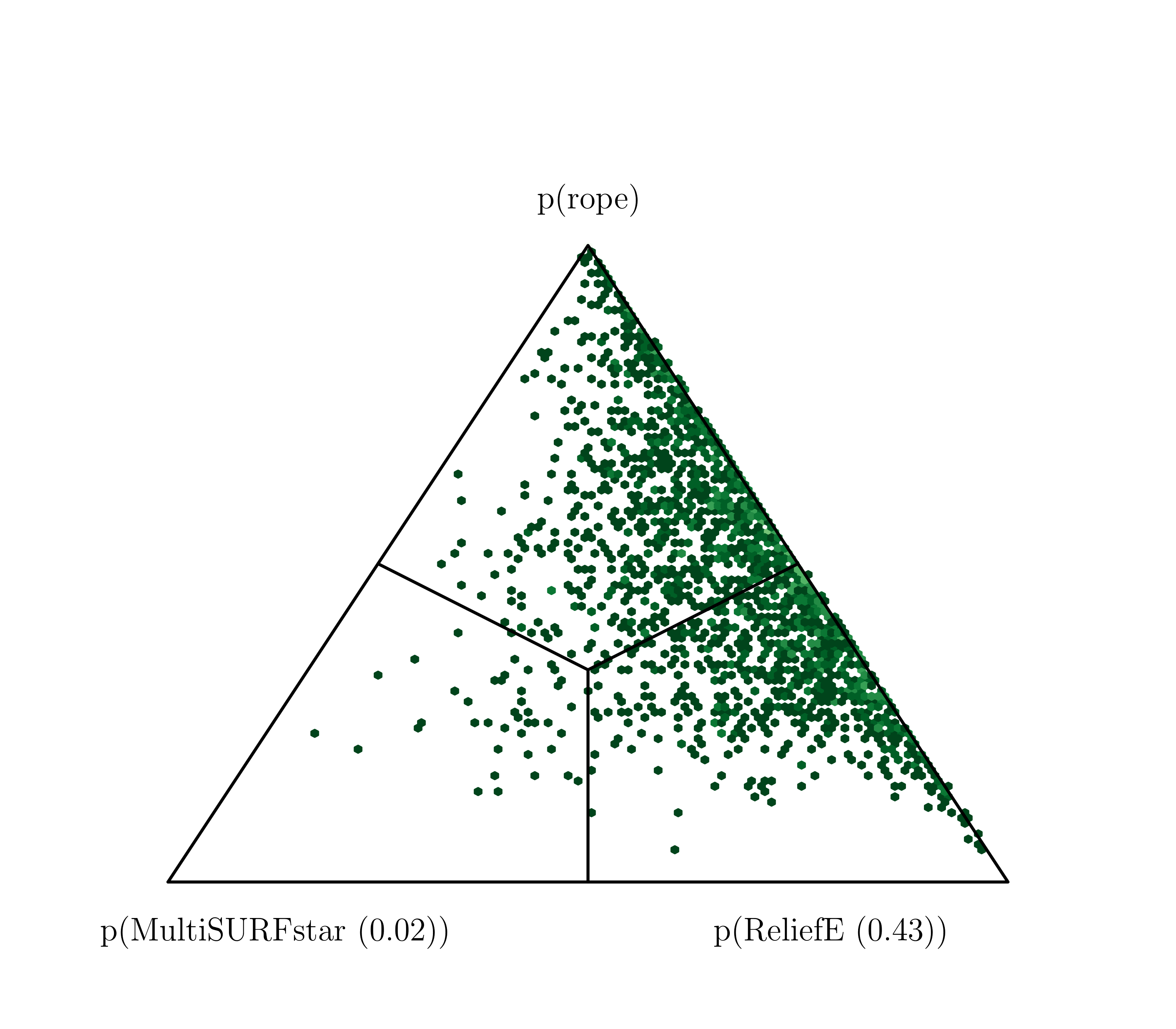}} \\
\subcaptionbox{MultiSURFstar vs. SURF}{\includegraphics[width = .45\linewidth]{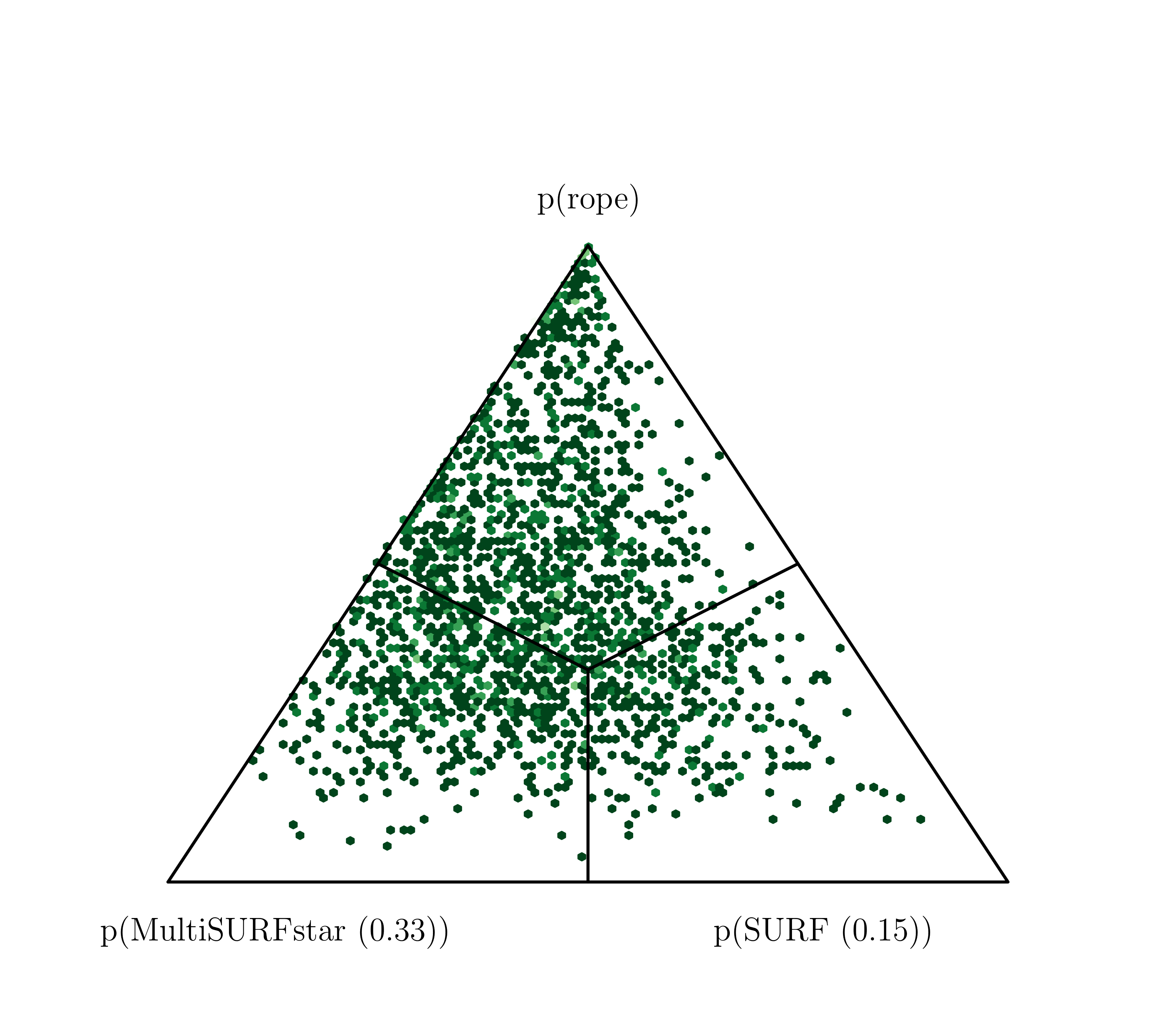}} &
\subcaptionbox{MultiSURFstar vs. ReliefF}{\includegraphics[width = .45\linewidth]{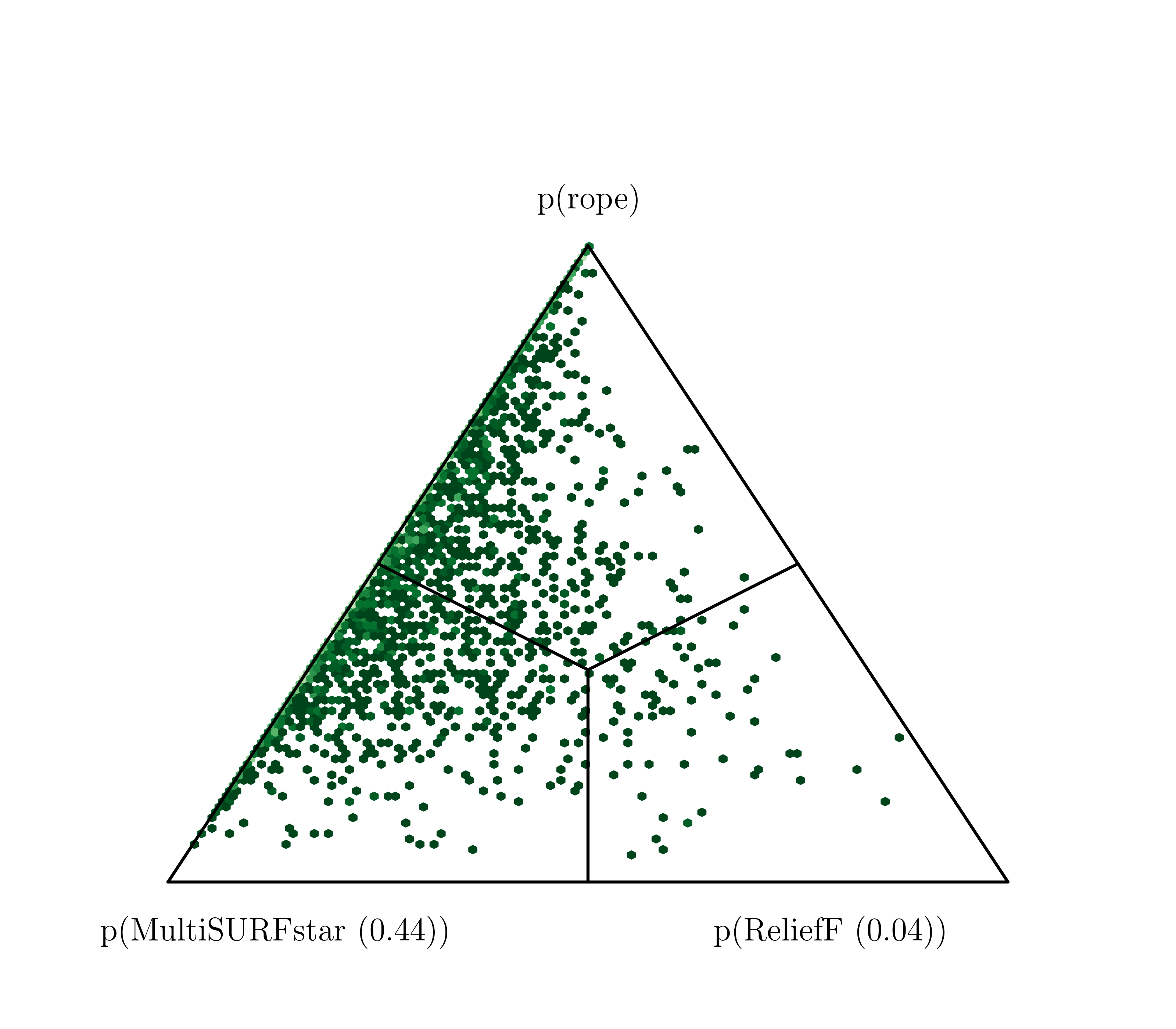}} 
\end{tabular}
    \caption{Bayesian comparisons of performance (ranking qualities) between MultiSURFstar and other feature ranking methods for MCC.}
        \label{fig:bayesians}
\end{figure}

\begin{figure}[h!]
\centering
    \begin{tabular}{cc}
\subcaptionbox{ReliefE-absMean-adaptive vs. SURF}{\includegraphics[width = .45\linewidth]{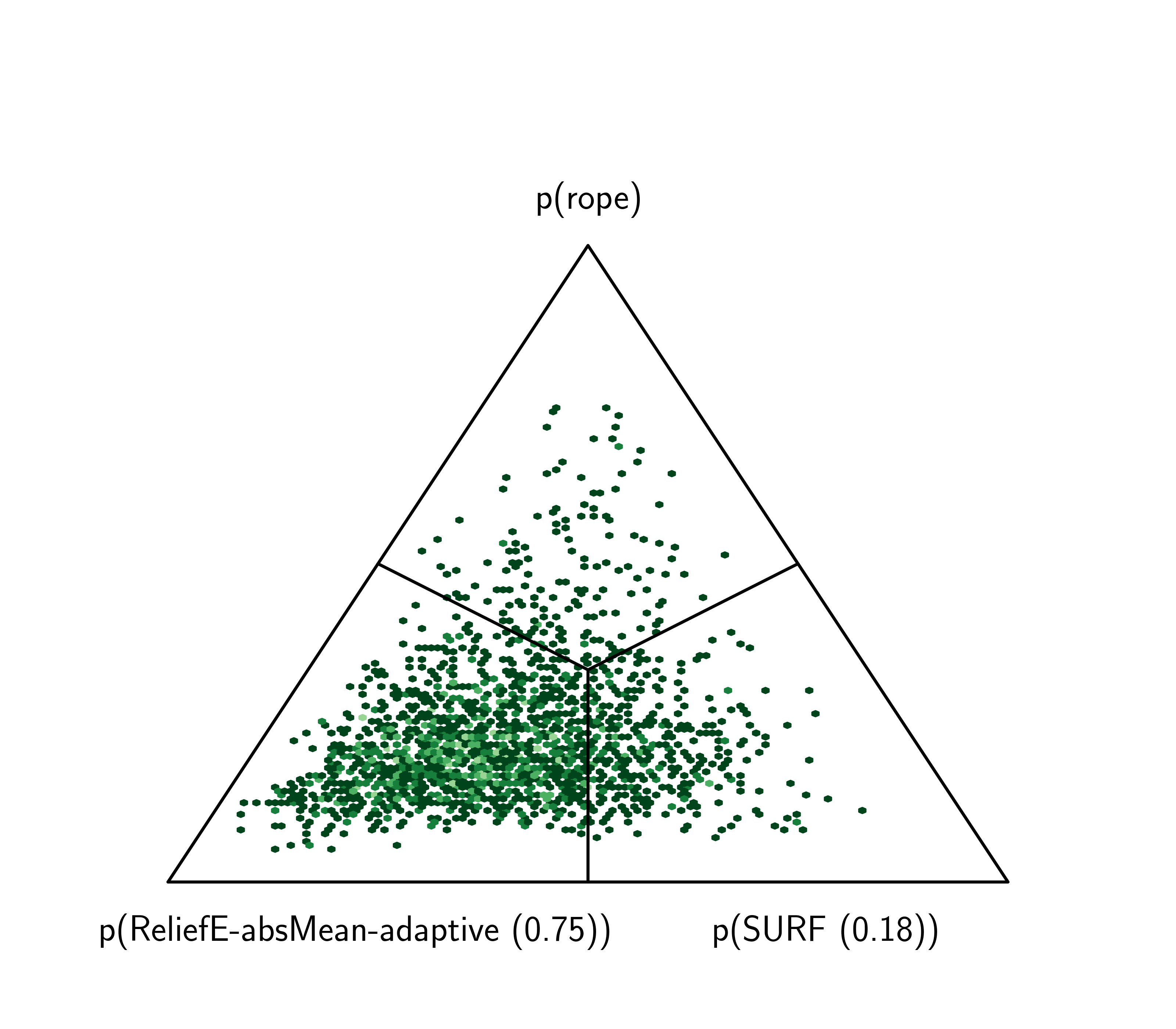}} &
\subcaptionbox{ReliefE-absMean-adaptive vs. MultiSURF}{\includegraphics[width = .45\linewidth]{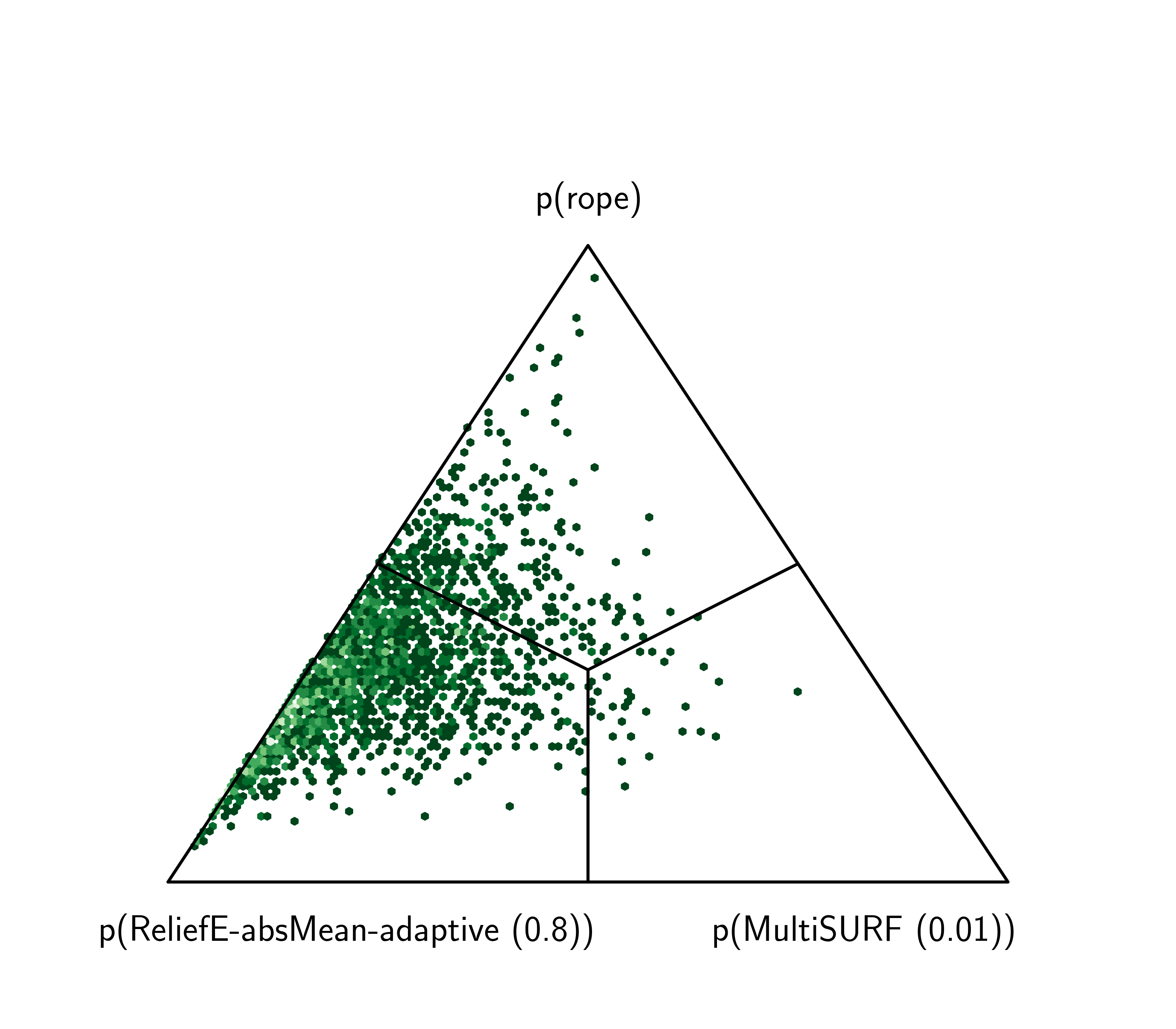}} \\
\subcaptionbox{ReliefE-absMean-adaptive vs. RandomForest}{\includegraphics[width = .45\linewidth]{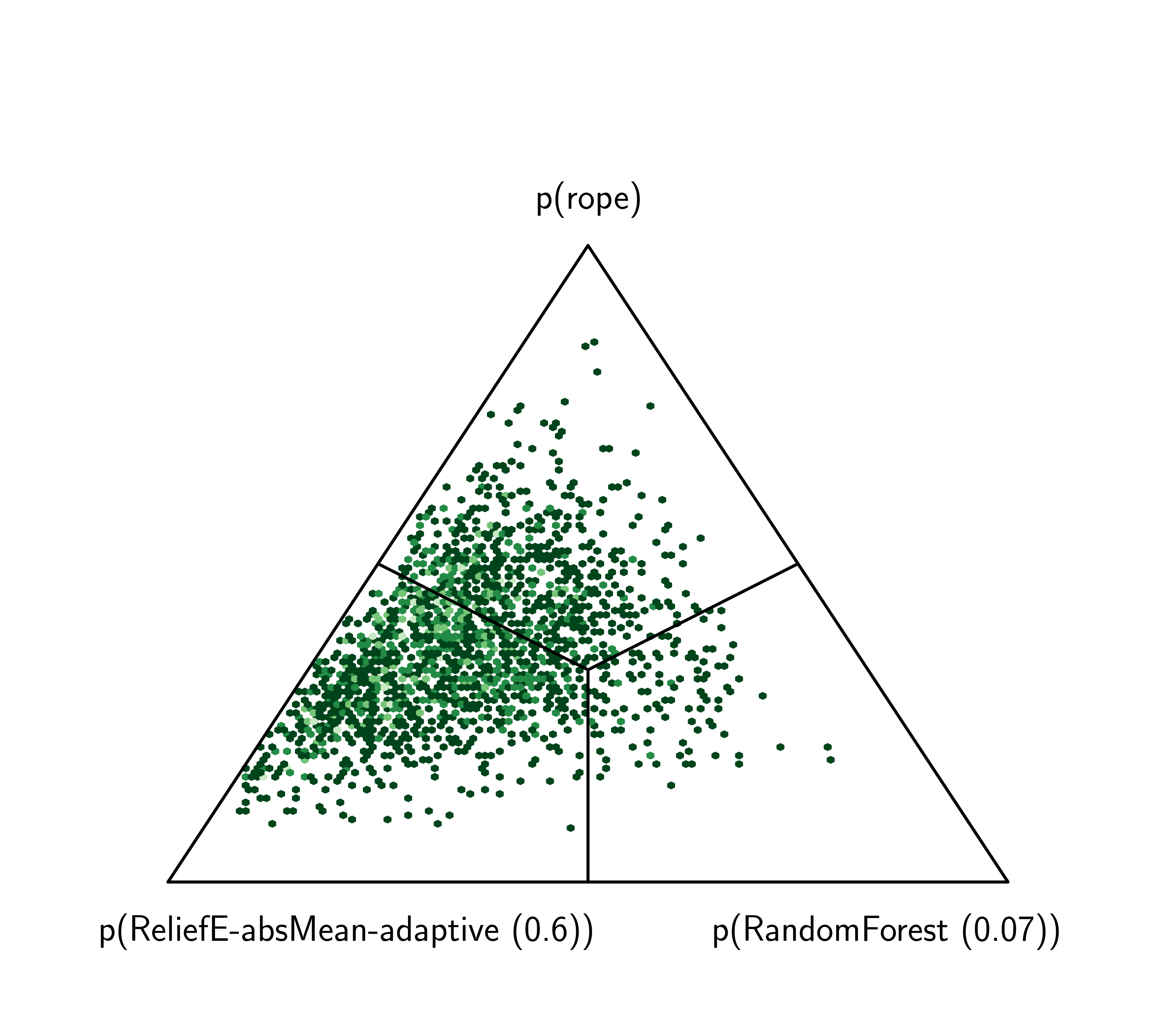}} &
\subcaptionbox{ReliefE-absMean-adaptive vs. ReliefF}{\includegraphics[width = .45\linewidth]{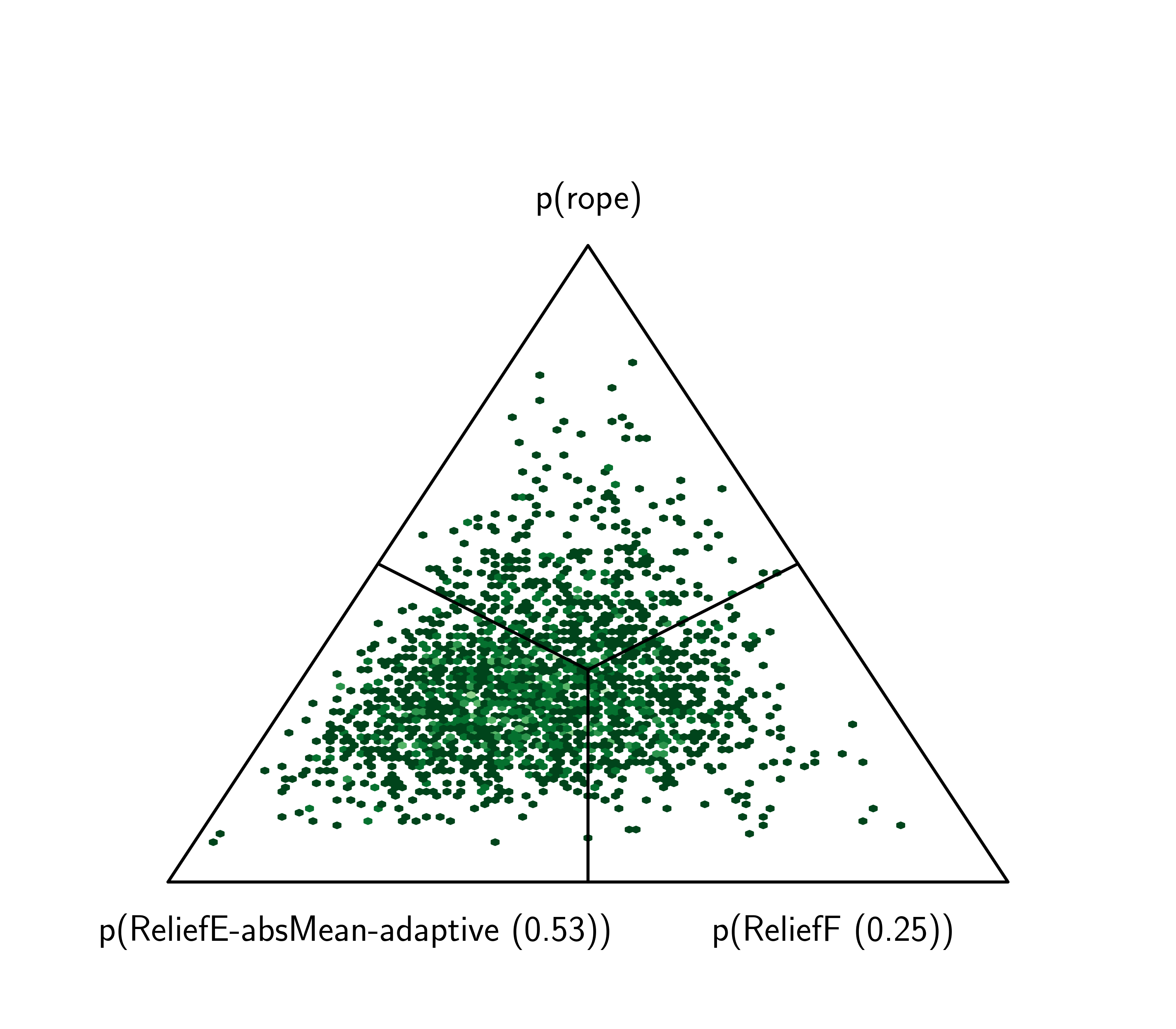}}
\end{tabular}
    \caption{Bayesian comparisons of performance (ranking qualities) between ReliefE-absMean-adaptive and other feature ranking methods for MCC.}
        \label{fig:bayesians2}
\end{figure}

In this section, we present selected Bayesian pairwise comparisons {of} classifiers' performance. Previously determined relationships, such as the dominance of {the} SURF branch of algorithms over mutual information were confirmed, and further extended by adding comparisons with {the proposed} ReliefE branch of algorithms. The comparisons are presented in Figures~\ref{fig:bayesians} and ~\ref{fig:bayesians2}.

{Each diagram has three main regions (parts of the pyramid). The two bottom regions correspond to the samples associated with the dominance of each of the two algorithms compared, and the rope region to the difference space, where the winner is not clearly defined. The probability density directly corresponds to the density of dots in the diagram, thus, the part of the diagram with the highest density implies the most probable situation. Individual (posterior) probabilities are also shown next to each diagram, and denote the probabilities of one algorithm outperforming the other or the algorithms being of similar performance.}

The key results of such pairwise comparisons can be summarized as follows. Very few comparisons yield clear winners. In the majority of the cases, when the most competitive methods are considered, less than 50\% probability that one of the ranking algorithms dominates is observed, giving no strong evidence for dominant ranking algorithms. This is the case also for the diagrams in Figure~\ref{fig:bayesians}.\\[-5mm]

The visualizations in Figure~\ref{fig:bayesians2} show that ReliefE-absMean-adaptive, the implementation proposed in this work, performs on par, or better than many existing, well established approaches such as MultiSURF and RandomForest-based rankings.
However, we observe, in the second part of Figure~\ref{fig:bayesians2}, that ReliefE-absMean-adaptive offers small, albeit incremental win rate when compared against the other methods. With the highest probability (80\%), we can claim ReliefE's dominance against MultiSURF, however, the observed probability ratio does not suffice for a significant difference with $>95\%$ probability (the commonly considered convention).
To further study the algorithm performance, we visualize the top $f$ features---rF1 curves and discuss the selected examples--such figures showing in detail the ranking performance of the different algorithms for the selected data sets are given in Appendix~\ref{appendix-cc:individual}.
Overall, considering the different statistical approaches to evaluating ReliefE's performance, the results indicate that the method has similar performance to its competitors, but offers up to two orders of magnitude faster ranking computation, which also confirms the theoretical findings from Section~\ref{sec:theoretical}.

\subsection{Multi-label classification}
\label{sec:res-mlc}
\begin{figure}[b!]
    \centering
    \includegraphics[width = .8\linewidth]{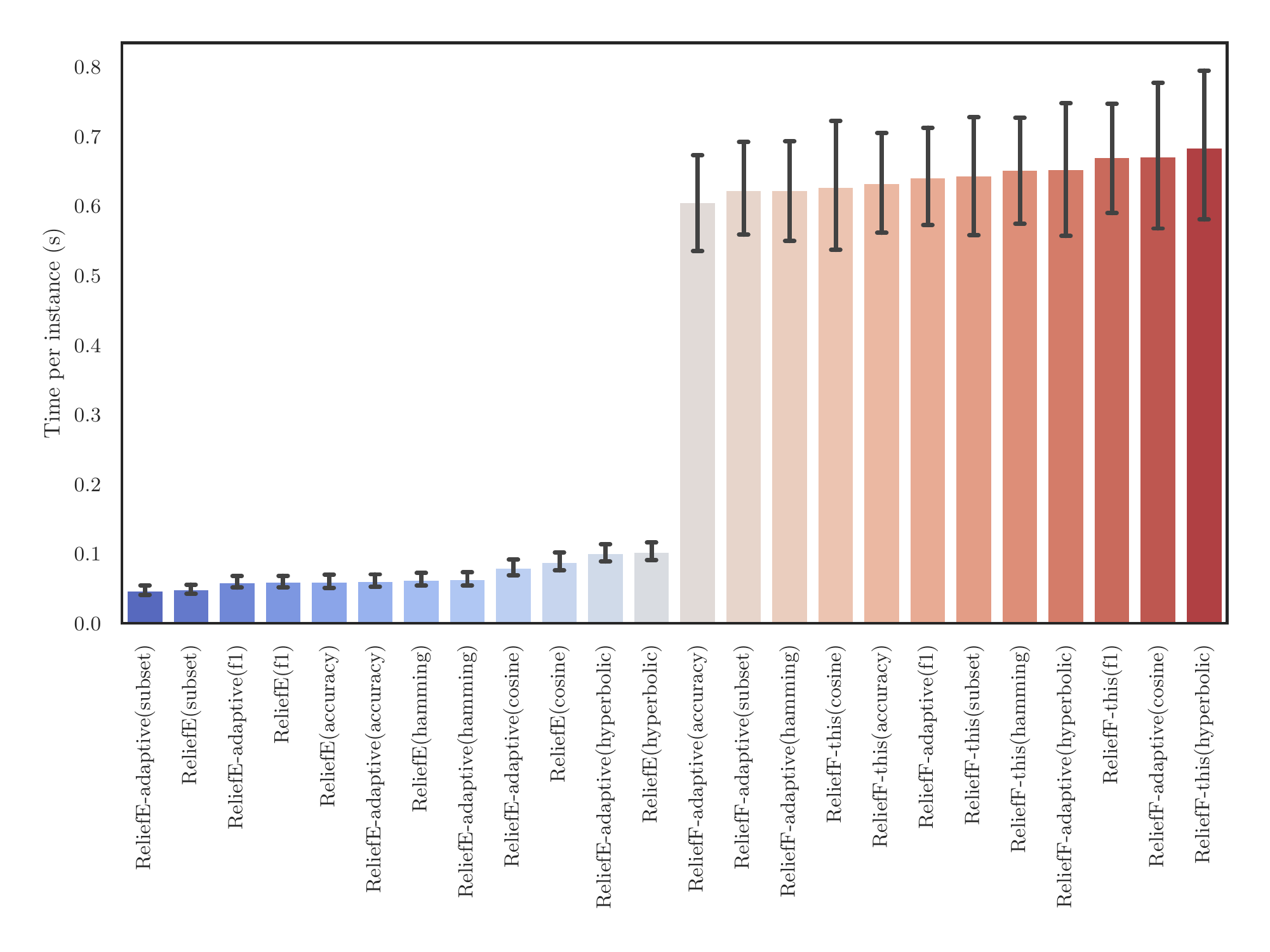}
    \caption{Running times for the MLC variants of ReliefE (and ReliefF, reimplemented in this work and denoted ReliefF-this).}
    \label{fig:timesMLC}
\end{figure}

We next present the results of feature ranking for multi-label classification.
For readability purposes, we present the average rank diagrams in Appendix~\ref{appendix-mlc}. The time required for {the} execution of various distance-ranking algorithm combinations is shown in Figure~\ref{fig:timesMLC}.
The differences in {the execution} times are apparent. The ReliefE branch (blue) offers more than an order of magnitude faster ranking computation.

\begin{figure}[h!]
    \centering
    \includegraphics[width = .8\linewidth]{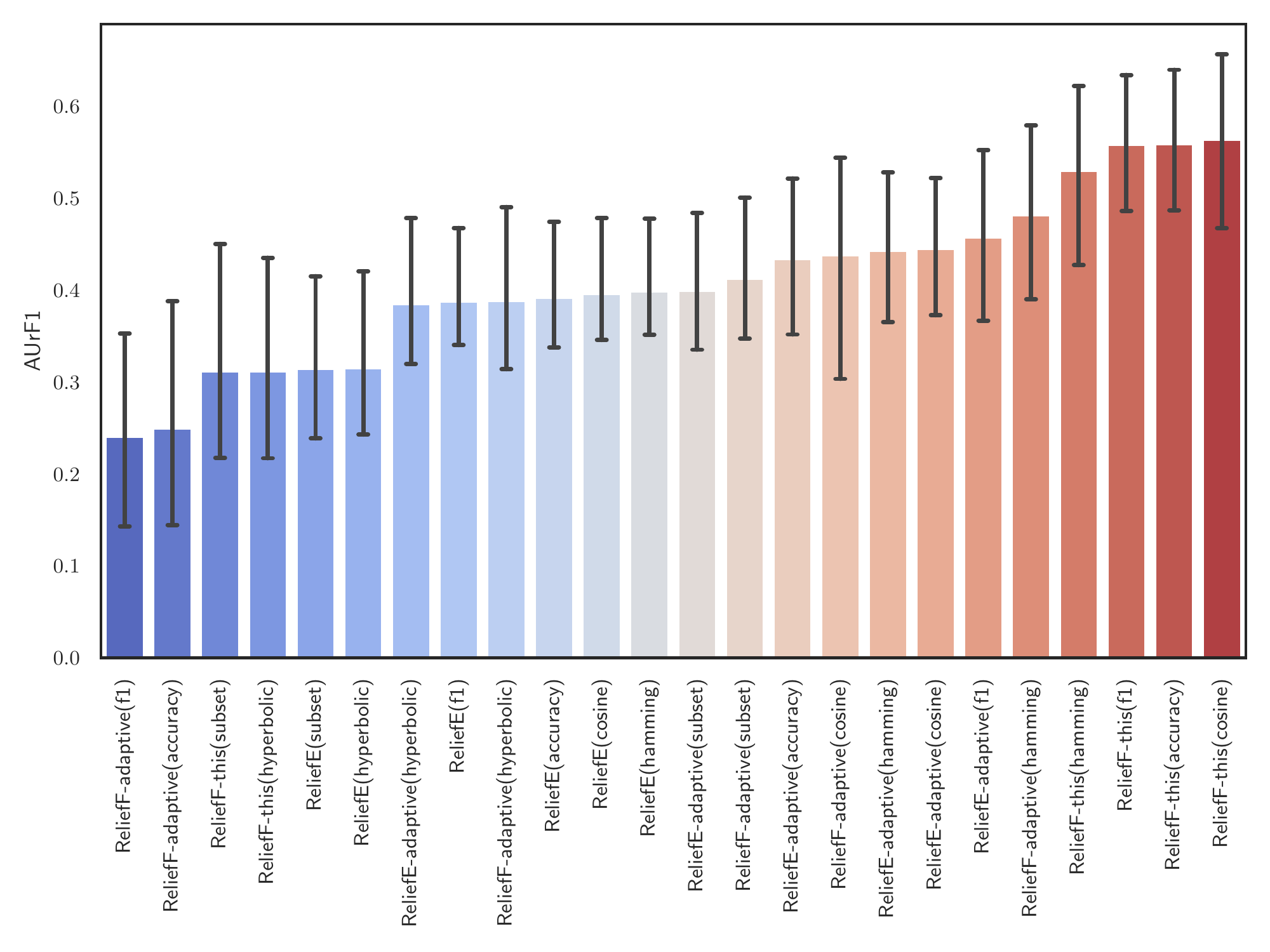}
    \caption{Area under the relative F1 for different ranking approaches in the context of multi-label classification.}
    \label{fig:aucMLC}
\end{figure}

The AUrF1 scores, averaged across data sets are shown in Figure~\ref{fig:aucMLC}.
The best performing ReliefF variants for multi-label classification do not embed the input space. However, the top performant variant employs Euclidean embeddings of the target space, where the distances are computed based on the cosine similarity score. This result indicates multi-label classification can benefit from embedding-based approaches. A case study, where the behavior of various ReliefE variants for MLC is considered in more detail can be found in Appendix~\ref{appendix-mlc:individual}.
% Additional figures showing performance details for a few more selected data sets are given in Appendix~\ref{appendix-mlc}.

\subsection{Relations between ranking algorithms}
\label{sec:results-relations}
\begin{figure}
    \centering
    \includegraphics[width = \linewidth]{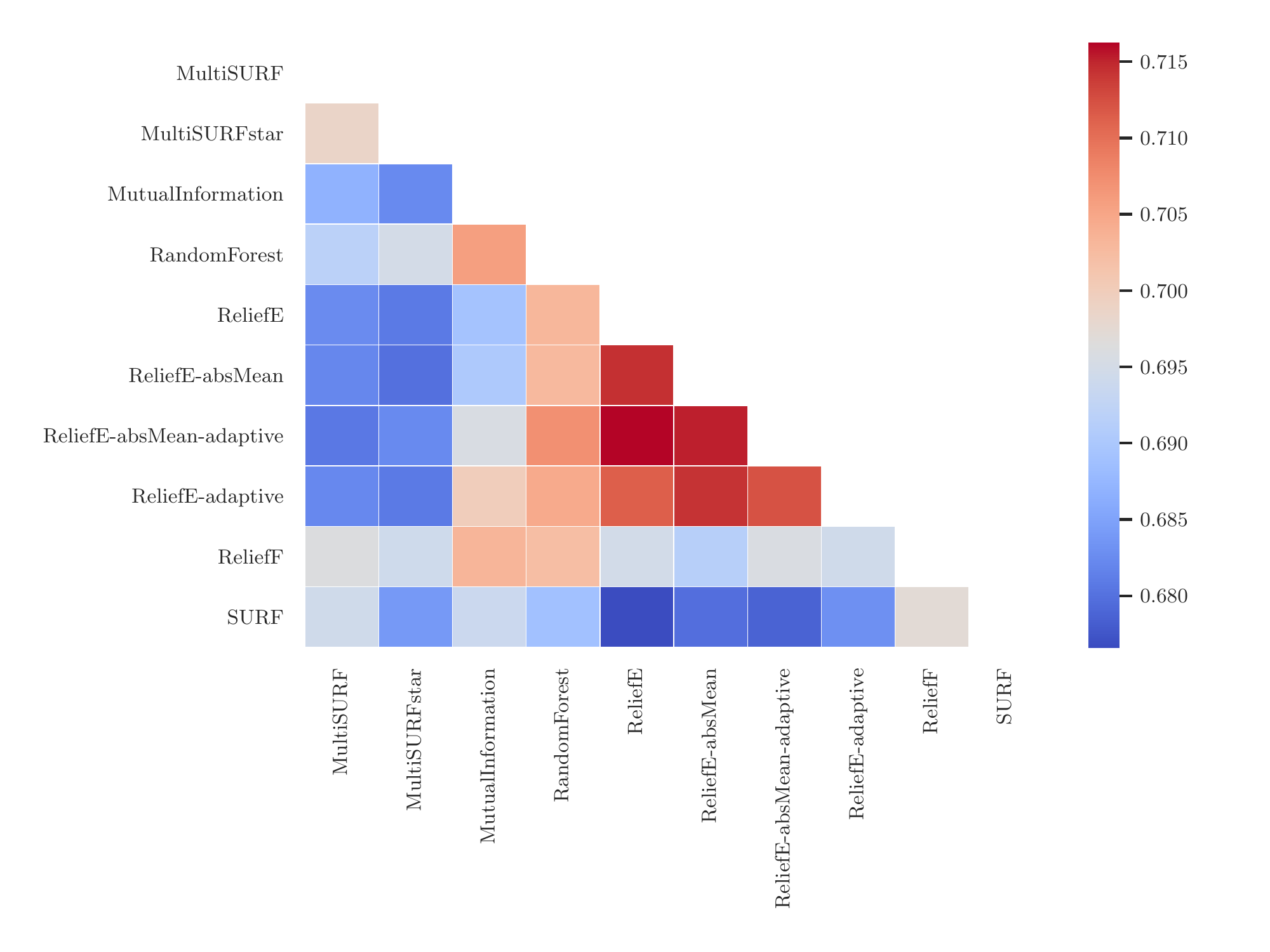}~\vspace*{-5mm}
    \caption{AU\textsc{FUJI} scores for multi-class rankings. Higher numbers (red colors) mean higher similarity between rankings.}
        \label{fig:fuji-cc}
\end{figure}

\begin{figure}
    \centering
        \includegraphics[width = \linewidth]{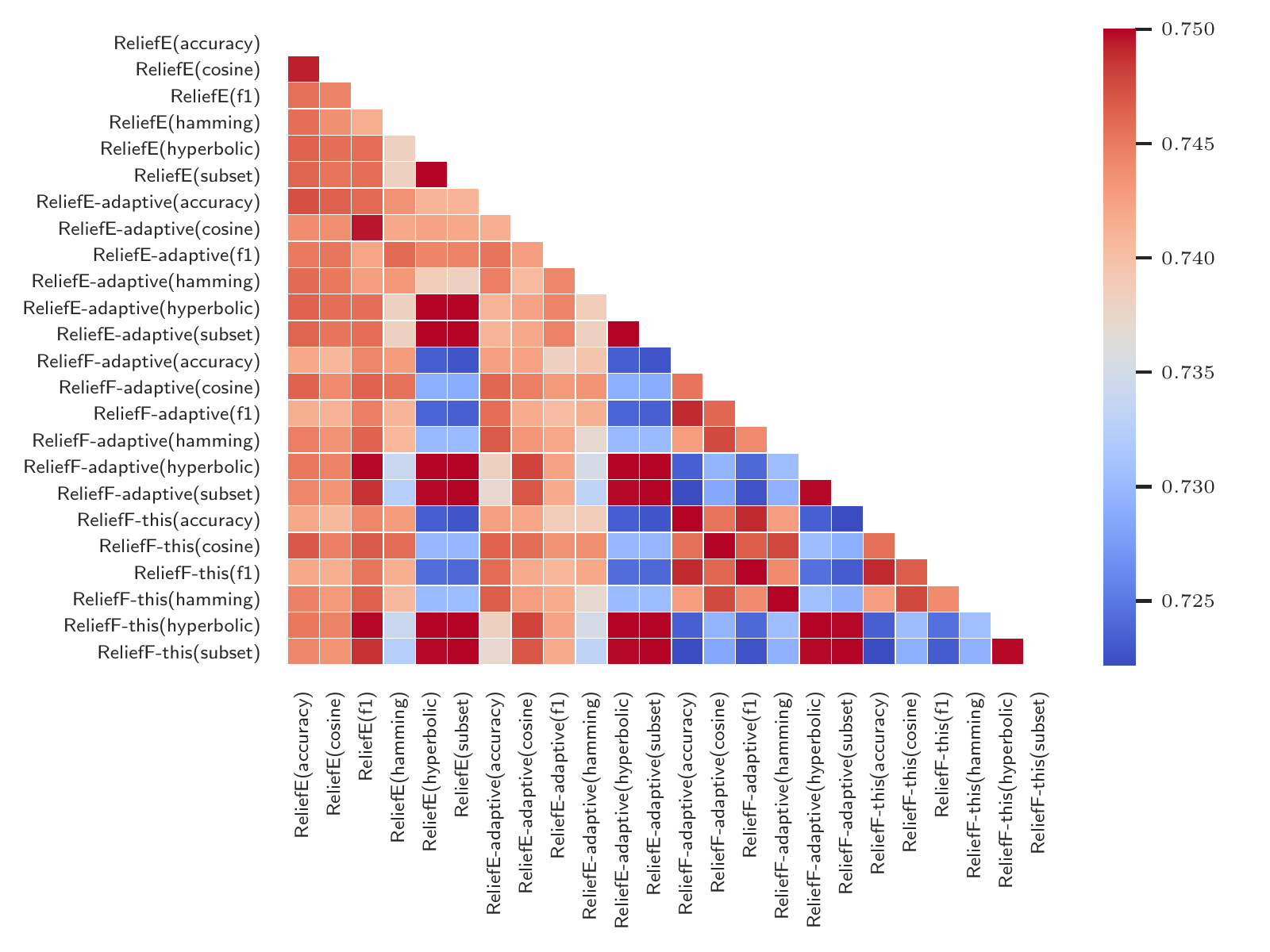}~\vspace*{-5mm}
    \caption{AU\textsc{FUJI} scores for multi-label rankings. The red block of cells in the upper left part of the triangle corresponds to various variants of ReliefE.}
    \label{fig:fuji-mlc}
\end{figure}

We employ the \textsc{FUJI} score, a recently introduced scale-free comparison of ordered positive real-valued lists, to study how different feature ranking algorithms relate to each other. This study employs the same methodology as discussed in \cite{petkovic2020fuzzy,vskrlj2020feature}.
The considered \textsc{FUJI} scores can, apart from the ranking, also take into account the differences between the elements that are being compared---this is not possible by using, e.g., the Jaccard score. We compare pairs of curves comprised of (rF1,top $f$) tuples, thus effectively comparing the \emph{shape} of {the} rankings' performance.
The results of these comparisons are shown in Figure~\ref{fig:fuji-cc} for multi-class classification and in Figure~\ref{fig:fuji-mlc} for multi-label classification.
The most apparent pattern that emerges when these comparisons is that embedding-based rankings (ReliefE variants) tend to give very similar rankings. This holds for both multi-class and multi-label classification rankings.
%Further, the similarity of other baselines, such as SURF and e.g., MultiSURFStar, to classic RelieF can also be observed.

\subsection{Convergence to the final ranking}
\label{sec:results-convergence}
Note that in all {the} examples up to this point, the number of iterations via which the weights corresponding to feature importances were updated was equal to the number of instances (hence the quadratic complexity). Having shown that this setting already offers state-of-the-art performance, we further explored how redundant is the iteration process, i.e., what is the minimum number of iterations needed to obtain a similar ranking. We investigated this question on MCC datasets following the approach described below.

\begin{figure}
    \centering
    \begin{tabular}{cc}
\subcaptionbox{DLBCL  \label{fig:ablation-dlbcl}}{\includegraphics[width = .47\linewidth]{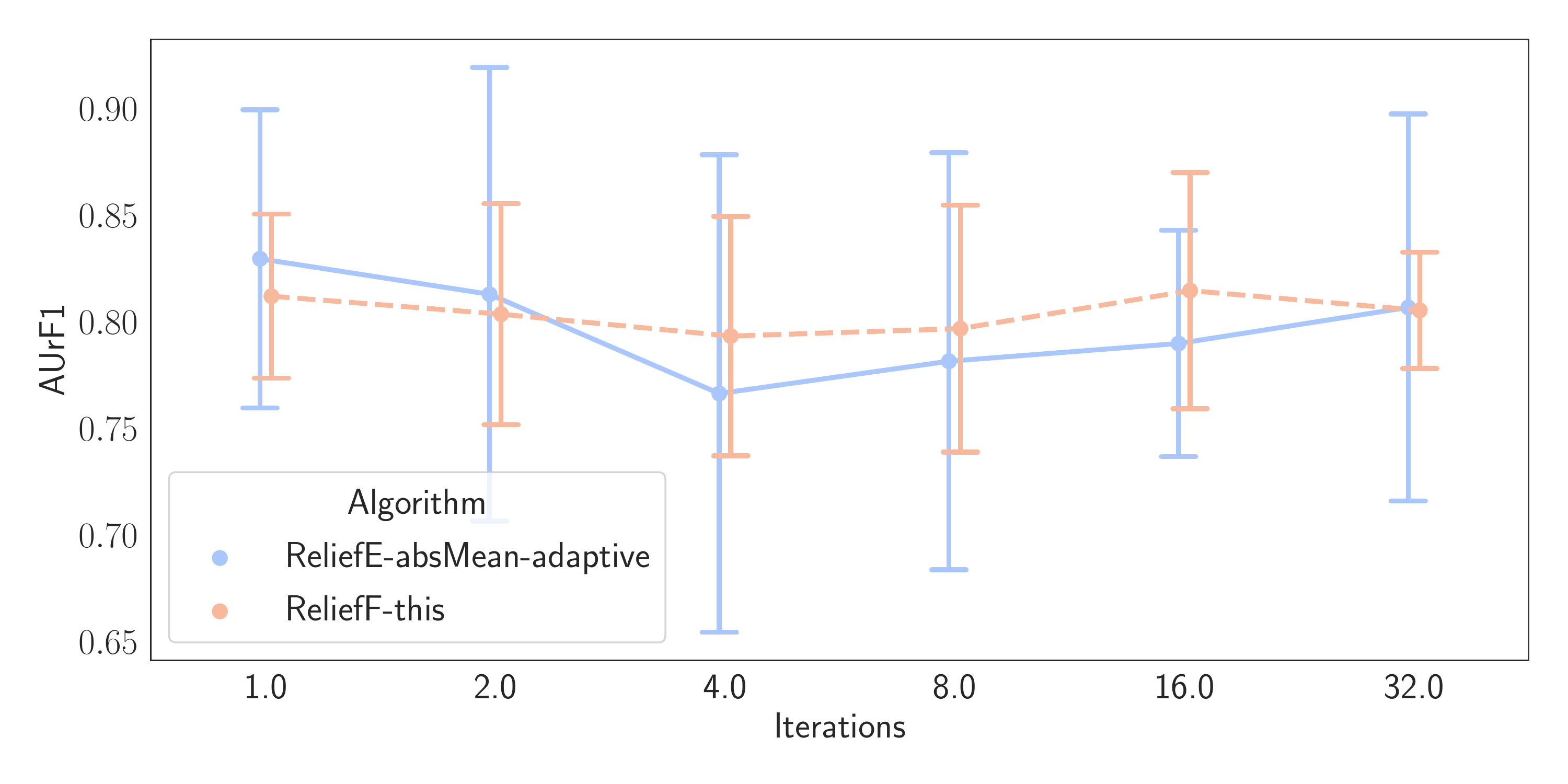}} &
\subcaptionbox{CHESS  \label{fig:ablation-chess}}{\includegraphics[width = .47\linewidth]{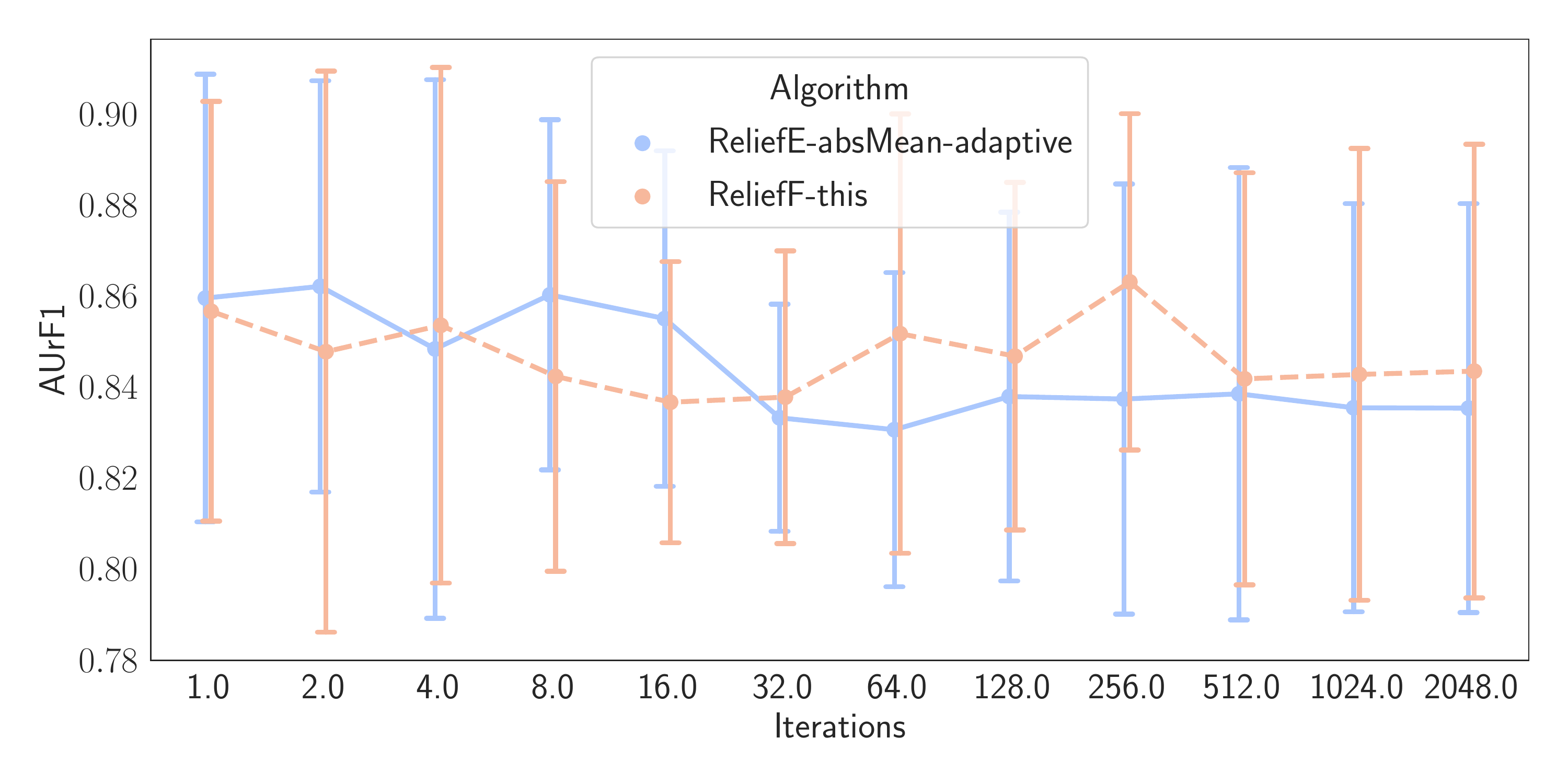}} \\
\subcaptionbox{Biodeg-p2-discrete \label{fig:ablation-biodeg}}{\includegraphics[width = .47\linewidth]{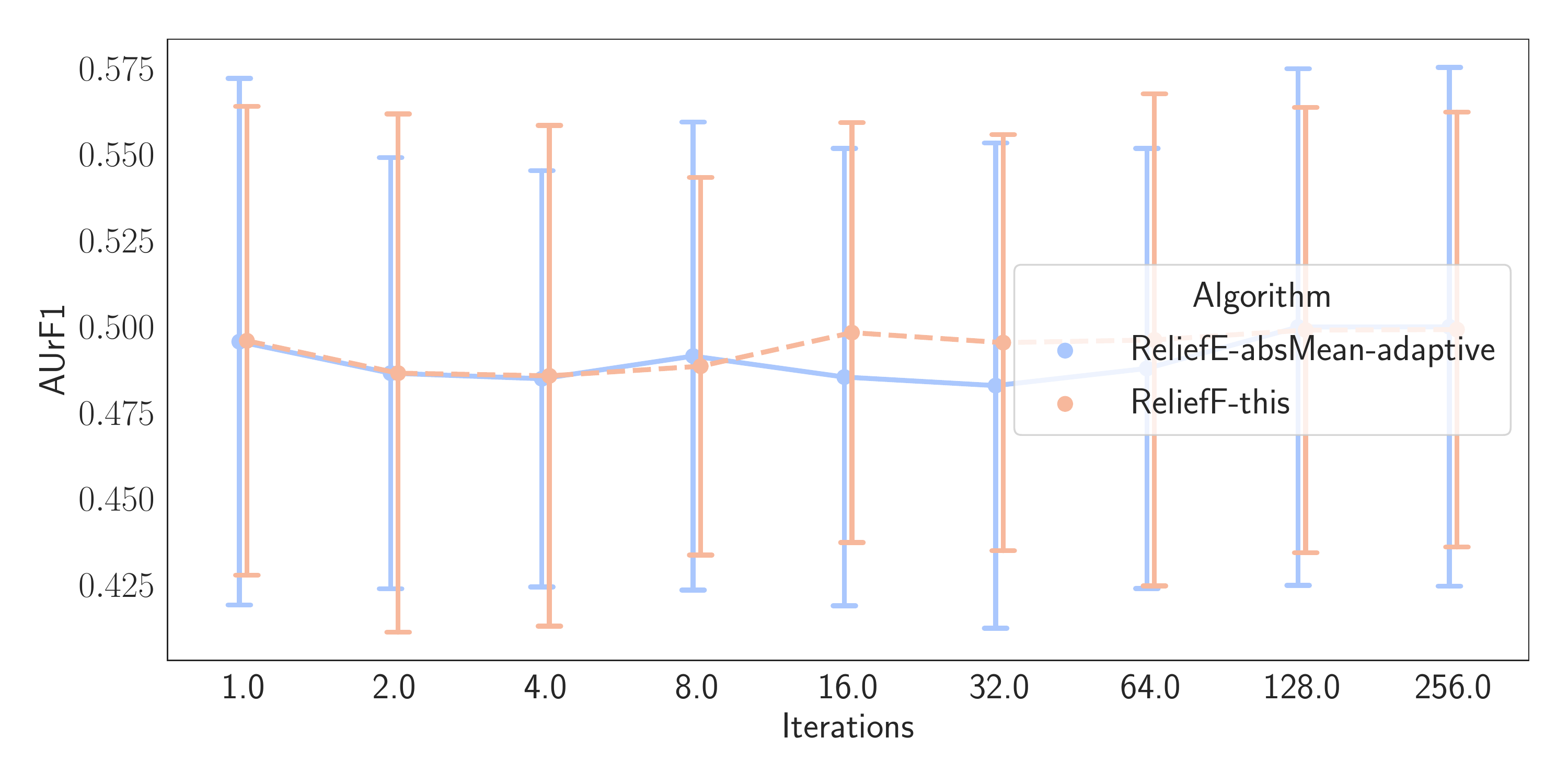}} & \subcaptionbox{Tumors C \label{fig:ablation-tumors}}{\includegraphics[width = .47\linewidth]{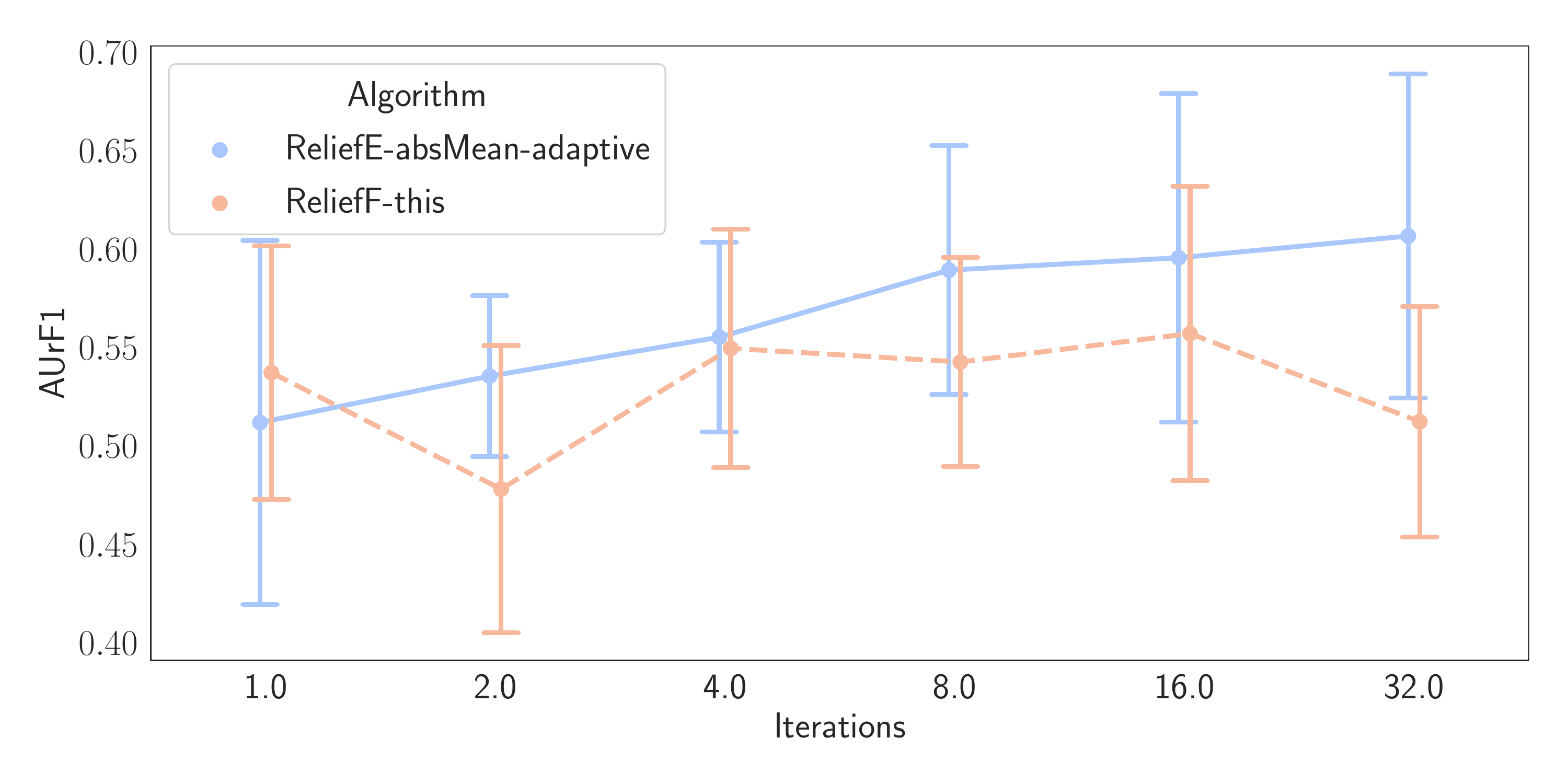}} 
 \\
\end{tabular}
    \caption{Impact of the number of ReliefF iterations on ranking quality.}
        \label{fig:ablation-iterations}
\end{figure}

For each number of considered iterations, we conducted 100 logistic regression runs building models with up to 100 top-ranked features. We computed the AUrF1 and inspected the curve induced by the obtained series of (top $f$, AUrF1) tuples. We conducted these experiments for the DLBCL, Tumors C, Biodeg-discrete and chess data sets, with the results shown in
%Figures~\ref{fig:ablation-dlbcl}, ~\ref{fig:ablation-chess}, ~\ref{fig:ablation-tumors} and Figure ~\ref{fig:ablation-biodeg}.
Figure \ref{fig:ablation-iterations}. We compared ReliefE-absMean-adaptive with ReliefF as implemented in this work, evaluating each iteration with three-fold cross validation (same splits).

It can be observed in Figure~\ref{fig:ablation-dlbcl} that the convergence is slower with the ReliefE-absMean-adaptive variant, however, once the performance is achieved, it is no longer impacted by additional iterations. This does not appear to be the case with ReliefF, where a decrease is observed when 32 iterations are considered. Overall, however,  ReliefE-absMean-adaptive offers state-of-the-art performance already after four iterations. A similar situation is observed in the case of Biodeg in Figure~\ref{fig:ablation-biodeg}. We also observed that on the Tumors C data set (Figure~\ref{fig:ablation-tumors}), ReliefE-absMean-adaptive was consistently outperformed by ReliefF. Being very high-dimensional, and with only tens of instances, this data set's intrinsic dimension is most likely under-estimated, yielding feature ranking based on representations that loose too much information. The ReliefE branch of algorithms is highly dependent on the underlying embeddings, where construction of high quality embeddings in such data scarce scenarios remains a lively research area on its own.

Potential speedups by decreasing the number of iterations {will be explored in} further work. The performance on the chess data set, however, remains consistent for both algorithms---this is a low-dimensional data set, where feature importance estimation via embedded space does not offer notable performance improvements, both with respect to top F1 and computation time.

\subsection{Relevant negative results}
\label{sec:negative}
Even though the paper proposes a promising Relief variant, capable of operating in high-dimensional sparse spaces, many intermediary steps did not perform as expected, and are summarized below:
\begin{enumerate}
    \item Due to pointer-based storage, using sparse matrix algebra can result in additional overhead, which can be significant in large dense data sets.
    \item Running UMAP with spectral decomposition resulted in an unexpected memory overhead. We circumvented this issue with $\nu$, however, the original implementation, once adapted for large scale embedding, could offer an alternative that is more native to UMAP's routines.
    \item Employment of Numba's parallel capabilities led to somewhat mixed results. On {the} one hand, trivially parallel routines such as independent looping and similar could easily be adapted to run in parallel, however, when the  parallel decorator was employed over the whole ReliefE weight update step, even though all cores were utilized, no notable speedups were observed. Additional study of the intricacies of such decorator-based parallelism is left for further work.
    \item When validating our and scikit-rebate's implementations against Weka's ReliefF, it turned out that ReliefF, as implemented in scikit-rebate differs with a somewhat negative effect on performance (as shown in this paper).
    \item We did not experiment with detailed typing of the most time-consuming methods, however we believe some of the routines could be, this way, made even faster.
    \item The intrinsic dimension algorithm (Algorithm~\ref{algo:dim}) appears to \emph{underestimate} the real dimension, leading to poorer performance in some cases.
    \item Embedding target instances in hyperbolic space either works well, or does not work at all. We believe the observed performances are due to the intrinsic geometry of the data, which we will explore in further work.
\end{enumerate}
We next discuss some of the general observations and their implications.

\section{Discussion}
\label{sec:discussion}
{In this work, we considered extensions of the original ReliefF approach with embedding-based distance computations to both multi-class and multi-label classification settings. We observed that, especially in MLC, embedding the target space can contribute both to lower running time and improve classifier's performance. The distance that showed the most promising results was based on the cosine similarity, which is widely used when considering embedding-based learning and exploration. The main contribution of this work, the ReliefE ranking approach is capable of operating via embeddings of both input and output spaces (e.g., in multi-label classification).}

In this section, we comment on the obtained results and discuss further implications of ReliefE. We first observe that adaptive neighbor selection empirically performs very similarly to implementations where neighbor selection is hard-coded. This positive results indicate that one hyperparameter less needs to be tuned, should the user not have the computational resources for extensive grid searches. Further, the simple adaptation of the update step to take into account the distance to the mean of the neighbors similarly offered competitive results. One of the possible reasons for such performance is the potential cancellation of noise, as with averaging, especially in the embedding space, a joint representation is obtained that can also carry some information on semantic similarity amongst the neighbors.

{Within the proposed ReliefE approach, we also explored how data sparsification can be leveraged to further speed up feature ranking in high-dimensional settings. The sparsification procedure was targeted at larger, higher-dimensional, data sets and did not affect smaller data sets as much.}

{In terms of multi-label classification performance, we observed that the classic adaptation of ReliefF with the proposed adaptive distance and the hamming loss was amongst the best performing options. Interestingly, the variant which used the cosine distance on the target space embeddings, was also amongst the top three best performing solutions, indicating that multi-label classification potentially benefits more by considering only the embeddings of the target space instances (and not of instances in the feature space). Similarly, the absMean variant of ReliefF was also  amongst the top five performed, indicating that this aggregation scheme is competitive to the widely accepted averaging, followed by the absolute value step. The best variant of ReliefE that considered both the feature and the target space embeddings is ranked 13th, indicating that by embedding the feature space, performance is lost (albeit significant speedups can be obtained): This hints at a trade-off between performance and ranking quality. Of the remaining metrics, the subset and hyperbolic distances were amongst the worst performing ones, indicating that hyperbolic embeddings operate well in rather limited settings, possibly where a hierarchical structure of the target space can be observed.}

This work is also one of the first (to our knowledge) to compare the performance curves of different ranking algorithms with the Fuzzy Jaccard Index. We observe that embedding-based algorithms proposed in this work behave very similarly, for both multi-label and multi-class classification. Especially in MLC, two consistent patterns emerge. All ReliefE variants are shown to be very similar to one another (red block in Figure~\ref{fig:fuji-mlc}). However, also the hyperbolic and subset  versions of ReliefF appear to behave similarly to the embedding-based ones, even though the input space was not embedded in these cases. For multi-label classification (Figure~\ref{fig:fuji-cc}), the ReliefE variants again emerge as the most similar (to one another). However, similarly to the MLC comparison, versions of the adapted ReliefF as implemented in this work are also shown to yield similar performance curves to ReliefE-based variants.

Following the results of ablation studies, we believe further speedups could be obtained by considering fewer iterations. Current experiments indicate that potentially quadratic speedup could be obtained, as adequate performance was already observed after $\sqrt{|I|}$ iterations in some cases. Further, the number of iterations could also be adapted dynamically, by monitoring the feature ranking scores and detecting convergence before all iterations are carried out.

When studying individual data sets, e.g., DLBCL and opt-digits, we observe that ReliefE offers superior performance at a fraction of the computation time required by the other methods, indicating that the development of approaches based on ideas introduced in this paper is a sensible research avenue.

{In this work, we have evaluated feature rankings based on classification performance obtained by robust learners, such as logistic regression, which have not been fine-tuned. The purpose of such evaluation was to emphasize the effect of feature ranking. However, extensive studies of the interplay between regularization regimes (e.g., L1 vs. L2) and ranking performance could also offer interesting insights into the robustness of rankings, and further, their purpose. For example, a L1 regularized learner could automatically discard large parts of the feature space: Although this would be considered as feature selection (and not ranking), it would potentially offer similar results. We leave this type of experiments for further work.}

Similarly, the Bayesian comparisons, involving mostly a state-of-the-art feature ranker MultiSURFstar and the proposed ReliefE algorithm(s), indicate that ReliefE is competitive and many times outperforms MultiSURFstar, even in a probabilistic sense. For example, the probability that ReliefE-absMean-adaptive outperforms MultiSURFstar is more than 30\%, with most of the remaining probability density lying in the equal performance (rope) region.

{Finally, we discuss several potentially interesting future empirical studies that would represent a non-trivial extension of the proposed work. Detailed analysis of the algorithms' performance with respect to various properties of the data sets could offer additional insights into when to use what type of ranking. We believe that meta-learning could be a promising research venue, as by linking the data sets' properties with suitable algorithms could largely benefit situations where embedding-based ranking is not the best option. Overall, if one optimizes for efficient performance on large, contemporary data sets, ReliefE offers a computationally efficient approach, that could serve as a first step to further study where to invest the remaining computational resources, and whether feature ranking is a sensible approach at all (it might not be for, e.g., image-based data). Similarly, understanding whether the choice of the distance score can be further \emph{transferred} between similar data sets also represents an interesting research direction worth of further study. Overall, the proposed paper provides an empirical, as well as a theoretical foundation for potentially more involved embedding schemes, such as e.g., (variational) autoencoder-based ones. We believe that a relevant factor influencing ReliefE's performance is the quality of the \emph{learned} representation, indicating that another promising research venue could be the investigation of different embedding approaches (this work explores different distances within a single embedding approach, but does not consider different embedding approaches).}

\section{Conclusions and further work}
\label{sec:conclusions}
In this paper, we have proposed one of the first embedding-based Relief implementations with both theoretical and practical grounding. We have explored whether embedding the input, but also output space onto a Riemannian manifold prior to feature ranking {yields} better rankings. The results indicate that, while being significantly faster, embedding-based ranking methods do not consistently outperform the ones that do not use embeddings. However, we show that they are indeed consistently faster than all other Relief-based ranking approaches. 

We also show that for multi-label classification, where additional complexity arises due to multiple label co-presence, ReliefE can offer more stable, and on data sets like Delicious, better performance. Further, we demonstrate that embedding the target label space is beneficial for the final ranking's quality in a multi-label setting. The proposed adaptive neighbor estimation procedure could be further developed in terms of the neighborhood dependence with respect to a given metric. Similarly, the current implementation potentially over-estimates the neighborhood size, which could be due to the nature of the embedded space or the method's bias. Both possibilities are to be explored in future work.

We believe that comparison of feature ranking algorithms should also be considered at the level of their properties and not only their performance. In this work, we show that embedding-based ranking gives rise to a fundamentally different type of rankings, which we believe are worthy of being studied {further}. To our knowledge, we are the first to perform such a large-scale comparisons of a long list of ranking approaches (using, e.g., different similarities in MLC) and take into account also the actual values of importance scores withing rankings (through the \textsc{FUJI} score), and not only the feature order.

We also observe that the variants of the original ReliefF, as re-implemented in this work, already offer superior performance to, e.g., the SURF branch of algorithms, indicating that their scikit-rebate implementations have some limitations in terms of numeric stability (and are not adapted at all to handle sparsity).

As further work, we believe the study of non-Euclidean spaces could yield many novel insights, as the target space is {frequently} of hierarchical nature, implying Euclidean geometry is not sufficiently good for its representation. In this work, we show initial results for embedding on a hyperboloid (Pincar\'e ball model). However, Lorenzian geometry can also be considered.

\begin{acknowledgements}
We would like to acknowledge the Slovenian Research Agency (ARRS) for funding the first and the last author (B\v{S}, MP) through young researcher grants and supporting other authors (SD, NL) through the research program \emph{Knowledge Technologies} (P2-0103) and the research project \emph{Semantic Data Mining for Linked Open Data} (financed under the ERC Complementary Scheme, N2-0078). This research was also partially supported by TAILOR  (a project funded by the EU Horizon 2020 research and innovation programme under GA No 952215) and  AI4EU (GA No 825619). We would also like to thank the administrators of the SLING supercomputing environment for the computing resources which made the empirical part of this study possible. \end{acknowledgements}


\bibliographystyle{spmpsci}       % APS-like style for physics
\bibliography{references}   % name your BibTeX data base

\appendix
\section{Theoretical considerations of embedding spaces}
\label{appendix:theory}
%{TODO -petkovic:} 
%{petkovic has done: js bi to ven vrgel, kvecjem morda tm pr fig 1 (kjer je delotok predstavljen), lahko omenimo, da se F --$>$ w da zapisat kot kompozija dveh funkcij, drugo pa tlele nima zares pomena?}
As many of the recently introduced embedding-based methods tend to replace earlier methods, whilst maintaining the representation power, we believe the comparison of the two mappings, when considered, can be represented as a simple commutative diagram. The example, considered to represent ReliefE's mapping compared to, e.g., that of the standard ReliefF's can be represented as
\begin{figure}[ht!]
\centering
\begin{tikzcd}
\boldsymbol{F} \arrow{r}{\phi} \arrow[swap]{rd}{q} & \boldsymbol{E} \arrow{d}{f} \\
& \boldsymbol{w}
\label{diag}
\end{tikzcd}
\end{figure}
\noindent Here, the initial, real valued feature matrix $\boldsymbol{F}$ is either directly ($q$), or indirectly ($\phi$ and $f$) mapped to the output weight vector $\boldsymbol{w}$. Note that $q: \mathbb{R}^{|I| \times |F|} \rightarrow \mathbb{R}^{|F|}$, $\phi: \mathbb{R}^{|I| \times |F|} \rightarrow \mathbb{R}^{|I| \times d}$ and $f: \mathbb{R}^{|I| \times d} \rightarrow \mathbb{R}^{|F|}$. ReliefE operates under the assumption that the initial ranking can be retrieved via latent space $\boldsymbol{E}$ in two steps ($\phi$ and $f$).

\section{Ablation study of data sparsity}
\label{appendix-sparseness}
\begin{figure}[b!]
    \centering
    \includegraphics[width = .8\linewidth]{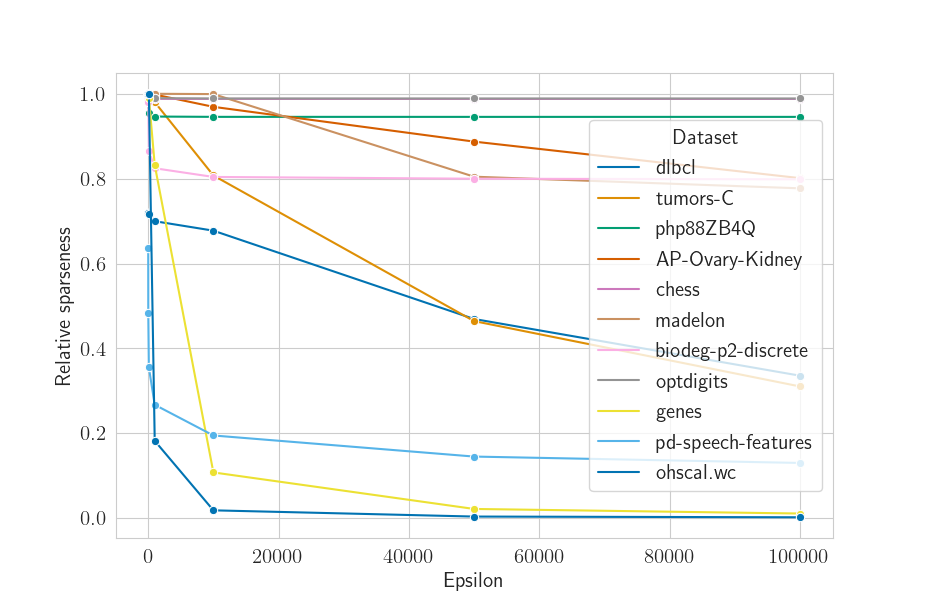}
    \caption{Dependence of sparsification on approximation error allowed.}
    \label{fig:sparsification-eps}
\end{figure}

\begin{figure}[h!]
    \centering
    \includegraphics[width = .7\linewidth]{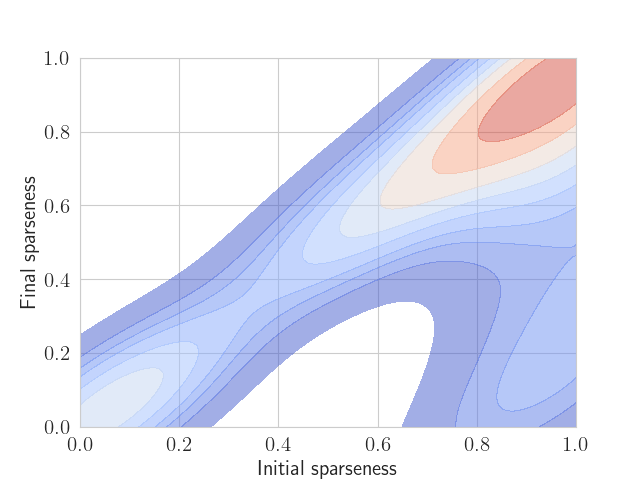}
    \caption{Kernel density estimation of the relation between the initial and final sparsity.}
    \label{fig:sparsification-kde}
\end{figure}

The considered sparsification procedure is dependent on the parameter \emph{epsilon}, i.e., the approximation error. The study of how different error thresholds impact the final sparsification result is shown in Figure~\ref{fig:sparsification-eps}.
{It can be observed that most of the data sets only get sparsified after a rather large epsilon is permitted.  The second ablation explores the relation between the initial data sparsity and the final sparsity, i.e., the sparsity of a given data set after the conducted sparsification procedure. The result is shown as a kernel density plot in Figure}~\ref{fig:sparsification-kde}.

\begin{figure}[h!]
    \centering
    \includegraphics[width = .7\linewidth]{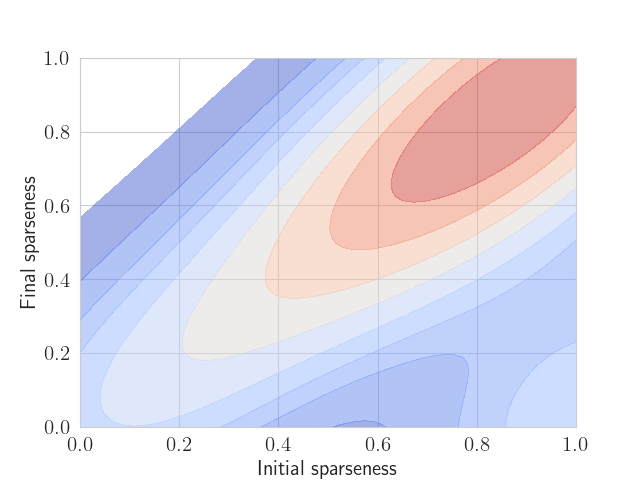}
    \caption{Kernel density estimation -- estimated epsilon values.}
    \label{fig:estimated-epsilon}
\end{figure}

{The observed result indicates that when the data set is sparse to begin with, the result will be, as expected, similarly sparse. However, the vertical density at the rightmost part of the figure demonstrates that the sparsification procedure indeed yields sparser data, albeit not in all cases.}
A similar visualization can be produced for the space based on estimated epsilon values, shown in Figure~\ref{fig:estimated-epsilon}.

{The considered estimate yields a similar landscape sparseness to that obtained via grid-search  (Figure~}\ref{fig:sparsification-kde}),{ indicating decrease in most cases. However, there are examples where a given data set's density was substantially lowered, such as for example pd-speech-features and biodeg-p2-discrete. The results indicate that the considered estimate could be further relaxed, albeit at the cost of worse approximation of the input matrix, which could negatively impact the final performance.}

\vspace*{-0.5cm}

\vspace*{-2cm}

\vspace*{-1cm}

\section{Adaptive $k$ distributions}
\label{apppendix:adaptive}
\begin{figure}[h!]
    \centering
    \vspace*{-3mm}~\includegraphics[width = 0.9\linewidth]{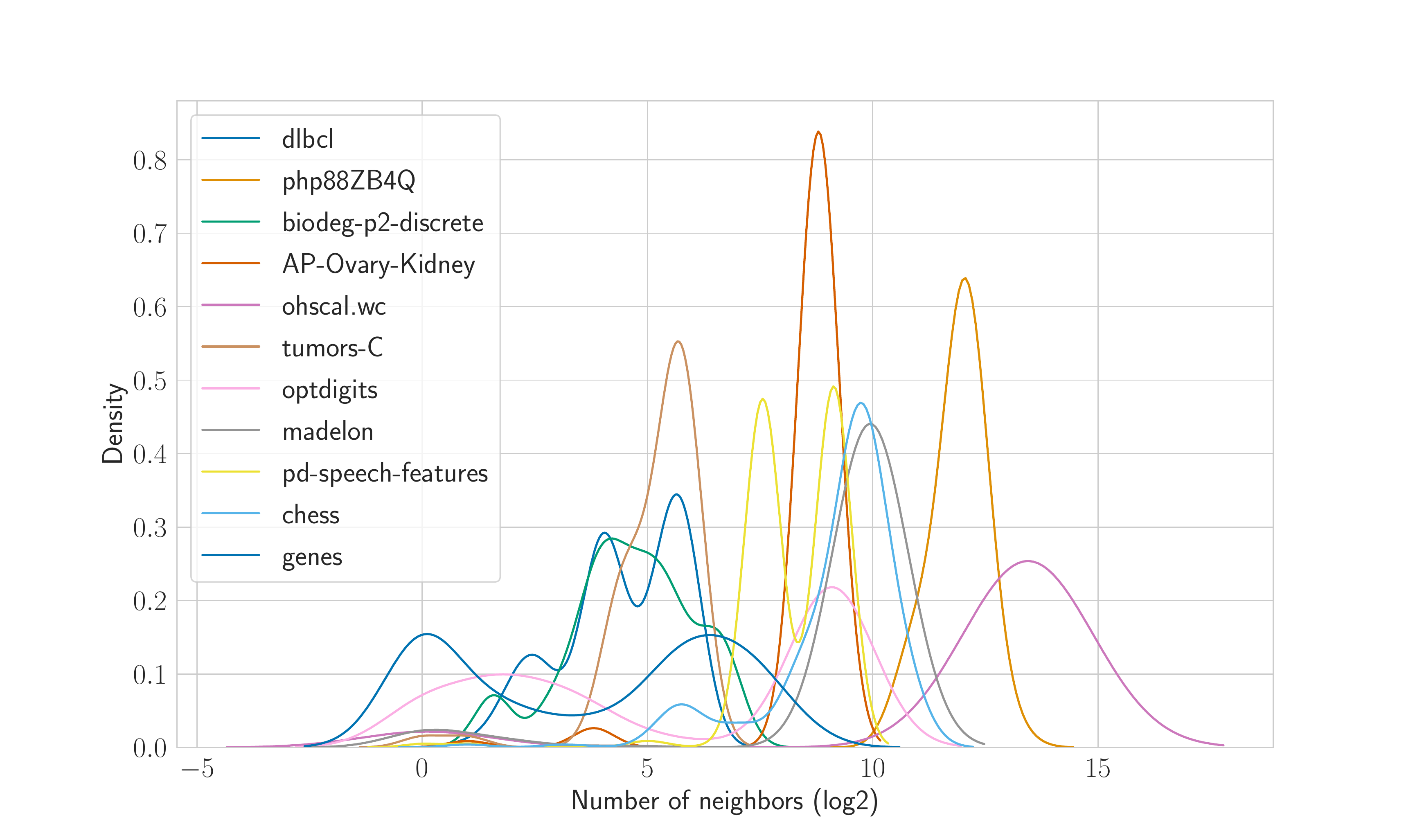}
    \caption{Density of estimated $k$ values for 100 iterations of ReliefE.}
    \label{fig:ablation-adaptive}
\end{figure}

{In this ablation study, we visualized the distributions of the neighborhoods across all considered MCC data sets. This plot demonstrates that for different data sets, differently sized neighborhoods were identified by the proposed heuristic (Figure~}\ref{fig:ablation-adaptive}).

\section{Area under the rF1}
\label{appendix:aurf1}
\begin{figure}[h!]
    \centering
    \includegraphics[width = 0.8\linewidth]{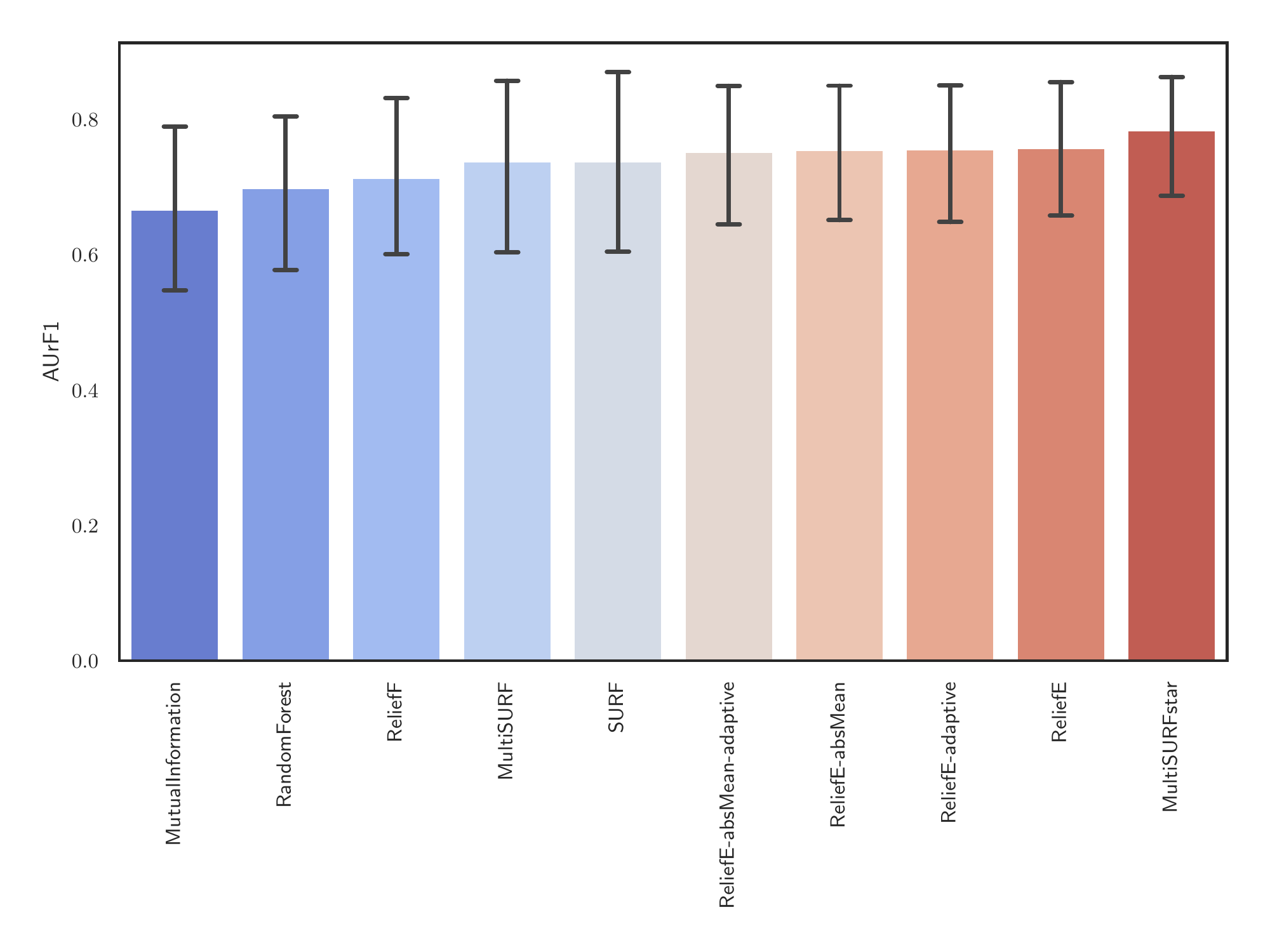}
    \caption{Area under the relative F1 curve for multi-class classification. All ranking approaches perform similarly with no notable differences. More insights into the relative performance of ranking algorithms are provided by Bayesian tests and $\textsc{FUJI}$-based comparisons of performance curves.}
    \label{fig:aucCC}
\end{figure}

The AUrF1 scores, averaged across data sets are shown in Figure~\ref{fig:aucCC}.
It can be observed that the first 5 rankings behave very similarly w.r.t. this measure. Thus, we emphasize other types of comparison, where the differences are more apparent.

\section{Detailed analysis of running time}
\label{appendix:time}

\begin{figure}[h!]
    \centering
    \vspace*{-5mm}~\includegraphics[width = .6\linewidth]{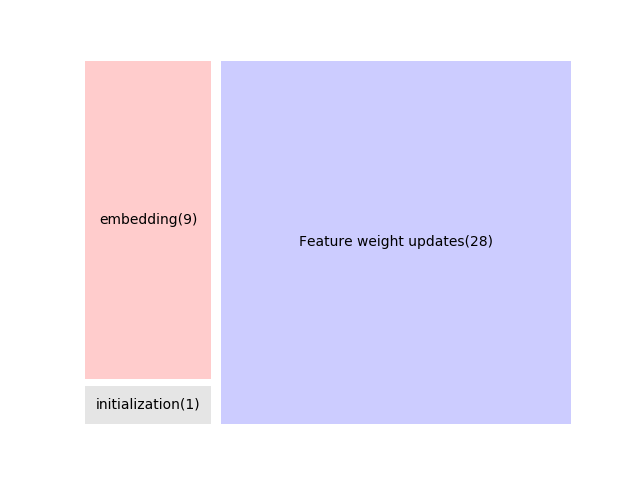}~\vspace*{-5mm}
    \caption{Proportions of time spent on different methods within ReliefE on the p53 dataset.  Majority of time is spent on weight update steps (as expected).}
    \label{fig:proportions}
\end{figure}

We additionally studied how different parts of ReliefE impact the total running time. For p53, one of the largest considered data sets, we visualize the proportions in Figure~\ref{fig:proportions}.

%This appendix presents the analysis of the proportions of time spent at different parts of the algorithm, showing that most time is spent on feature weight updates (Figure~\ref{fig:rt}).

\section{Multi-class classification, additional rank diagrams}
\label{appendix:additional-rank}
\begin{figure}[htb!]
    \centering
        \includegraphics[width = 1.1\linewidth]{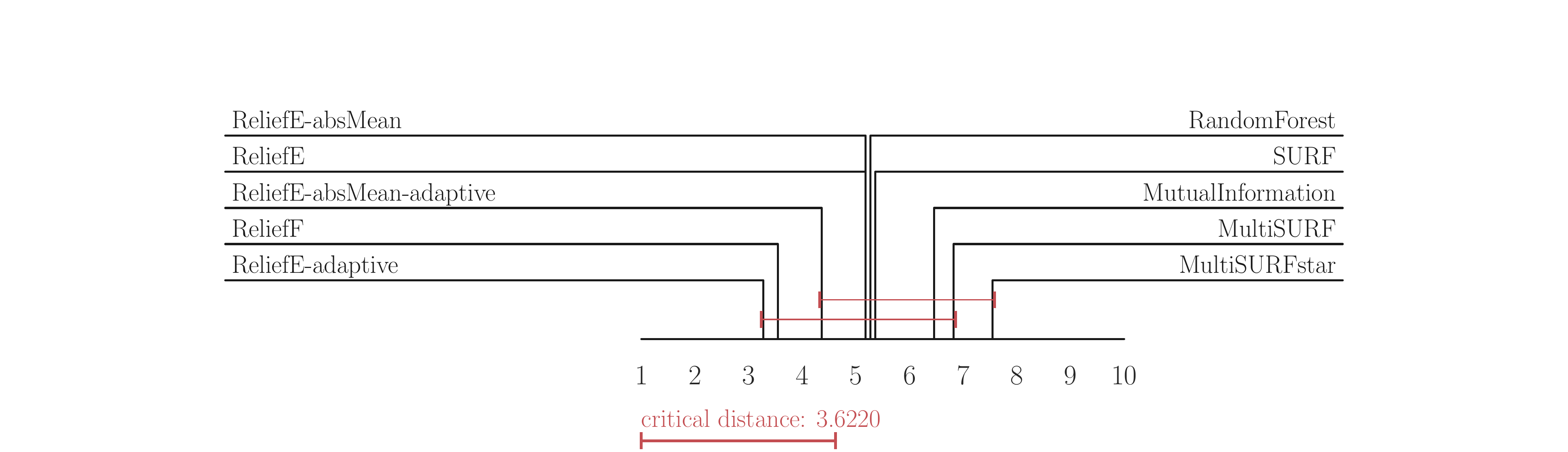}
    \caption{Max first 100 features. {Similarly to the situation with 50 top features, the ReliefE variants, including the adaptive one, perform well for the multi-class classification task.}}
    \label{fig:ablation-mcc-cd-10}
\end{figure}

\begin{figure}[htb!]
    \centering
        \includegraphics[width = 1.1\linewidth]{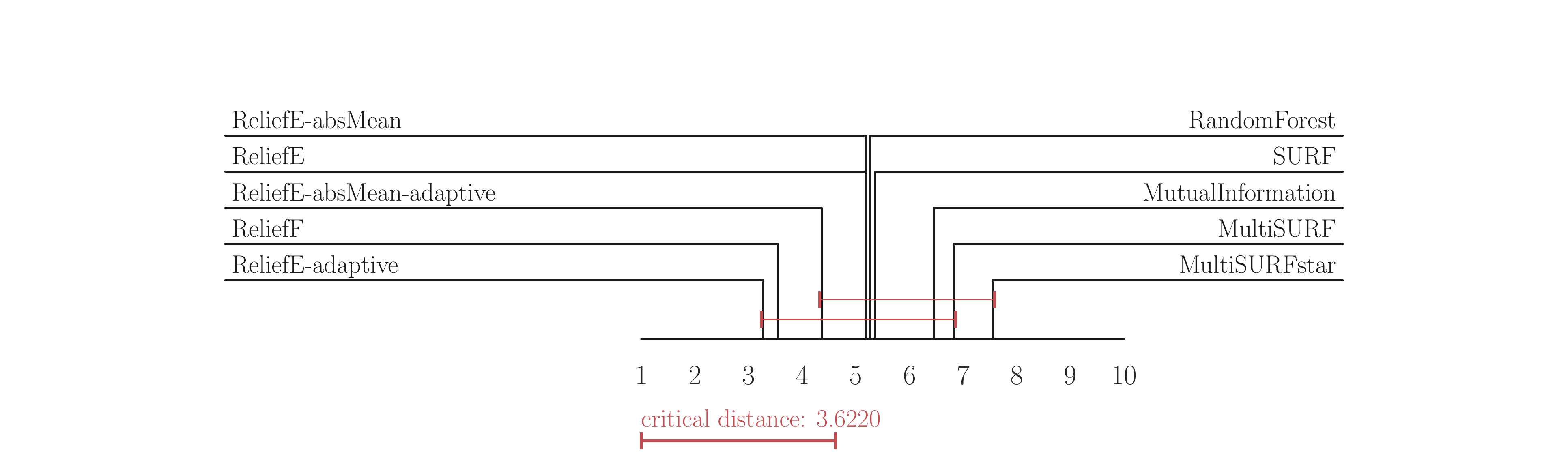}
    \caption{Max first 50 features. {The performance if considering only first 50 features. The adaptive version of ReliefE performs on average the best in this scenario.}}
    \label{fig:mcc-cd-50}
\end{figure}

This appendix includes additional ablation studies in the form of critical distance diagrams presenting the performance for multi-class ranking (Figures~\ref{fig:ablation-mcc-cd-10} and ~\ref{fig:mcc-cd-50}).


\section{Multi-class classification -- case study with Madelon, DLBCL and genes}
\label{appendix-cc:individual}
\begin{figure}[b!]
    \centering
    \vspace*{-5mm}~\includegraphics[width = .8\linewidth]{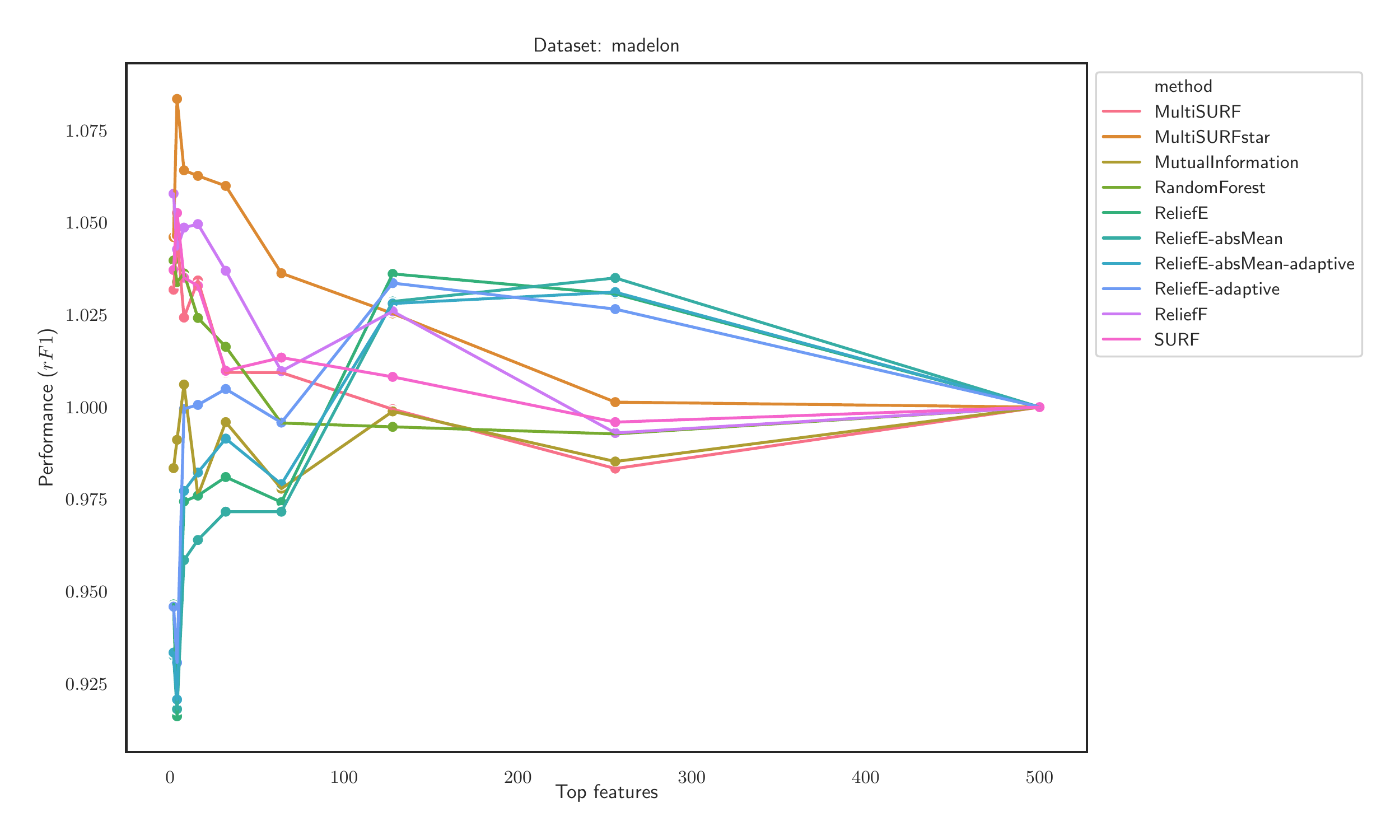}
    \caption{Madelon performance curves. {This result indicates that there exist situations where initially better rankings are obtained via e.g., the SURF branch of the algorithms, however, when considering more features, ReliefE variants are the only ones that find rankings which perform well.}}
    \label{fig:apx-madelon}
\end{figure}

This section contains feature rankings, visualized for the Madelon data set, where either the ReliefE or a variant of ReliefF equiped with one of the proposed heuristics shows different behavior (better performance) (Figure~\ref{fig:apx-madelon}.) The visualized performances offer insights into behavior of the algorithms. For example, the ReliefF branch of adaptations (and vanilla ReliefF) peak at less than a hundred features, however, another performance peak where feature ranking is sensible ($rF1 > 1$) is around 250 features, where ReliefE-type algorithms are consistently amongst the best-performing ones. 

\begin{figure}[h!]
    \centering
    \vspace*{-5mm}~\includegraphics[width = .8\linewidth]{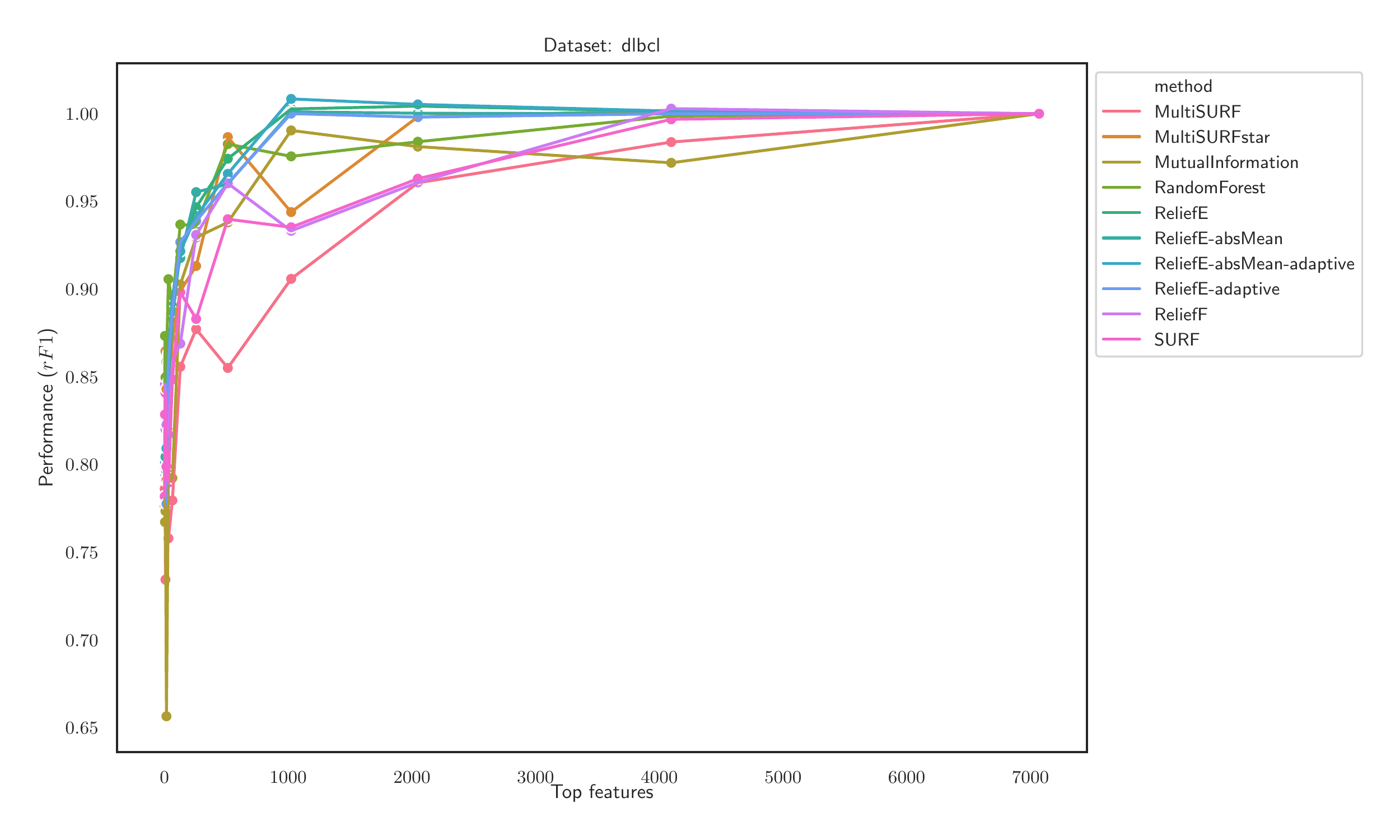}
    \caption{DLBCL performance curves.  {The DLBCL is a very high-dimensional data set and reflects the ReliefE's capability to operate with high-dimensional feature spaces.}}
    \label{fig:apx-dlbcl}
\end{figure}
\begin{figure}[h!]
    \centering
    \vspace*{-5mm}~\includegraphics[width = .8\linewidth]{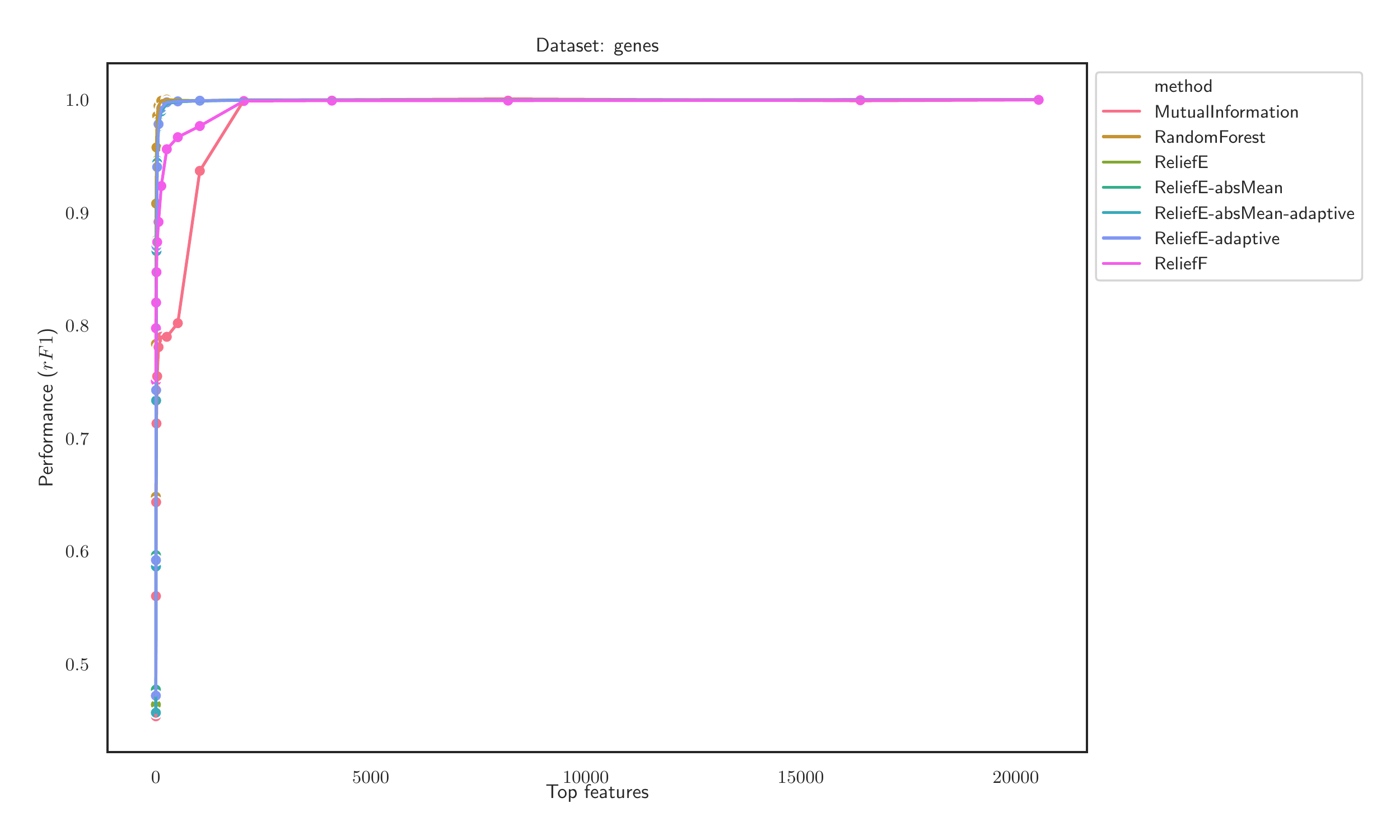}
    \caption{Genes performance curves. {Compared to mutual information (myopic)-based rankings ReliefE performs consistently better (steeper curve at the beginning in the first around hundred features.}}
    \label{fig:apx-genes}
\end{figure}

The power of ReliefE is apparent when considering DLBCL data set (very high dimensional with not many instances). Results are shown in Figure~\ref{fig:apx-dlbcl}.
Finally, the results for the genes data set are shown in Figure~\ref{fig:apx-genes}. Note how the more time expensive SURF variants were not able to finish in dedicated time. Further, ReliefF is notably worse, requiring more information to detect the relevant signal. On the other hand ReliefE variants perform consistently well.

\section{Multi-label classification -- additional rank diagrams}
\label{appendix-mlc}

\begin{figure}
    \centering
        \includegraphics[width = \linewidth]{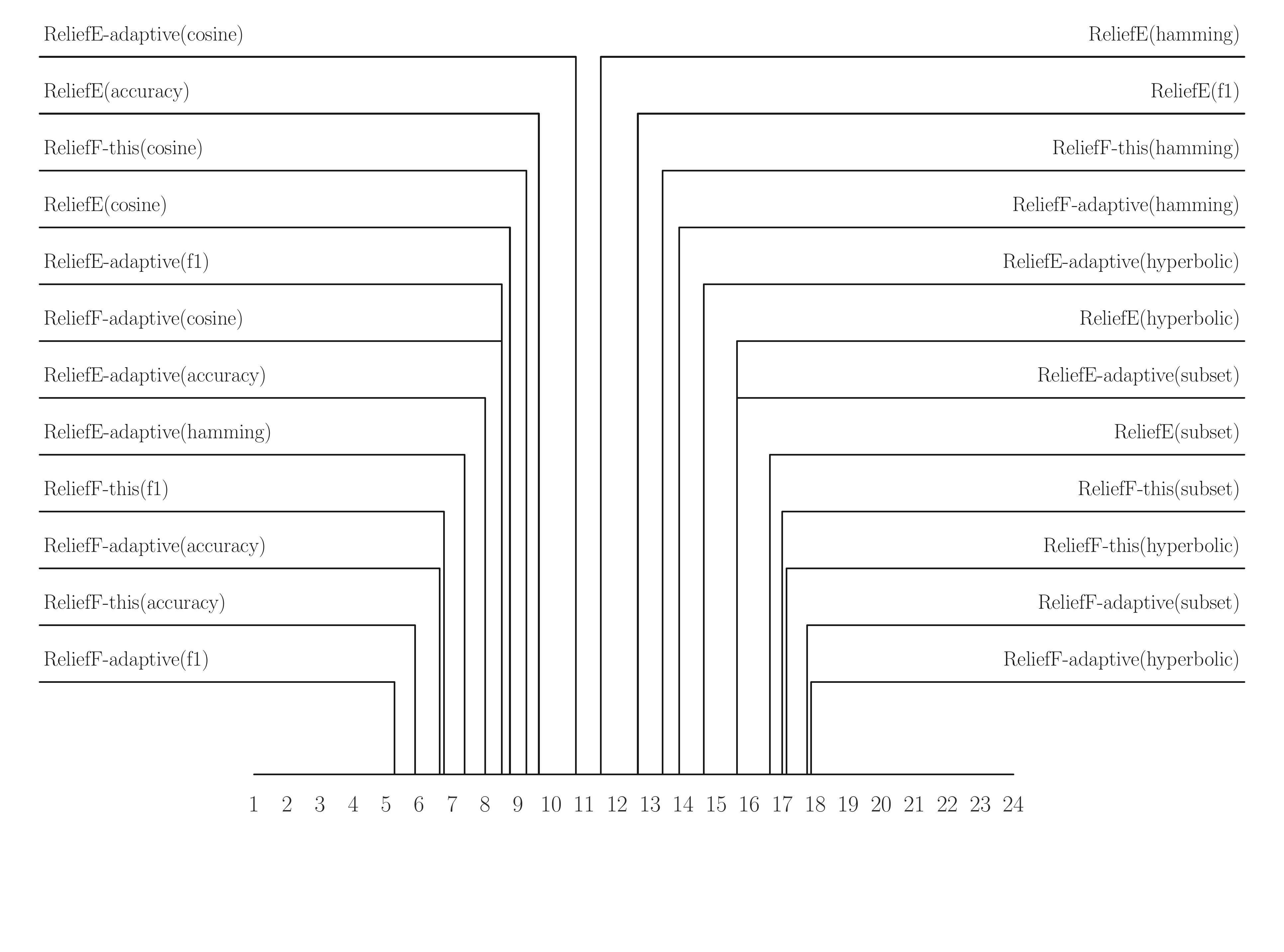}~\vspace*{-10mm}
    \caption{Max (upper) F1 scores for top 10 highly-ranked features. {The results indicate that the adaptive threshold step impacts the placing of the top features (adaptive) version of ReliefF.}  \label{fig:mlc1}}
   
\end{figure}

\begin{figure}
    \centering
        \includegraphics[width = \linewidth]{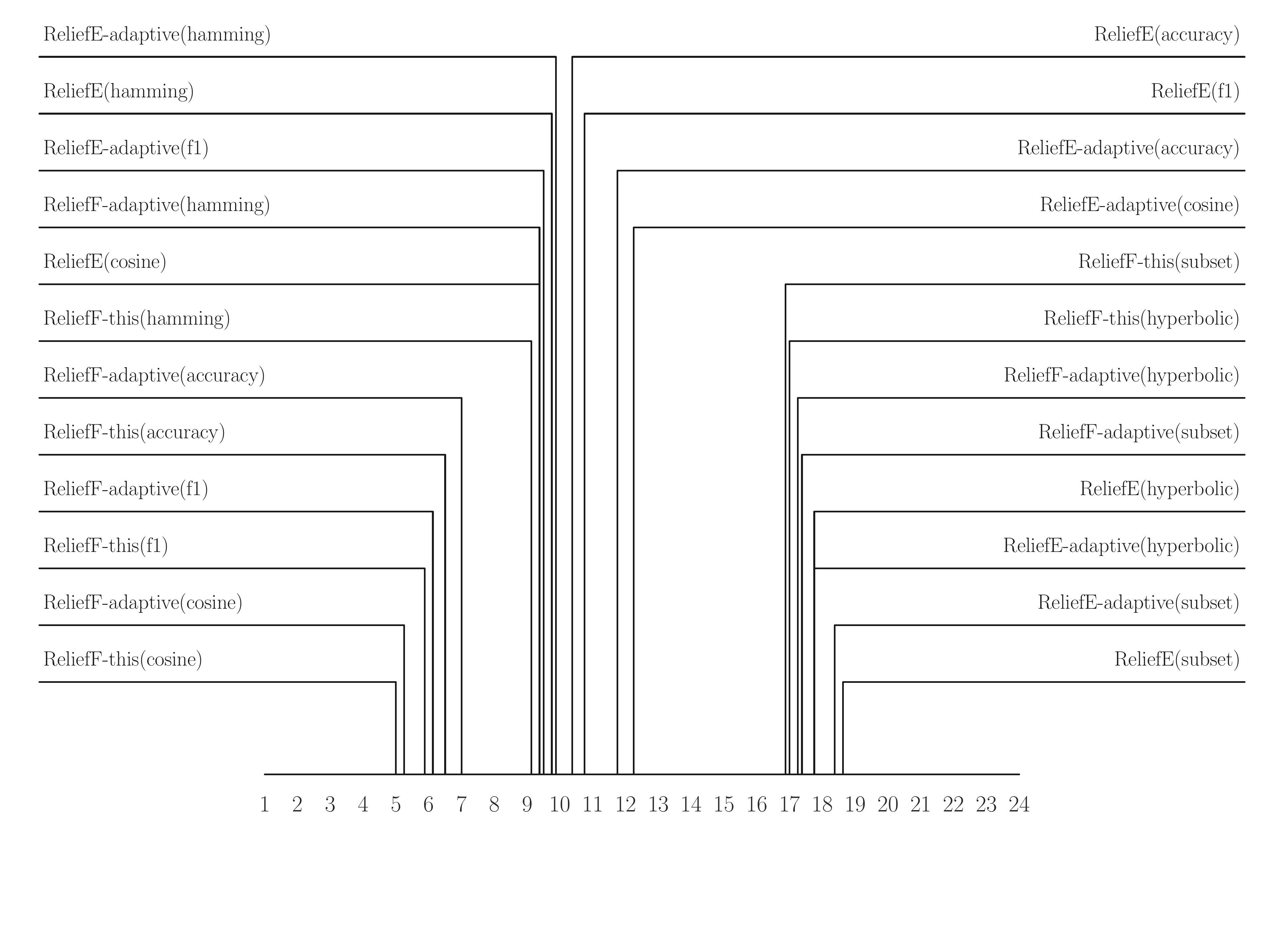}~\vspace*{-10mm}
    \caption{Mean (upper) F1 scores for top 10 highly-ranked features. {The average rank diagrams confirm the finding that if the target space is embedded via cosine distance, the MLC ReliefF variant performs the best.}         \label{fig:mlc2}}

\end{figure}

In this section, we present the average rank diagrams that offer insights into global distribution of the performances when multi-label classification setting is considered (Figures \ref{fig:mlc1} and \ref{fig:mlc2}).


\protect{\pagebreak}
%\floatbarrier

\section{Multi-label classification -- case study with Delicious}
\label{appendix-mlc:individual}

\begin{figure}[hb!]
\centering
    \includegraphics[width = .94\linewidth]{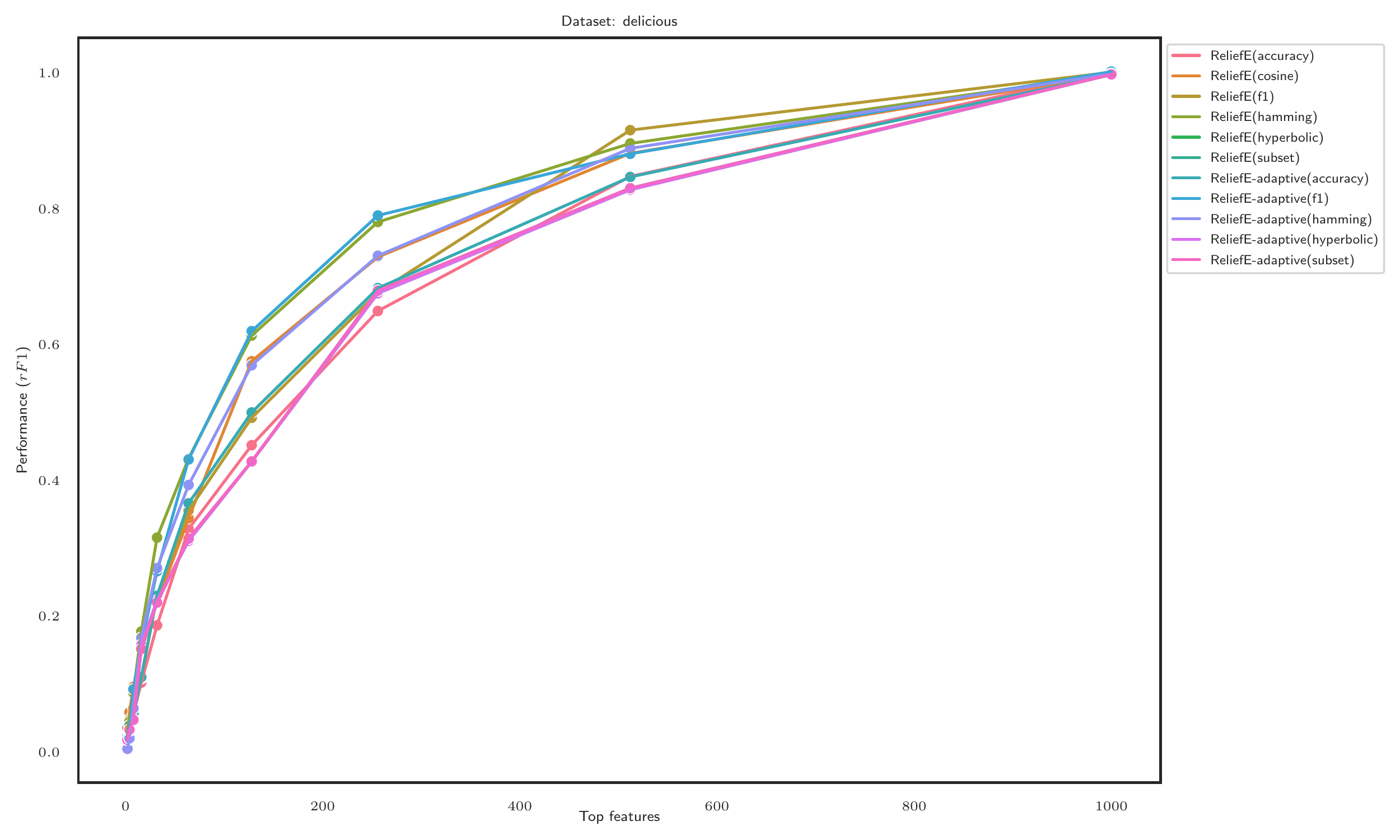}
    \caption{Delicious performance curves. {The ReliefE variants perform consistently better the first for up to 250 features.}}
    \label{fig:apx-del}
\end{figure}

We study in more detail the performance on the Delicious data set, as it offers interesting insights into the algorithms' performances (Figure~\ref{fig:apx-del}).
The algorithms' performances are overall consistent. Note how cosine-based embeddings of the target space emerge as the best option (orange line), indicating embedding-based distances amongst the target instances can already offer competitive performance.

\end{document}